\documentclass{article}
\usepackage[round, sort]{natbib}

\usepackage[preprint]{neurips_2024}

\usepackage[utf8]{inputenc} % allow utf-8 input
\usepackage[T1]{fontenc}    % use 8-bit T1 fonts
\usepackage{hyperref}       % hyperlinks
\usepackage{url}            % simple URL typesetting
\usepackage{booktabs}       % professional-quality tables
\usepackage{amsfonts}       % blackboard math symbols
\usepackage{nicefrac}       % compact symbols for 1/2, etc.
\usepackage{microtype}      % microtypography
\usepackage{xcolor}         % colors

\usepackage{graphicx}
\usepackage{amsmath}
\usepackage{amssymb}
\usepackage{multirow}
\usepackage[caption=false,font=scriptsize,labelfont=sf,textfont=sf]{subfig}
\usepackage{makecell}
\usepackage{pifont}
\usepackage{mathtools}
\usepackage{colortbl}
\usepackage[export]{adjustbox}
\usepackage{float}
\usepackage{subcaption}

\title{SSRFlow: Semantic-aware Fusion with Spatial Temporal Re-embedding for Real-world Scene Flow}
\author{%
  Zhiyang Lu \\
  Department of Computer Science\\
  Xiamen University\\
  \texttt{zhiyang@stu.xmu.edu.cn} \\
  \And
  Qinghan Chen \\
  Department of Computer Science \\
  Xiamen University \\
  \texttt{chenqinghan@stu.xmu.edu.cn} \\
  \AND
  Zhimin Yuan \\
  Department of Computer Science \\
  Xiamen University \\
  \texttt{zhiminyuan@stu.xmu.edu.cn} \\
  \And
  Chenglu Wen \\
  Department of Artificial Intelligence \\
  Xiamen University \\
  \texttt{clwen@xmu.edu.cn} \\
  \And
  Ming Cheng\thanks{Corresponding Author} \\
  Department of Computer Science \\
  Xiamen University \\
  \texttt{chm99@xmu.edu.cn} \\
  \And
  Cheng Wang \\
  Department of Computer Science \\
  Xiamen University \\
  \texttt{cwang@xmu.edu.cn} \\
}

\begin{document}

\maketitle

\begin{abstract}
    Scene flow, which provides the 3D motion field of the first frame from two consecutive point clouds, is vital for dynamic scene perception. However, contemporary scene flow methods face three major challenges. Firstly, they lack global flow embedding or only consider the context of individual point clouds before embedding, leading to embedded points struggling to perceive the consistent semantic relationship of another frame. To address this issue, we propose a novel approach called Dual Cross Attentive (DCA) for the latent fusion and alignment between two frames based on semantic contexts. This is then integrated into Global Fusion Flow Embedding (GF) to initialize flow embedding based on global correlations in both contextual and Euclidean spaces. Secondly, deformations exist in non-rigid objects after the warping layer, which distorts the spatiotemporal relation between the consecutive frames. For a more precise estimation of residual flow at next-level, the Spatial Temporal Re-embedding (STR) module is devised to update the point sequence features at current-level. Lastly, poor generalization is often observed due to the significant domain gap between synthetic and LiDAR-scanned datasets. We leverage novel domain adaptive losses to effectively bridge the gap of motion inference from synthetic to real-world. Experiments demonstrate that our approach achieves state-of-the-art (SOTA) performance across various datasets, with particularly outstanding results in real-world LiDAR-scanned situations. Our code will be released upon publication.
\end{abstract}

\section{Introduction}
% Understanding 3D scenes is a crucial task of computer vision, especially for autonomous driving. 
3D scene flow estimation captures the motion information of objects from two consecutive point clouds and produces the motion vector for each point in the source frame. It serves as a foundational component for perceiving dynamic environments and provides important motion features to downstream tasks, such as object tracking~\cite{zhou2022pttr-objectTrack1,yang2022temporal-objecttrack,zou2022traffic-TITS-ObjectTrack}, point cloud label propagation~\cite{zhao2023semanticflow-semanticSegmentation} and pose estimation~\cite{ding2023milliflow}. Early approaches\cite{basha2013multi-rgbd,vogel2013piecewise-rgbd,wedel2008efficientSF-stereo-rgbd,huguet2007variational-rgbd-stereo} rely on stereo or RGB-D images as input. While recent advances in deep learning-based point cloud processing have paved the way for numerous end-to-end algorithms specifically designed for scene flow prediction~\cite{liu2019flownet3d,wu2020pointpwc,puy2020flot,cheng2022bi-flow,wang2022whatmatters,li2022rppformer}. Among them, FlowNet3D~\cite{liu2019flownet3d} presents a pioneering approach that integrates deep learning into the estimation of scene flow. 
By incorporating the principles of the PWC (Pyramid, Warp and Cost volume) optical flow algorithm~\cite{Sun_2018_CVPR_PWC_opticalflow}, PointPWC~\cite{wu2020pointpwc} introduces the coarse-to-fine strategy to scene flow prediction. 
However, the PWC frameworks~\cite{cheng2022bi-flow, li2022rppformer, wu2020pointpwc, Cheng_2023_ICCV_MSBRN, wang2021hierarchical-fullSF, wang2022residual-Res3DSF-fullSF} only account for scene flow regression of the local receptive field within each level, which neglects global feature matching. Hence, it is difficult to estimate precise motion for long-distance displacements and complex situations such as repetition and occlusion. 

FlowStep3D\cite{kittenplon2021flowstep3d} and WM3DSF\cite{wang2022whatmatters} tackle this issue by using global flow initialization in the all-to-all manner. However, they neglect the alignment of semantic space between the embedded points and the context of another frame, see \figurename~\ref{fig::distVis} of Appendix. This hard approach to global flow embedding results in ambiguous flows. Hence, inspired by the fusion and alignment capability of cross-attention\cite{gheini2021cross-crossAttention,xu2023cross-crossAttentionAlignment,mital2023neural-crossAttentionAlignment}, we introduce the Dual Cross Attentive (DCA) Fusion to merge the semantic contexts of point clouds from two frames in latent space, which allows for perceiving the semantic context of another frame before embedding. By integrating into the Global Fusion Flow Embedding (GF) module for global flow embedding, DCA Fusion aggregates embedded features in both context and Euclidean spaces, leveraging the global correlations of two consecutive point clouds.

The second issue is attributed to the warping layer, which upsamples sparse scene flow from the previous level and accumulates to the current level. Previous methods~\cite{wu2020pointpwc,cheng2022bi-flow,li2022sctn,wang2022whatmatters, Cheng_2023_ICCV_MSBRN} simply employ the information preceding the warping layer to predict the residual flow for the subsequent layer. However, the temporal relation between the consecutive frames changes during warping since the two frames become closer, and the relative spatial position of points within the source frame also transforms. Utilizing the original features could introduce bias in residual flow estimation after warping layer-by-layer. To overcome this limitation, we propose a Spatial Temporal Re-embedding (STR) module to re-embed the temporal features between the warped source frame and target frame, along with spatial features within the warped source frame per se.

Furthermore, as a point-level task, obtaining the ground truth~(GT) of scene flow from real-world point clouds is difficult\cite{menze2018object-kitti2018,menze2015joint-kitti2015}, and previous methods resort to synthetic datasets\cite{mayer2016large-fly3d} for training. However, they suffer from domain gaps when applied to real-world LiDAR-scanned scenes. To address this issue, we propose novel Domain Adaptive Losses (DA Losses) based on the intrinsic properties of point cloud motion, including local rigidity of dynamic objects and the cross-frame feature similarity after motion, suggesting promising results when generalized to real-world datasets.

Overall, our contributions are as follows:
\begin{itemize}
\item Our GF module leverages the dual cross-attentive mechanism to fuse and align the semantic context from both frames and further matches the all-to-all point-pairs globally from both latent context space and Euclidean space, enabling accurate flow initialization for subsequent residual scene flow prediction. 
\item We elaborate the STR module to tackle the problems caused by distortion in surface spatiotemporal sequence features of two consecutive frames after warping.
\item We propose novel DA Losses that address the synthetic-to-real challenge of the scene flow task by considering the local rigidity and cross-frame feature similarity.
\item Experiments demonstrate that our model achieves SOTA performance on datasets of various patterns and exhibits strong generalization on real-world LiDAR-scanned datasets.
\end{itemize}

\section{Methodology}

\subsection{Problem Definition}
The scene flow task aims to estimate point-wise 3D motion information between two consecutive point cloud frames. The input includes the source frame $ S = \{ s_{i} \}_{i=1}^{N} = \{x_i,f_i \}_{i=1}^N  $ and target frame $ T = \{ t_{j} \}_{j=1}^{M} =  \{y_j,g_j \}^M_{j=1} $, where $x_i$, $y_j\in\mathbb{R}^3$ are 3D coordinates of the points, and $f_i$, $g_j\in\mathbb{R}^d$ represent the feature of the corresponding point at a specific level. It should be noted that $N$ and $M$ may not be equal due to the uneven point density and occlusion. The prediction of the model is the 3D motion vector $ SF = \{ sf_{i}\in\mathbb{R}^{3}  \}_{i=1}^{N} $ of each source frame point, representing the non-rigid motion towards the target frame. 

% \subsection{Network Architecture}
% Our proposed network architecture (shown in \figurename~\ref{fig: framework}) mainly consists of three components: 
% (1)	Hierarchical point cloud feature extraction.
% (2)	Global attentive flow initialization.
% (3)	Local flow refinement.
% %这一段与framework的caption高度重合，是否可以放在补充材料中去%
% We first leverage the pyramid feature extraction network to hierarchically capture the semantic features of the point clouds. Then, we construct a global attentive flow embedding based on both the high-dimensional feature space and Euclidean space. Afterward, through the upsampling and warping layers, the sparse flow from higher levels is upsampled to acquire the dense flow at lower levels, which is then accumulated onto the source frame to result in the warped source frame. Among this, we hierarchically refine the residual scene flow between the warped source frame and the target frame through spatiotemporal re-embedding features. This iterative refinement process leads to the full resolution scene flow.

\begin{figure}[t]
    \centering
    \includegraphics[width=1.0\linewidth]{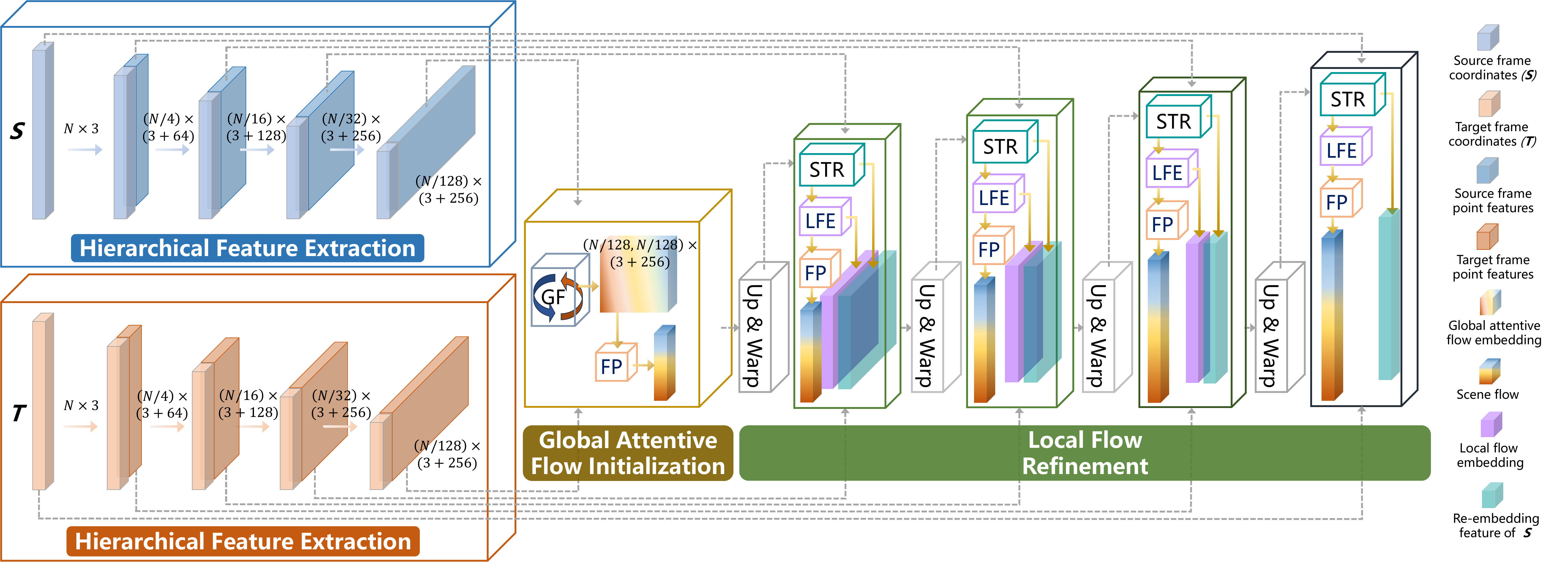}
    \caption{Illustration of the proposed network. Firstly, semantic features are hierarchically extracted and sent to GF to achieve global embedding between the two point clouds at the highest level. Then, the Flow Prediction (FP) module produces the initial scene flow. Subsequently, the flow and features are upsampled level by level, and the upsampled flow is accumulated onto the source frame by the warping layer. Afterwards, Spatial Temporal Re-embedding (STR) and  Local Flow Embedding (LFE) are performed in turn, and FP yields the refined flow at a specific level.}
    \label{fig: framework}
    % \vspace{-10pt}
\end{figure}

\subsection{Hierarchical Feature Extraction}
The overview of our proposed network is shown in \figurename~\ref{fig: framework}. We utilize PointConv~\cite{wu2019pointconv} as the feature extraction backbone to build a pyramid network. To extract the higher-level semantic feature $S_{l+1}$ of level $ (l+1) $, we apply a three-step process to the previous lower-level feature $ S_{l} $. Farthest Point Sampling (FPS) is first employed to extract $ N_{l+1} $ center points from  $ S_{l} $, where $N_{l+1}<N_l$. Next, K-Nearest Neighbor (KNN) is used to group the neighbor points around each center point. Finally, PointConv is utilized to aggregate the local features for each group, resulting in the desired semantic feature $ S_{l+1} $.

% The semantic feature of the target frame is obtained in exactly the same way.

\subsection{Global Fusion Flow Embedding}
\begin{figure}[t]
    \centering
    \includegraphics[width=1\linewidth]{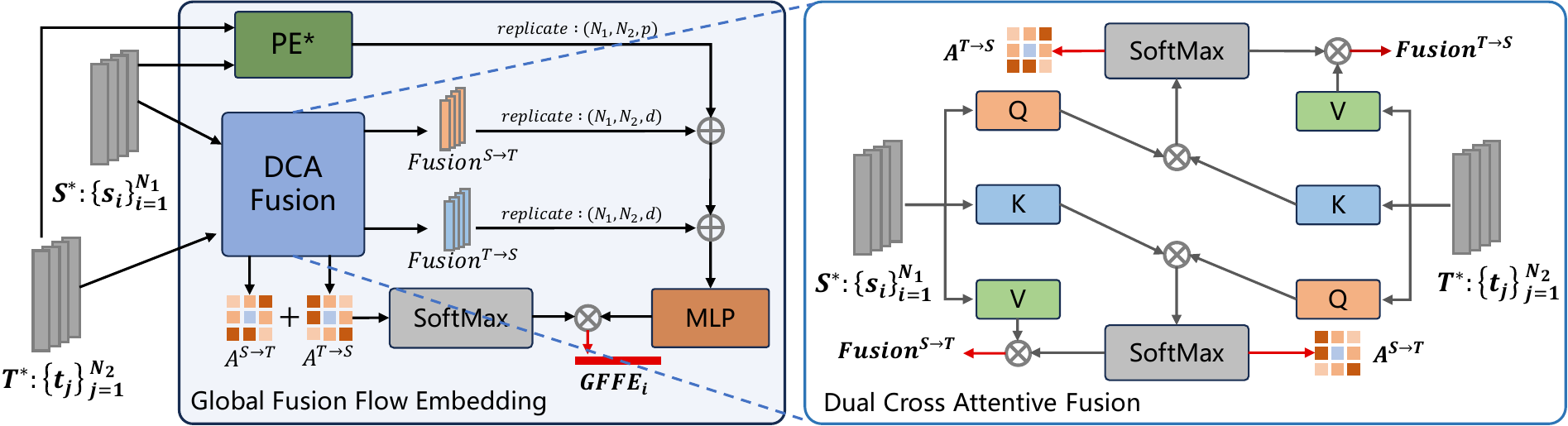}
    \caption{Flowchart of global flow embedding. $\otimes$ and $\oplus$ denote multiplication and concatenation, respectively.}
    \label{fig: GF}
\end{figure}

%前面有写，这里可以直接去掉%
%Scene flow initialization plays a crucial role in coarse-to-fine methods which focus on locally refining the residual scene flows layer by layer while lacking global matching information. 
The GF module is designed to capture the global relation between consecutive frames during the flow initialization. After performing the multi-level
%down-sampling
feature extraction, we obtain $ S^{*} $ and $ T^{*} $ at the highest level of the semantic pyramid. Following that, the global fusion flow embedding is constructed from $ S^{*} $ to all points in $ T^{*} $ in both semantic context space and Euclidean space, as shown in \figurename~\ref{fig: GF}.
The previous algorithms\cite{wang2022whatmatters,kittenplon2021flowstep3d} merely utilized the individual and unaligned semantic features of two consecutive point clouds for hard global embedding. However, the flow embedding of a point is generated in response to the semantic context of another frame, necessitating the simultaneous consideration of the fused features in a consistent semantic space between the two frames during embedding.
To enhance mutual understanding of semantic context between two frames of point clouds, we first utilize the DCA module to fuse and align semantic context from both frames. This equips each frame with the ability to perceive the global semantic environment of the other frame, leading to a more reliable latent correlation.

Specifically, within the DCA module, we employ a cross-attentive mechanism to merge the semantic context of the highest layers in the feature pyramid, yielding an attentive weight map used for subsequent global aggregation, as illustrated in Figure \ref{fig: GF}.
During the dual cross-attentive fusion phase, the semantic context in the latent feature space is obtained for $ S^{*} \text{ and } T^{*} $ through linear networks Q K and V. Subsequently, Q($ T^{*} $) serves as the Query, while K($ S^{*} $) serves as the Key for computing the cross-attention map $ A^{S\to T} $. The final fused features from $ S^{*} $ to $ T^{*} $ are calculated via V($ S^{*} $).
\begin{equation}
    A^{S \to T} = \sigma (\frac{\text{Q}(T^{*})\cdot \text{K}(S^{*})}{\sqrt{d_a}}),
\end{equation}
\begin{equation}
    Fusion^{S \to T} = A\cdot \text{V}(S^{*}),
\end{equation}
where $d_{a}$ is the output dimension of linear network K and V, and $\sigma$ denotes SoftMax. The fusion of context features from $T^{*}$ to $S^{*}$ follows a similar procedure.

After acquiring the fused features in the semantic space, the GF module initializes the flow embedding for two frames. We elucidate the process using a point $s_{i}$ from $S^{*}$ as a case to enhance clarity.
Firstly, to establish the relative positional association between each point-pair, a position encoder $PE^{*}$ in Euclidean space is introduced as follows, where $ \eta $ denotes concatenation.
\begin{equation}
    PE_{ij}= \eta ( x_{i}, y_{j} , y_{j}-x_{i}),
    \label{eq: PEm}
\end{equation}
\begin{equation}
        PE^{*}_{ij} = \eta (PE_{ij},\mathrm{MLP}(PE_{ij})).
    \label{eq: PEm*}
\end{equation}
The external position encoder $ PE^{*} $ instead of internal integration in the DCA module provides explicit position context during global flow embedding.
Then we proceed to construct the initial global flow embedding $ GFE=\{GFE_{ij}\} $ from both fusion semantic context and Euclidean space. The latent embedding of point-pair $ s_{i} $ and $ t_{j} $ is represented as:
\begin{equation}
    GFE_{ij}=\mathrm{MLP}(\eta (Fusion^{T->S}_{i},Fusion^{S->T}_{j}, PE^{*}_{ij})),
    \label{eq: AC}
\end{equation}
where $ \eta $ denotes dimension concatenation.
After obtaining the dual cross-attentive maps $A^{S->T}$ and $A^{T->S}$ within the DCA module, we perform element-wise addition and then pass them through SoftMax to obtain the aggregation weights $ W = \{W_{ij}\} $ for the global flow embedding aggregation, which was later proven to be superior to the MaxPooling.
Finally, the initial global flow embedding is aggregated by utilizing the aggregation map $ W $, to obtain the global fusion flow embedding from the specific point $s_{i}$ to all points in the target:
\begin{equation}
    GFFE_{i} =  {\textstyle \sum_{j}} W_{ij} \cdot GFE_{ij}.
    \label{eq: GFFE}
\end{equation}
Once $ GFFE =  \{GFFE_i\}  $ has been obtained, it is then fed into the flow predictor (described in Section~\ref{sec::fp}) to generate the global initial scene flow.

% 一种表示方法【上下标】：M^{i}_{j} = \sigma (MLP(g_{j}\odot f_{i} ) )
% 另一种表示方法【矩阵】：\sideset{}{}{\mathop{\sigma}}_{j=1,\cdots,m}

\subsection{Warping Layer}
We employ distance-inverse interpolation to upsample the coarse sparse scene flow from level $ (l+1) $ to obtain the coarse dense scene flow of level $ l $. The obtained coarse dense flow is directly accumulated onto the source frame $ S_{l} $ to generate the warped source frame $ WS_{l}=\{ ws_{i} \}_{i=1}^{N_{l}}=\{wx_{i}=x_{i}+sf_{i},f_{i} \}_{i=1}^{N_{l}} $, which brings the source and target frames closer and allows the subsequent layers to only consider the estimation of residual flow~\cite{wu2020pointpwc, cheng2022bi-flow,wang2022whatmatters,wei2021pv,fu2023pt-flownet-transformerSF-fullSF,kittenplon2021flowstep3d}.

\subsection{Spatial Temporal Re-embedding}
\begin{figure*}[t]
   \centering
   \includegraphics[width=1.0\textwidth]{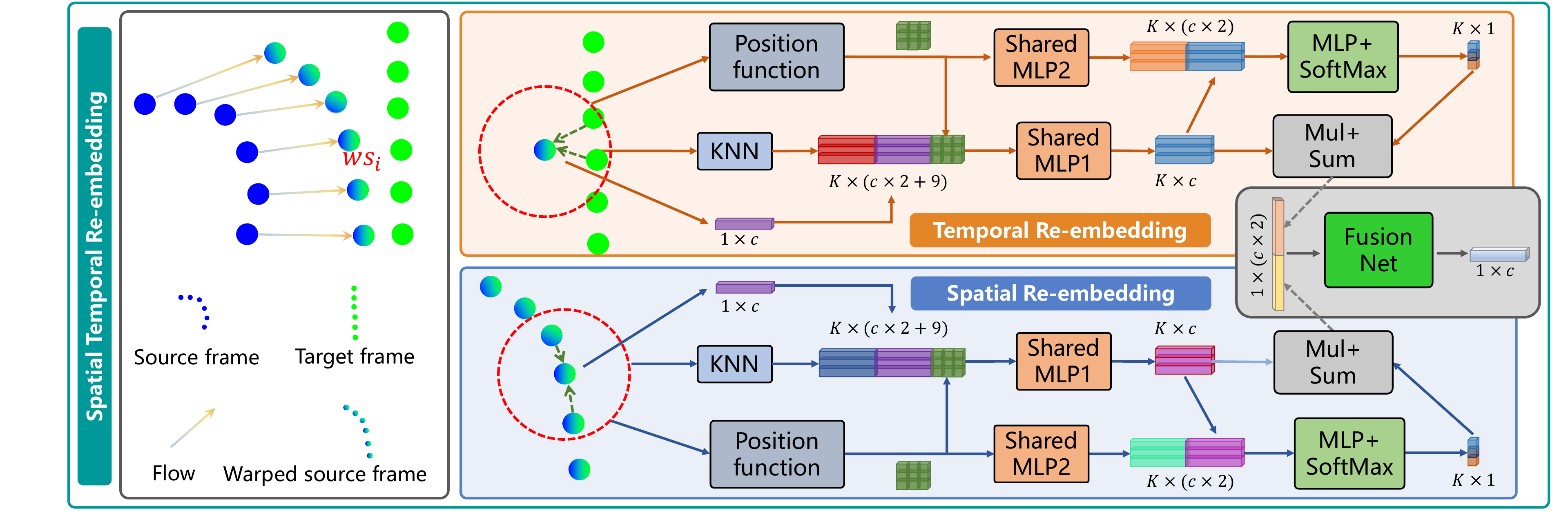}
   \caption{The details of STR module.}
   \label{fig: STR}
\end{figure*}

After the warping layer, the spatiotemporal relation between the consecutive frames may change. Specifically, the temporal features of points from the warped source frame to the target change since the position between the two point clouds is closer. Furthermore, dynamic non-rigid objects in the source frame may encounter surface distortion during warping, resulting in different spatial features. Therefore, it is necessary to re-embed spatiotemporal point features before the Local Flow Embedding (LFE), which is implemented in a patch-to-patch manner between the two frames following~\cite{wu2020pointpwc}. Based on this consideration, we re-embed the spatiotemporal features of each point $ ws_{i}$ at level $l$ based on the warped source frame $ WS_{l} $ and the target frame $ T_{l} $, as depicted in \figurename~\ref{fig: STR}.

\textbf{Temporal Re-embedding} First, we locate the $K$ nearest neighbor points group $\mathcal{N}_{T}(ws_{i})$ of point $ ws_{i} $ in $ T_{l} $. For each target point $ t_{j} \in \mathcal{N}_{T}(ws_{i}) $, by employing position encoder as~\eqref{eq: PEm}, a 9D positional feature $PE_{ij}$ is acquired for this group, representing the positional relation between the two frames after warping. Then, the initial temporal re-embedding feature is derived using the following formula:
\begin{equation}
    TRF_{ij}= \mathrm{MLP}(\eta (g_{j},f_{i},PE_{ij})).
    \label{eq: TRF}
\end{equation}
Instead of employing the hard aggregation method of MaxPooling, which results in flow bias due to the non-corresponding points between the two frames, we leverage local similarity map $ LM_{i} =  \{ LM_{ij}\} $ in both feature space and Euclidean space to derive the soft aggregation weights, 
\begin{equation}
    LM_{ij} = \sigma (\mathrm{MLP}(\eta (TRF_{ij},\mathrm{MLP}(PE_{ij})))),   
    \label{eq: LM}
\end{equation}
\begin{equation}
    TRF^{*}_i =  {\textstyle \sum_{j}} LM_{ij} TRF_{ij}.
    \label{eq: TRF*}
\end{equation}
% The adverse effects of noisy points within the group can be mitigated by employing the soft aggregation method, decreasing the presence of Out3D in the scene flow. (Experimental analysis is conducted in Appendix.)

\textbf{Spatial Re-embedding} 
%As can be seen from \figurename~\ref{fig: STR}, 
Spatial Re-embedding shares the same framework as Temporal Re-embedding, with the only distinction being that the embedding object changes to the warped source frame itself.
Upon acquiring the temporal re-embedding features $ TRF^{*} = \{ TRF^{*}_i\} $ and spatial re-embedding features $ SRF^{*} = \{ SRF^{*}_{i}\} $ of each point in the warped source frame of level $l$, we fuse them by leveraging the Fusion Net module to derive the ultimate comprehensive features
\begin{equation}
    STRF_{i} = \mathrm{MLP} ( \eta (TRF^{*}_i,SRF^{*}_i)   ),
    \label{eq: STRF}
\end{equation}
and the warped frame updates to $ WS_{l} = \{ wx_{i},STRF_{i} \}_{i=1}^{N_{l}} $.
As shown in \figurename~\ref{fig: framework}, the STR module is followed by LFE, which computes the patch-to-patch cost volume of each point $ws_{i}$ by utilizing the spatiotemporal re-embedding features.

\subsection{Flow Prediction}\label{sec::fp}
This module is constructed by combining PointConv, MLP, and a Fully Connected (FC) layer. For each point $s_{i}$ in the source frame, its local flow embedding feature, along with the warped coordinates and $STRF_{i}$ are input into the module. PointConv is first employed to incorporate the local information of each point, followed by non-linear transformation in the MLP layer. The final output is the scene flow $ sf_{i} $, regressed through the FC layer.

\section{Training Losses}
\subsection{Hierarchical Supervised Loss}
A supervised loss is directly hooked to the GT of scene flow, and we leverage multi-level loss functions as supervision to optimize the model across various pyramid levels. 
The GT of scene flow at level $l$ is represented as $ \tilde{SF}_{l}={ \{ \tilde{sf}{}_{i}^{l}  \}} {}_{i=1}^{N_{l}} $ and the predicted flow is $SF_{l}=  {  \{ sf_{i}^{l}  \}}{}_{i=1}^{N_{l}}$. The multi-level supervised loss is as follows:
\begin{equation}
    \mathcal{L}_{sup}=\sum_{l=1}^{5}\frac{\delta_{l}}{N_{l}}  \sum_{i=1}^{N_{l}}   \|{\tilde{sf}{}_{i}^{l}}- {sf{}_{i}^{l}}  \| _{2},
\end{equation}
where $\delta$ is the penalty weight, with $\delta_{1}=0.02,\delta_{2}=0.04,\delta_{3}=0.08,\delta_{4}=0.16$, and $\delta_{5}=0.32$.

\subsection{Domain Adaptive Losses}
\textbf{Local Flow Consistency (LFC) Loss}
Dynamic objects in real-world scenes may not exhibit absolute rigid or regular motion. Instead, they typically undergo local rigid motion, which is manifested through the consistency of local flow. The degree of predicted flow difference between each point $s_{i}$ and its KNN+Radius points group $ \mathcal{N}_{S}^{R}(s_{i}) $ in the source frame at the full resolution level ($ N_1 =8192$) is defined as the LFC loss, where point $ p \in \mathcal{N}_{S}^{R}(s_{i}) $ denotes $p \in \mathcal{N}_{S}(s_{i})$ and the $\ell_2$ distance between $ p \text{ and } s_{i}$ is less than $ R $. The KNN+Radius search strategy effectively mitigates the influence of noise points resulting from occlusion and sparsity in point clouds, as demonstrated in Sec \ref{sup-sec::searchMethods} of Appendix. Formally, the LFC loss is represented as follows:
\begin{equation}
     \mathcal{L}_{lfc} = \frac{1}{N_{1}}{\sum\limits_{i=1}^{N_1}} \frac{1}{|\mathcal{N}_{S}^{R}(s_{i})|} { \sum\limits_{ s_{j} \in { \mathcal{N}_{S}^{R}(s_{i})} } ^ {}} { \| sf_{i}-sf_{j}  \| }_{2},
\end{equation}
where $ |\cdot| $ is the number of points in a group. 

%It is noteworthy that 
%To mitigate the influence of noise points resulting from occlusion and sparsity in point clouds, we employ the KNN+Radius search method to acquire the neighborhood around $s_{i}$ located within radius $ R $, as demonstrated in Figure \ref{Fig: knn_radius}.

\textbf{Cross-frame Feature Similarity (CFS) Loss} 
% The local features of the points in the source frame accumulated by scene flow are similar to those in the surrounding target frame
The semantic features of the points in the warped source frame are similar to those in the surrounding target frame, as they should be in a dynamic registered state. Specifically, we accumulate the GT scene flow $\tilde{sf} $ directly onto the source frame at the full resolution level, as described by 
$ \tilde{ws_{i}} = \{ x_{i} + \tilde{sf_{i}}, STRF_{i} \}. $
Next, we utilize cosine similarity to compute the similarity between $\tilde{ws_{i}}$ and $ t_{j} \in \mathcal{N}_{T}^{R}(\tilde{ws_{i}})  $ in the target frame:
%, as illustrated by
\begin{equation}
    CS ( \tilde{ws_{i}},t_{j}  )=\frac{{STRF_{i}} \odot  g_{j}}{{ \| STRF_{i}  \| }_{2}{ \| g_{j}  \| }_{2}}.
\end{equation}
% It is noteworthy that owing to the non-rigid deformation that transpires during motion, the initial features of points within the source frame and the target frame may not be similar. However, through continual warping and Spatial Temporal Re-embedding, the corresponding points between the two frames gradually reach a high local feature similarity effect. For this reason, 
We utilize the features of source frame points derived from the last layer of the STR module (the rightmost re-embedding feature in \figurename~\ref{fig: framework}) as input instead of the initially extracted features. Lastly, we establish a similarity threshold $ TH $ and employ function $ F $ to penalize points that exhibit a similarity lower than $ TH $:
\begin{equation}
    \mathcal{L}_{cfs}=\frac{1}{N_{1}} \sum\limits_{i=1}^{N_{1}}  \sum\limits_{t_{j}\in\mathcal{N}_{T}^{R}(\tilde{ws_{i}})}^{} \frac{F(CS(\tilde{ws_{i}},t_{j})-TH)}{|\mathcal{N}_{T}^{R}(\tilde{ws_{i}})|},
\end{equation}
where $F(x)=-x$ if $x<0$ and 0 otherwise,
%\begin{equation}
%    F(x)= \begin{cases}
%               0, & x\ge 0 \\ 
%               - x, & x< 0 
%\end{cases},
%\end{equation}
and $g_{j}$ is updated by the Temporal Re-embedding module of the STR module with the warped source frame for a reliable and precise loss. 
The final loss of our model is :
\begin{equation}
    \mathcal{L}_{all}=\lambda_{1} \mathcal{L}_{sup} + \lambda_{2} \mathcal{L}_{lfc} + \lambda_{3} \mathcal{L}_{cfs},
\end{equation}
where $ R=0.05$, $ TH=0.95$, and $\lambda_{1}=0.7$, $\lambda_{2}=0.15$, $\lambda_{3}=0.15$  by default.

\section{Experiments}

\subsection{Datasets and Data Preprocessing}
\begin{figure}[t]
    \centering
    \subfloat[FT3Ds]{
    \includegraphics[width=0.24\linewidth]{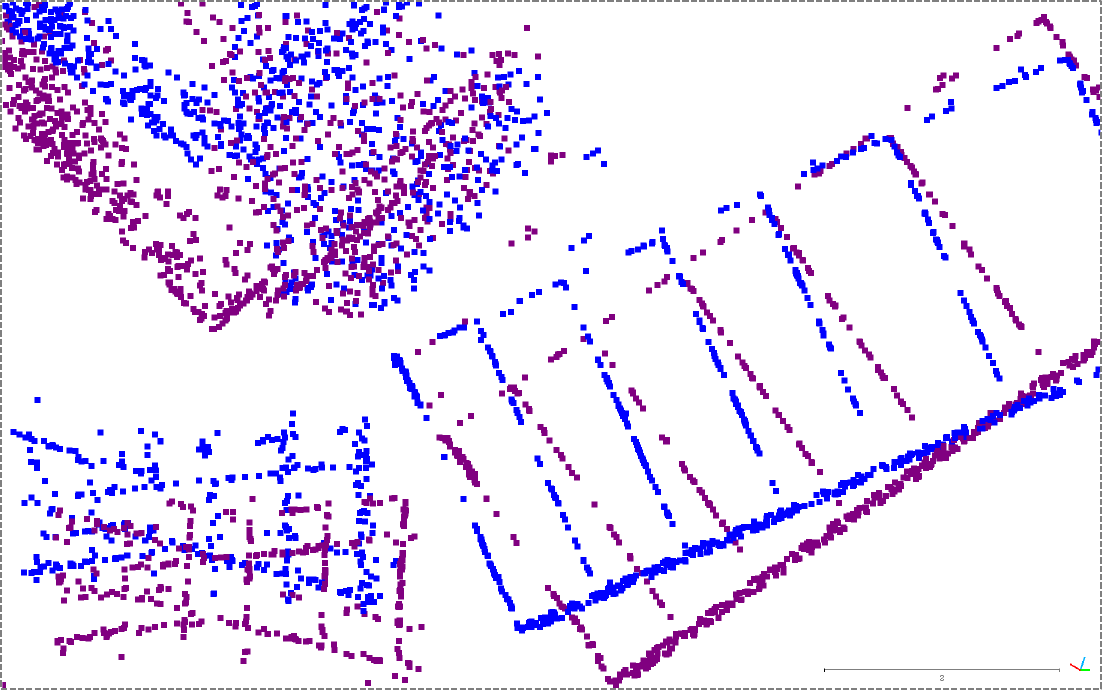}}
    \hfil
    \subfloat[KITTIs]{
    \includegraphics[width=0.24\linewidth]{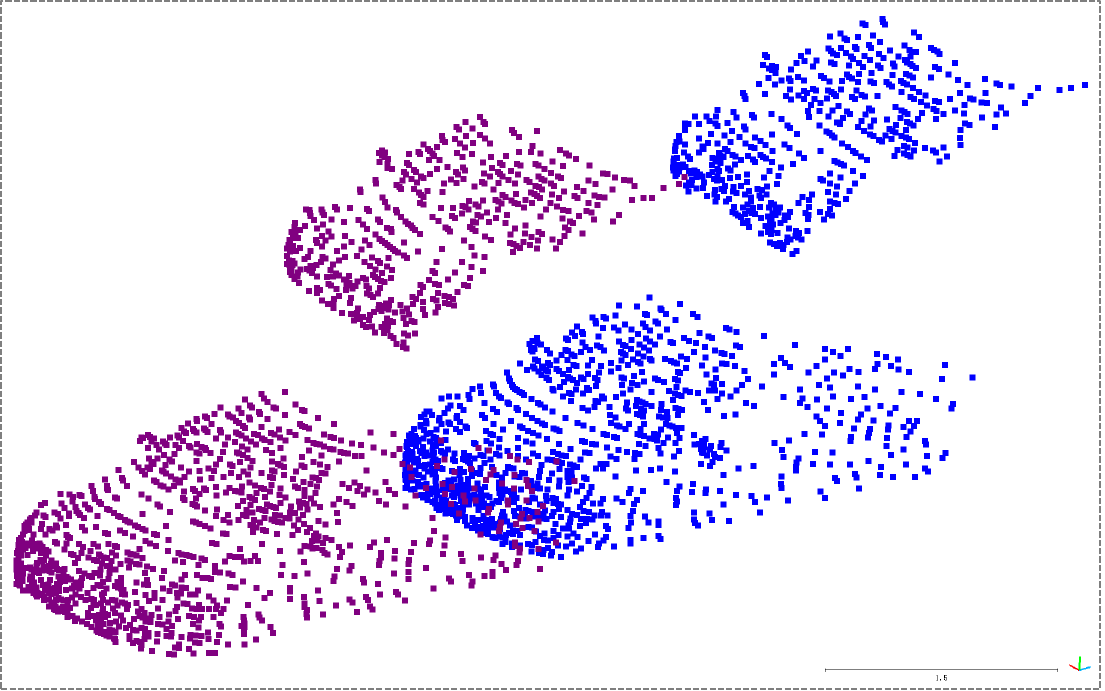}}
    \hfil
    \subfloat[SF-KITTI]{
    \includegraphics[width=0.24\linewidth]{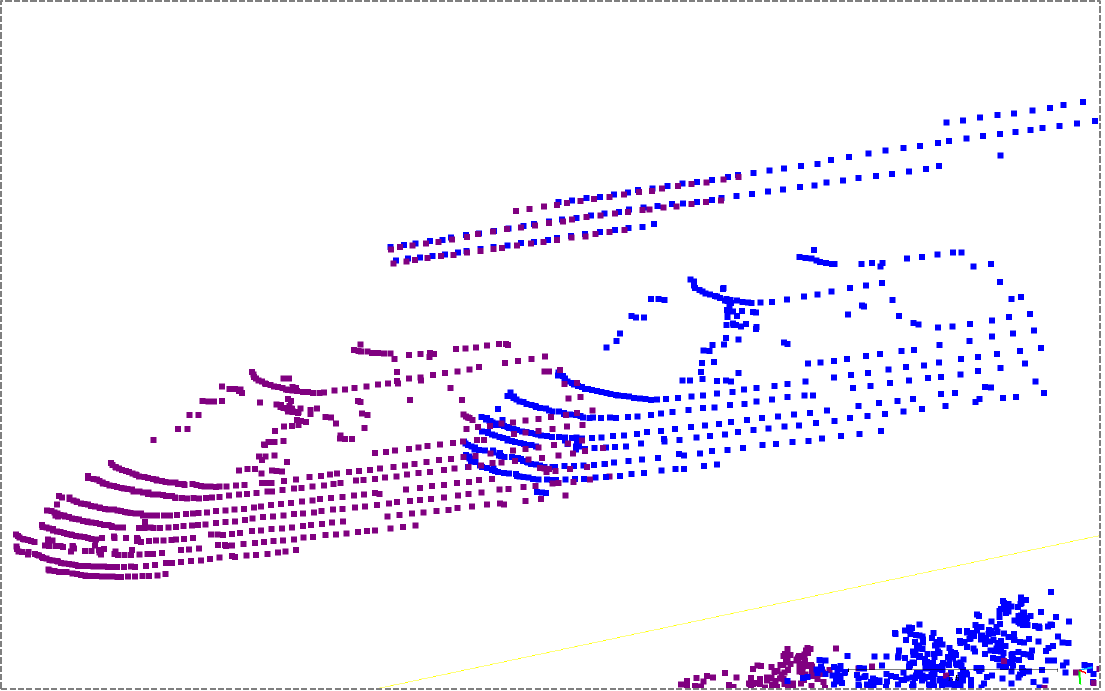}}
    \hfil
    \subfloat[LiDAR-KITTI]{
    \includegraphics[width=0.24\linewidth]{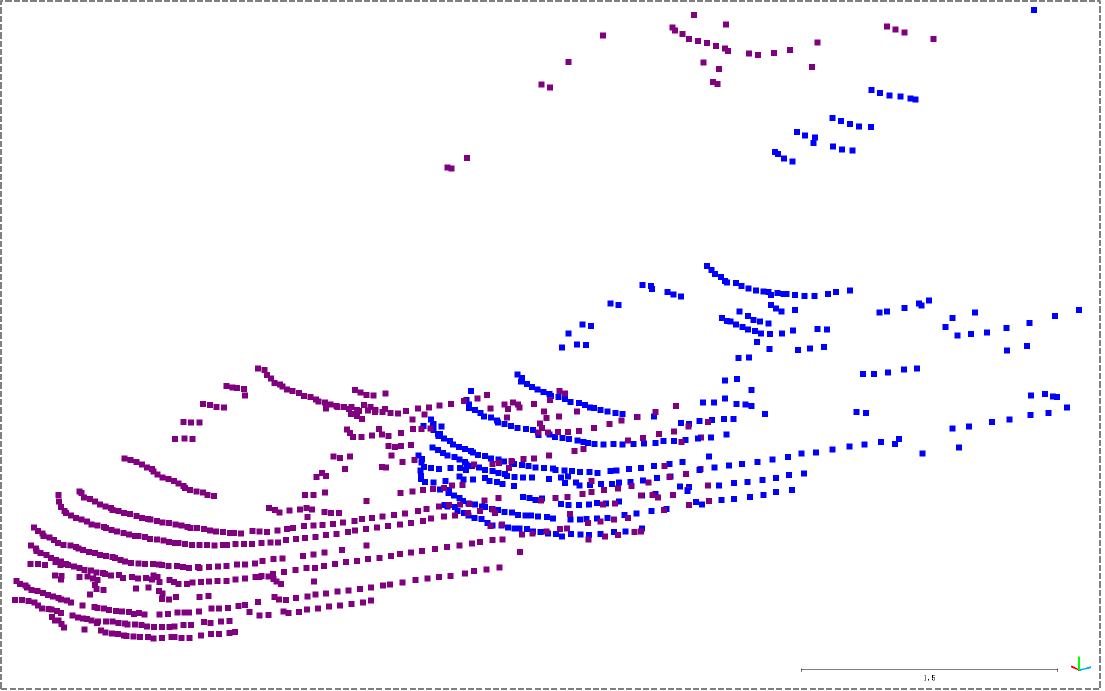}}
    
    \caption{Comparisons of scene flow datasets, including (a) synthetic stereo, (b) real-world stereo, and (c)(d) real-world LiDAR-scanned. Blue and purple denote the source and target frames, respectively.}
    \label{fig:CompareDatasets}
\end{figure}

The experiments were performed on four datasets: the synthetic dataset FlyThings3D (FT3D)~\cite{mayer2016large-fly3d} and three real-world datasets including Stereo-KITTI~\cite{menze2015joint-kitti2015,menze2018object-kitti2018},  SF-KITTI~\cite{ding2022fhnet-flow-SFKT}, and LiDAR-KITTI~\cite{geiger2012we-lidar-kt,gojcic2021weakly-rigid3dSF}, as shown in \figurename~\ref{fig:CompareDatasets}.
% FT3D is a synthetic dataset generated from ShapeNet~\cite{chang2015shapenet}, where random motions are assigned to each object within every scene, and Stereo-KITTI is a real-world dataset. 
These datasets are preprocessed in two ways~\cite{gu2019hplflownet,liu2019flownet3d}: FT3Ds and KITTIs remove non-corresponding points between consecutive frames, while FT3Do and KITTIo retain occluded points using mask labels.
% FT3D and Stereo-KITTI datasets are derived from dense and regular disparity images, distinct from real-world LiDAR-scanned datasets, as shown in \figurename~\ref{fig:CompareDatasets}. To showcase the robustness of our SSRFlow, we also conducted experiments on real-world LiDAR-scanned datasets
%\cite{ding2022fhnet-flow, gojcic2021weakly-rigid3d,geiger2012we-lidar-kt} SF-KITTI and LiDAR-KITTI, emphasize sparse and non-corresponding point attributes. Following~\cite{ding2022fhnet-flow-SFKT}, the SF-KITTI dataset is only used for training. 
Further dataset details can be found in Sec \ref{sup-sec::datasets} of Appendix.

% --------------non-occluded FT3D and KT2015 ------------------%
\def\w{20}
\begin{table}[t] \footnotesize
% \vspace{-5pt}
\centering
\setlength{\tabcolsep}{4.7pt}
\caption{ Performance comparisons on the FT3Ds and KITTIs datasets. All models in the table are only trained on FT3Ds and no fine-tuning is applied when tested on KITTIs. The best results for each dataset are marked in bold. * denotes the methods with an inference time exceeding 250 ms.} \label{tab:FTsKTs} 
% \resizebox{\textwidth}{60mm}{
\begin{tabular}{c|l|c|cccc|cc}
\toprule
Dataset & Method & Sup. & EPE3D${\downarrow}$ & AS3D${\uparrow}$ & AR3D${\uparrow}$  & Out3D${\downarrow}$ & EPE2D${\downarrow}$ & Acc2D${\uparrow}$ \\
\midrule
\multirow{8}{*}{FT3Ds} 
%---self--------------------------------
& \cellcolor{gray!\w}SPFlowNet* \cite{shen2023selfsuperpoint-selfSF}  & Self & 0.0606 & 0.6834 & 0.9074 & 0.3876 & -- & -- \\
\cmidrule(l){2-9}
%---full--------------------------------
% & FlowNet3D\cite{liu2019flownet3d}  & Full & 0.1136 & 0.4125 & 0.7706 & 0.6016 & 5.9740 & 0.5692 \\

% & HPLFlowNet\cite{gu2019hplflownet}  & Full & 0.0804 & 0.6144 & 0.8555 & 0.4287 & 4.6723 & 0.6764 \\

& PointPWC\cite{wu2020pointpwc} & Full & 0.0588 & 0.7379 & 0.9276 & 0.3424 & 3.2390 & 0.7994 \\

% & FLOT\cite{puy2020flot} & Full & 0.0520 & 0.7322 & 0.9276 & 0.3578 & -- & -- \\

% & HALFlow\cite{wang2021halflow} & Full & 0.0588 & 0.7379 & 0.9276 & 0.3424 & 3.2390 & 0.7994 \\
& \cellcolor{gray!\w}PV-RAFT\cite{wei2021pv}* & Full & 0.0461 & 0.8169 & 0.9574 & 0.2924 & -- & -- \\

& HCRF\cite{li2021hcrf-selfSF} & Full & 0.0488 & 0.8337 & 0.9507 & 0.2614 & 2.5652 & 0.8704 \\

% & FlowStep3D\cite{kittenplon2021flowstep3d} & Full & 0.0455 & 0.8162 & 0.9614 & 0.2165 & -- & -- \\

& SCTN\cite{li2022sctn} & Full & 0.0383 & 0.8474 & 0.9681 & 0.2686 & -- & -- \\

% & Bi-PointFlow\cite{cheng2022bi-flow} & Full & 0.0282 & 0.9184 & 0.9781 & 0.1436 & 1.5822 & 0.9296 \\

& WM3DSFNet\cite{wang2022whatmatters} & Full & 0.0281 & 0.9290 & 0.9817 & 0.1458 & 1.5229 & 0.9279 \\

% & RPPformer-Flow\cite{li2022rppformer} & Full & 0.0270 & 0.9211 & 0.9783 & 0.1178 & -- & -- \\

% & PT-Flow\cite{} & Full & 0.0304 & 0.9142 & 0.9814 & 0.1735 & 1.6150 & 0.9312 \\

& \cellcolor{gray!\w}MSBRN\cite{Cheng_2023_ICCV_MSBRN}* & Full & 0.0158 & 0.9733 & 0.9923 & \textbf{0.0568} & 0.8335 & 0.9703 \\

& SSRFlow (Ours)  & Full & \textbf{0.0122} & \textbf{0.9790} & \textbf{0.9942} & 0.0575 & \textbf{0.7891} & \textbf{0.9821} \\
\midrule

\multirow{9}{*}{KITTIs} 
%---self--------------------------------
& \cellcolor{gray!\w}SPFlowNet* \cite{shen2023selfsuperpoint-selfSF}  & Self & 0.0362 & 0.8724 & 0.9579 & 0.1771 & -- & -- \\
\cmidrule(l){2-9}
%---full--------------------------------
% & FlowNet3D\cite{liu2019flownet3d}  & Full & 0.1767 & 0.3738 & 0.6677 & 0.5271 & 7.2141 & 0.5093 \\

% & HPLFlowNet\cite{gu2019hplflownet}  & Full & 0.1169 & 0.4783 & 0.7776 & 0.4103 & 4.8055 & 0.5938 \\

& PointPWC\cite{wu2020pointpwc} & Full & 0.0694 & 0.7281 & 0.8884 & 0.2648 & 3.0062 & 0.7673 \\

% & FLOT\cite{puy2020flot} & Full & 0.0560 & 0.7550 & 0.9080 & 0.2420 & -- & -- \\

% & HALFlow\cite{wang2021halflow} & Full & 0.0622 & 0.7649 & 0.9026 & 0.2492 & 2.5140 & 0.8128 \\

& PV-RAFT\cite{wei2021pv} & Full & 0.0560 & 0.8226 & 0.9372 & 0.2163 & -- & -- \\

% & HCRF\cite{li2021hcrf} & Full & 0.0531 & 0.8631 & 0.9444 & 0.1797 & 2.0700 & 0.8656 \\

% & FlowStep3D\cite{kittenplon2021flowstep3d} & Full & 0.0546 & 0.8051 & 0.9254 & 0.1492 & -- & -- \\

% & SCTN\cite{li2022sctn} & Full & 0.0375 & 0.8730 & 0.9592 & 0.1793 & -- & -- \\

& Bi-PointFlow\cite{cheng2022bi-flow} & Full & 0.0307 & 0.9202 & 0.9603 & 0.1414 & 1.0562 & 0.9493 \\

& WM3DSFNet\cite{wang2022whatmatters} & Full & 0.0309 & 0.9047 & 0.9580 & 0.1612 & 1.1285 & 0.9451 \\

& \cellcolor{gray!\w}PT-Flow\cite{li2022rppformer}* & Full & 0.0224 & 0.9551 & 0.9838 & 0.1186 & 0.9893 & 0.9667 \\

& RPPformer-Flow\cite{li2022rppformer} & Full & 0.0284 & 0.9220 & 0.9756 & 0.1410 & -- & -- \\

& \cellcolor{gray!\w}MSBRN\cite{Cheng_2023_ICCV_MSBRN}* & Full & 0.0118 & 0.9713 & 0.9893 & 0.0856 & 0.4435 & 0.9853 \\

& SSRFlow (Ours)  & Full & \textbf{0.0059} & \textbf{0.9961} & \textbf{0.9993} & \textbf{0.0762} & \textbf{0.3292} & \textbf{0.9981} \\
\bottomrule
\end{tabular}
% }
\end{table}

\begin{table}[!htbp] \footnotesize
%\begin{table}[!htbp] \footnotesize
\vspace{-5pt}
%\centering
% \hspace{-1cm}
%----------------------subTab-------------------------
\begin{minipage}[c]{0.37\textwidth}
\centering
\setlength{\tabcolsep}{1.5pt}
% \vspace{-1cm}
\caption{Runtime and performance of the methods evaluated on KITTIs.} \label{tab:runtime}
\begin{tabular}{l|cc}
\toprule
Method & EPE3D${\downarrow}$ & \makecell{Run \\time (ms)} \\
\midrule
PointPWC\cite{wu2020pointpwc} & 0.0694  & 76ms \\
Bi-PointFlow\cite{cheng2022bi-flow} & 0.0307  & 80ms \\    
WM3DSF\cite{wang2022whatmatters} & 0.0309  & 63ms \\
PT-Flow\cite{fu2023pt-flownet-transformerSF-fullSF} & 0.0224  & 376ms \\
MSBRN\cite{Cheng_2023_ICCV_MSBRN} &0.0118 & 275ms \\
SSRFlow (Ours) & \textbf{0.0059} & 101ms \\
\bottomrule
\end{tabular}
\end{minipage}
%\hspace{0.08\textwidth}
\hspace{0.06\textwidth}
\begin{minipage}[c]{0.5\textwidth}
\centering
\setlength{\tabcolsep}{1.5pt}
\caption{Comparisons on the FT3Do and KITTIo datasets. All methods are trained only on FT3Do.}\label{tab:FtoKto}
% \resizebox{\linewidth}{22.5mm}{
\begin{tabular}{c|l|cccc}
\toprule
Dataset & Method & \makecell{EPE3D${\downarrow}$} & \makecell{AS3D${\uparrow}$} & \makecell{AR3D${\uparrow}$}  & \makecell{Out3D${\downarrow}$} \\
\midrule
\multirow{3}{*}{FT3Do} 
% & PointPWC\cite{wu2020pointpwc}  &  0.1552 & 0.4160 & 0.6990 & 0.6389 \\

% & OGSF\cite{ouyang2021ogsf}  & 0.1217 & 0.5518 & 0.7767 & 0.5180 \\

% & Bi-PointFlow\cite{cheng2022bi-flow}  & 0.0730 & 0.7910 & 0.8960 & 0.2740 \\

& WM3DSF\cite{wang2022whatmatters}  & 0.0630 & 0.7911 & 0.9090 & 0.2790 \\

& MSBRN\cite{Cheng_2023_ICCV_MSBRN}  & 0.0535 & 0.8364 & 0.9261 & 0.2314 \\

& SSRFlow (Ours)  & \textbf{0.0326} & \textbf{0.9152} & \textbf{0.9742} & \textbf{0.1398} \\
\midrule

\multirow{3}{*}{KITTIo} 

% & PointPWC\cite{wu2020pointpwc} & 0.1180 & 0.4031 & 0.7573 & 0.4966 \\

% & OGSF\cite{ouyang2021ogsf}  & 0.0751 & 0.7060 & 0.8693 & 0.3277 \\

% & Bi-PointFlow\cite{cheng2022bi-flow}  & 0.0650 & 0.7690 & 0.9060 & 0.2640 \\

& WM3DSF\cite{wang2022whatmatters}  & 0.0730 & 0.8190 & 0.8900 & 0.2610 \\

& MSBRN\cite{Cheng_2023_ICCV_MSBRN}  & 0.0448 & 0.8732 & 0.9500 & 0.2085 \\

& SSRFlow (Ours)  & \textbf{0.0298} & \textbf{0.9606} & \textbf{0.9740}  & \textbf{0.1237} \\
\bottomrule
\end{tabular}
\end{minipage}
% \hspace{-1cm}
\vspace{-5pt}
\end{table}

\subsection{Experimental Settings}
\textbf{Implementation Details}
Our model is implemented with PyTorch 1.9, and both training and testing are conducted on a single NVIDIA RTX3090 GPU. The AdamW optimizer~\cite{loshchilov2017adamw} with $ \beta_{1}=0.9 $ and $ \beta_{2}=0.99 $ is used for model tuning during the training phase, with an initial learning rate of $0.001$ which was decayed by half every 80 epochs. We train our model in an end-to-end manner for 900 epochs (or reached convergence) with batch size 8. The cross-attention is utilized with head $=8$ and $d_a=128$. Our model code and weights will be released upon publication. More architectural details are listed in Sec \ref{sup-sec::network} of Appendix.

\textbf{Evaluation Metrics}
Following previous methods~\cite{cheng2022bi-flow,wang2022whatmatters,ding2022fhnet-flow-SFKT,gojcic2021weakly-rigid3dSF,puy2020flot,wu2020pointpwc}, we employ the same evaluation metrics for fair comparisons, including EPE3D, AS3D, AR3D, Out3D, EPE2D, and Acc2D, which are discussed in detail in Sec \ref{sup-sec::metrics} of Appendix. 

\subsection{Results and Analysis}
%It is worth mentioning that while several methods are tailored to specific datasets. However, Our method exhibits remarkable generalization ability across various scenarios regardless of whether it is synthetic or real-world scenes or dense or sparse point clouds.
Our method exhibits remarkable generalization ability across various scenarios, encompassing both synthetic and real-world scenes, as well as dense or sparse point clouds. In contrast, some previous methods are tailored to specific datasets.

\textbf{FT3Ds and KITTIs}
We compare with recent SOTA methods on the FT3Ds and KITTIs datasets. 
% Additionally, we evaluate the generalization of SSRFlow by directly testing it on the KITTIs dataset without any fine-tuning. 
The quantitative results presented in Table~\ref{tab:FTsKTs} indicate that SSRFlow outperforms the other methods by a large margin, especially in real-world datasets.
% and exhibits strong generalization capabilities.
Specifically, on the FT3Ds dataset, SSRFlow is on par with previous SOTA\cite{Cheng_2023_ICCV_MSBRN} while achieving a 63\% reduction in inference time, as listed in Table~\ref{tab:runtime}. Further, our model exhibits exceptional generalization performance on the KITTIs dataset, surpassing the second place by 50\% on EPE3D. Qualitative analysis is shown in \figurename~\ref{fig:FtsKts}.

\def\w{0.32}
\begin{figure}[!htbp]
\centering
\captionsetup[subfloat]{labelsep=none,format=plain,labelformat=empty}
\subfloat{
\includegraphics[width=\w\textwidth, frame]{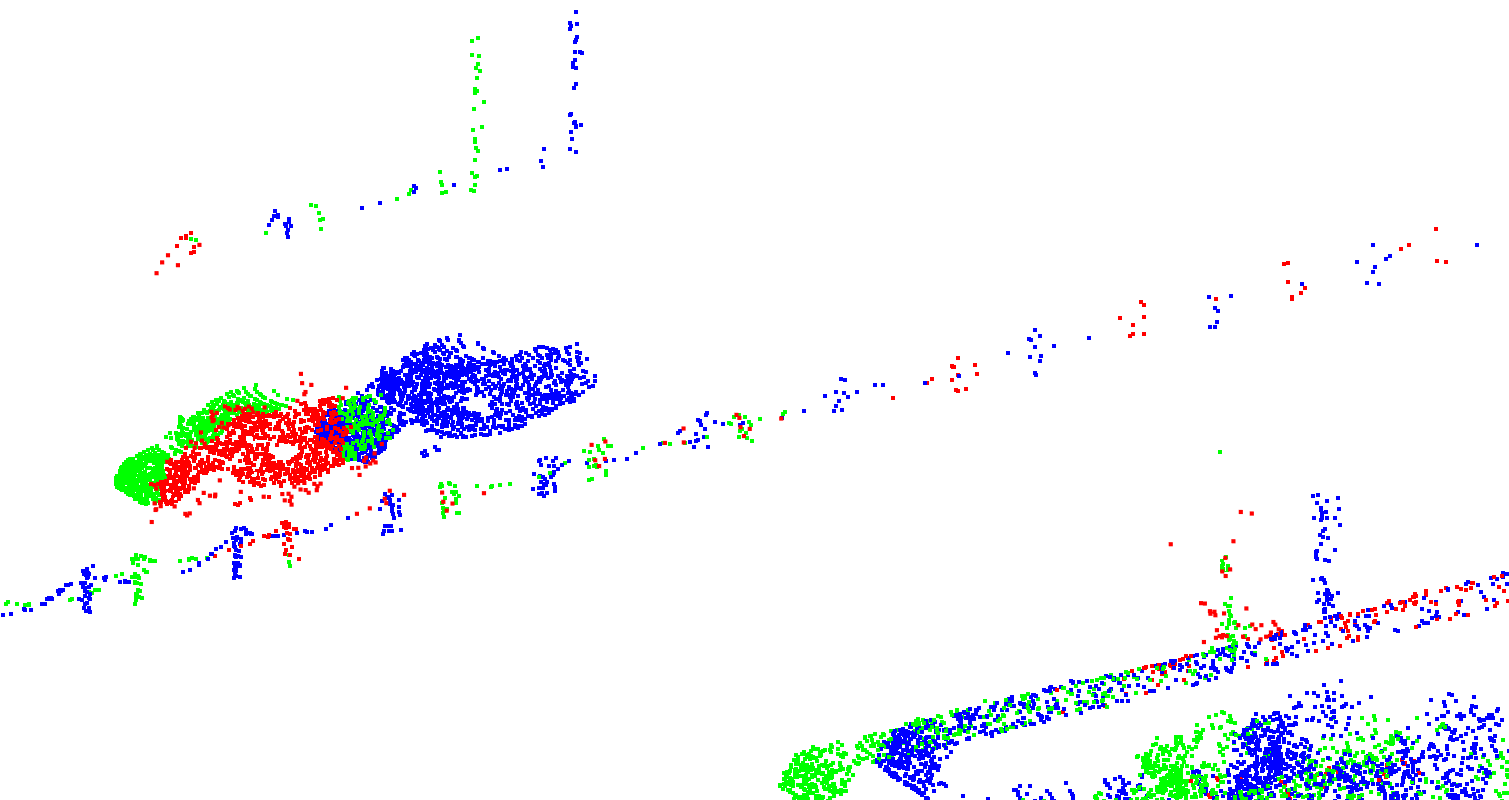}}
\hfil
\subfloat{
\includegraphics[width=\w\textwidth, frame]{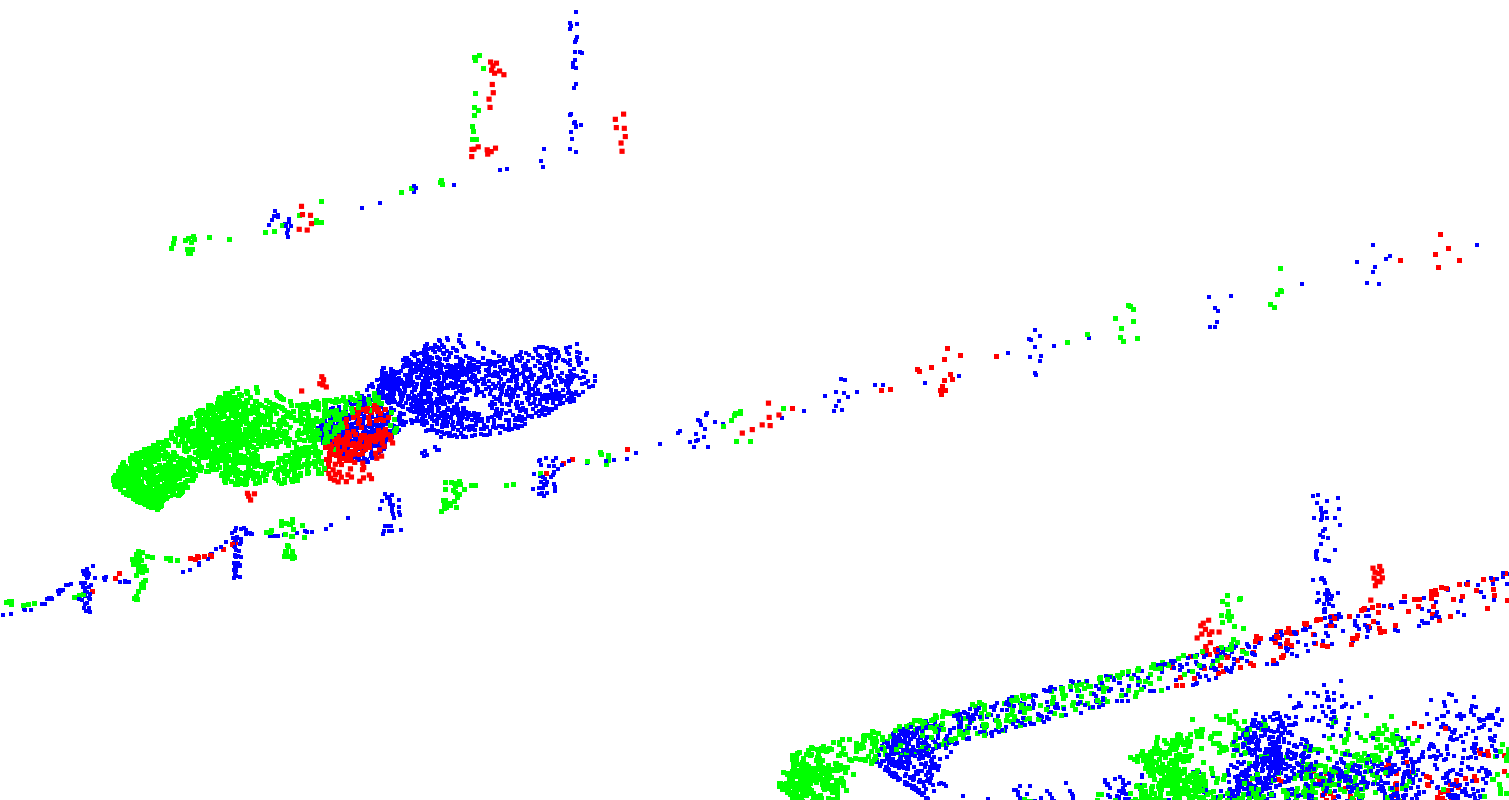}}
\hfil
\subfloat{
\includegraphics[width=\w\textwidth, frame]{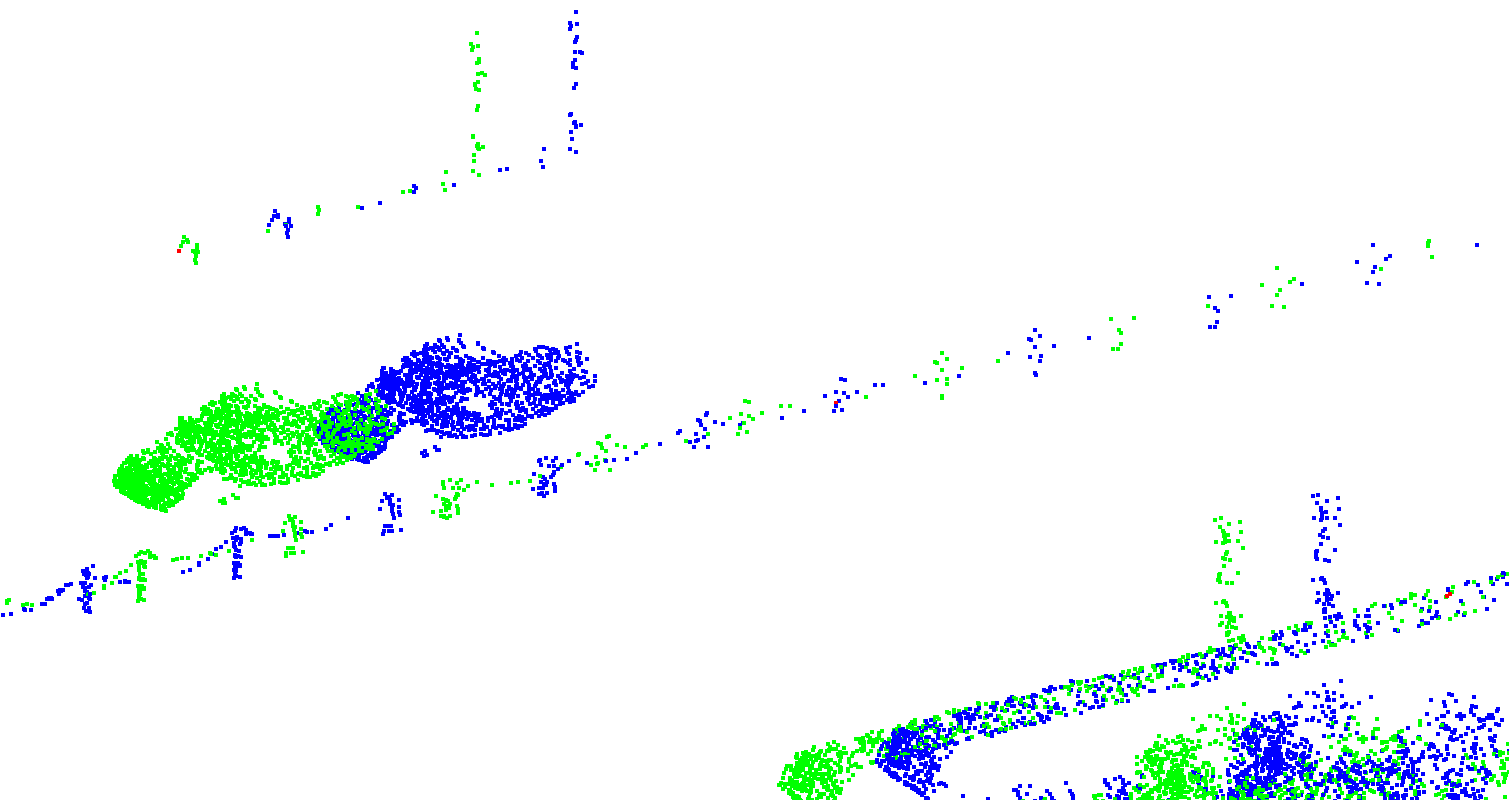}}
\vspace{-8pt}
\quad
\subfloat[PointPWC~\cite{wu2020pointpwc}]{
\includegraphics[width=\w\textwidth, frame]{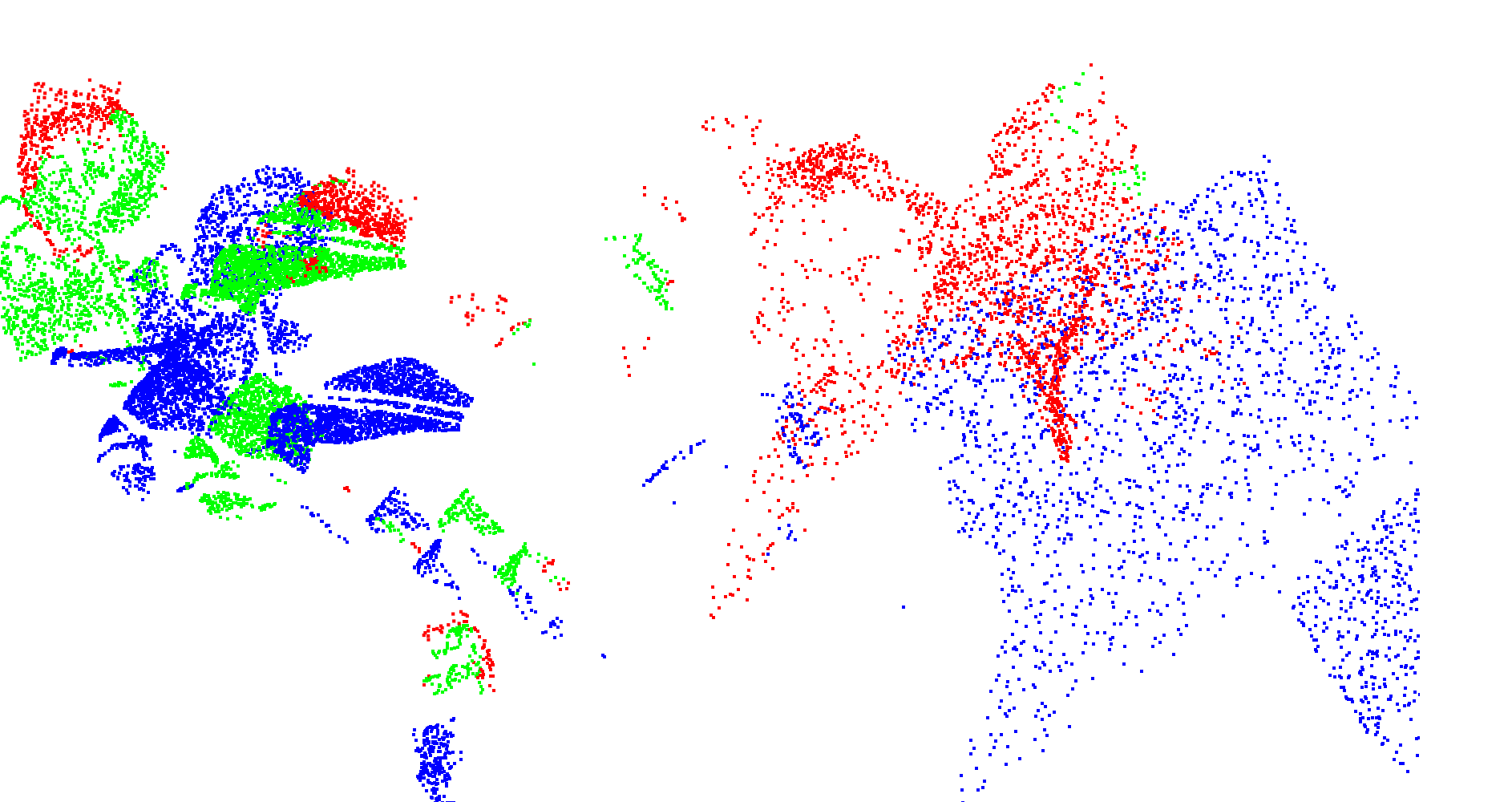}}
\hfil
\subfloat[Bi-PointFlow~\cite{cheng2022bi-flow}]{
\includegraphics[width=\w\textwidth, frame]{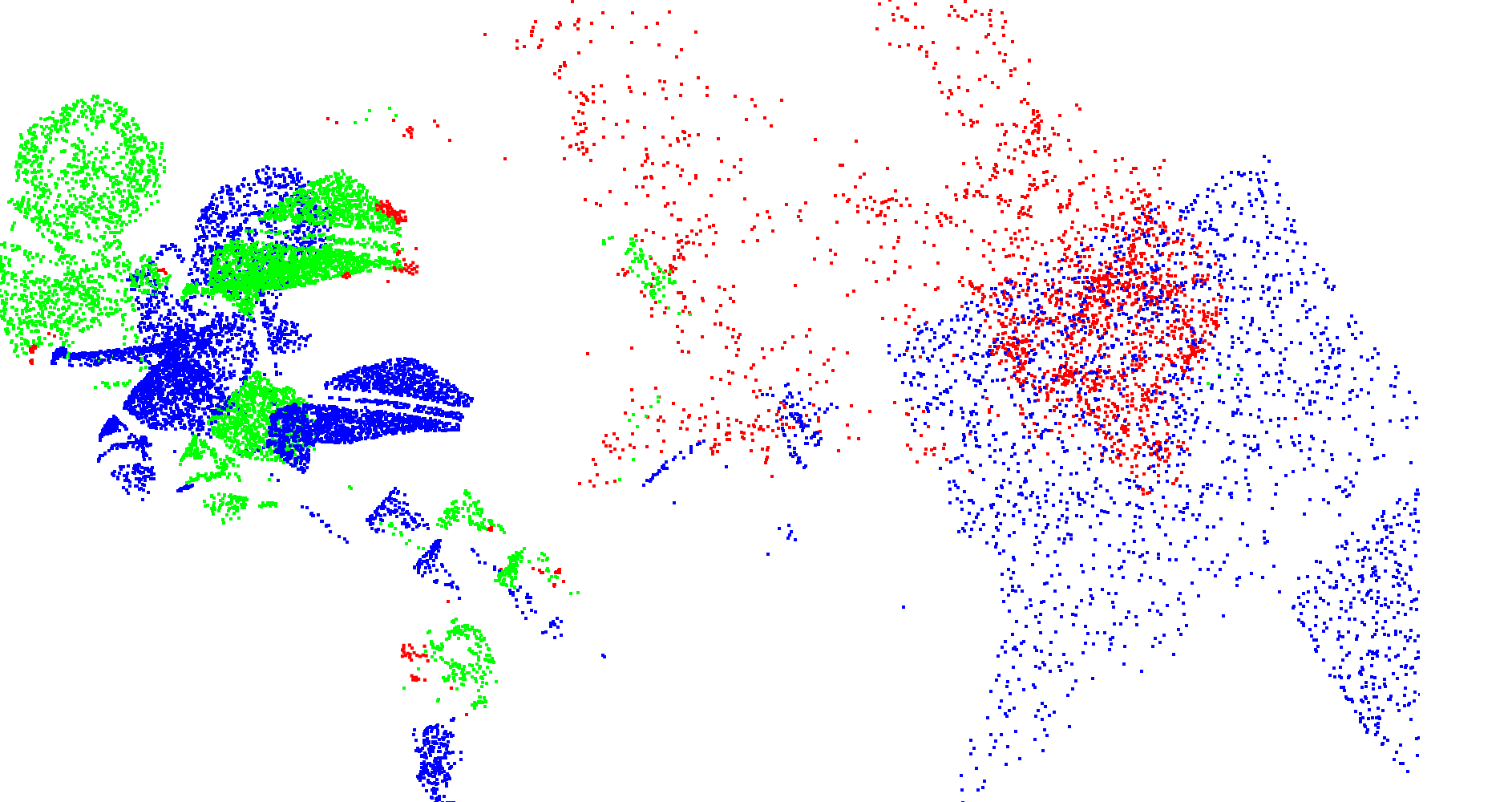}}
\hfil
\subfloat[SSRFlow (Ours)]{
\includegraphics[width=\w\textwidth, frame]{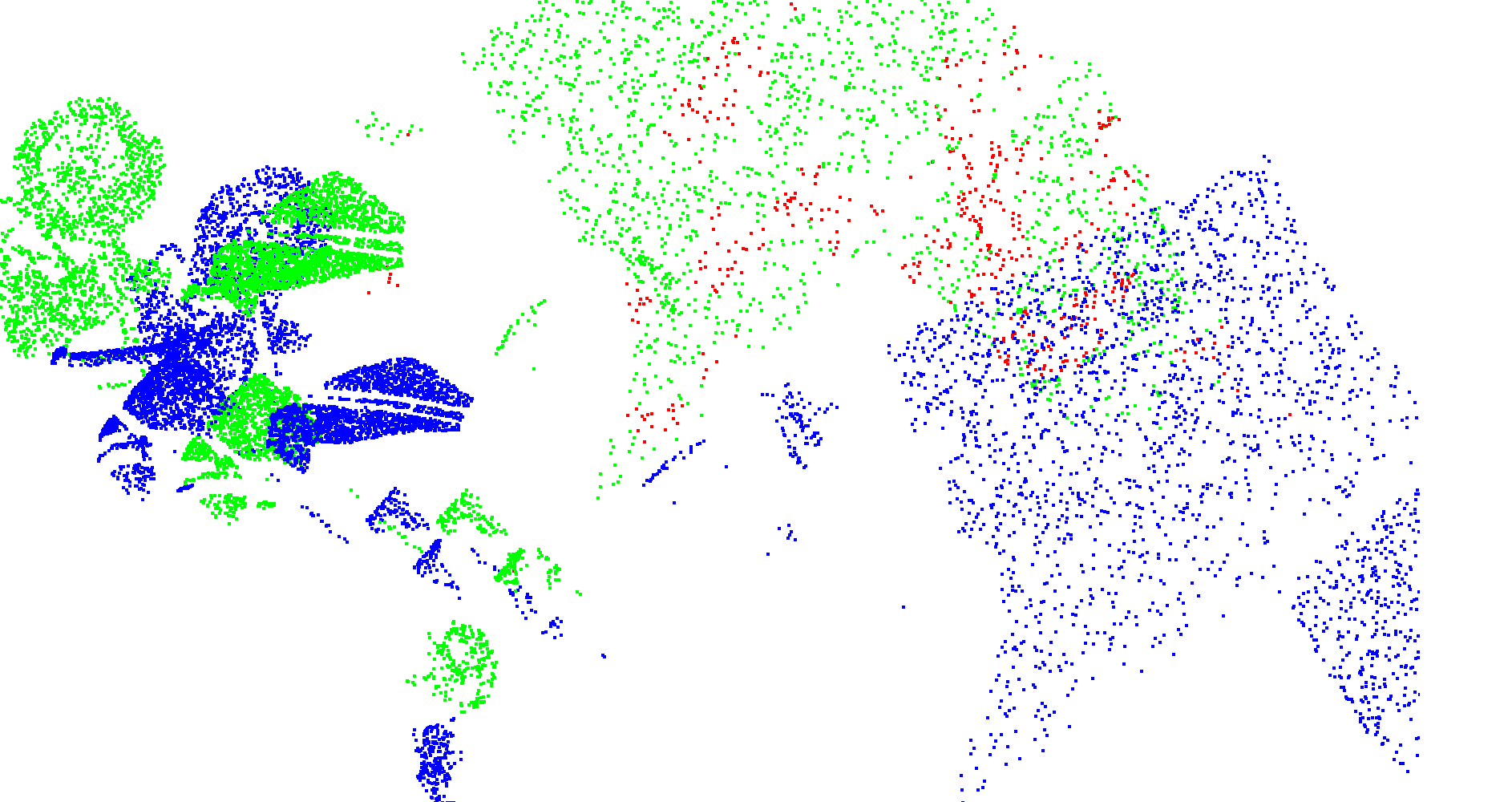}}
\caption{The visualization comparisons on KITTIs (first row) and FT3Ds (second row). The blue represents the source frame, and the green represents the result of warping the source frame using predictions. The red signifies incorrectly predicted warped points whose EPE3D $>$ 0.1m.}
\label{fig:FtsKts}
\vspace{-5pt}
\end{figure}

\def\w{0.243}
\begin{figure}[!htbp]
\centering
\subfloat[FT3Do~\cite{mayer2016large-fly3d}]{
\includegraphics[width=\w\textwidth]{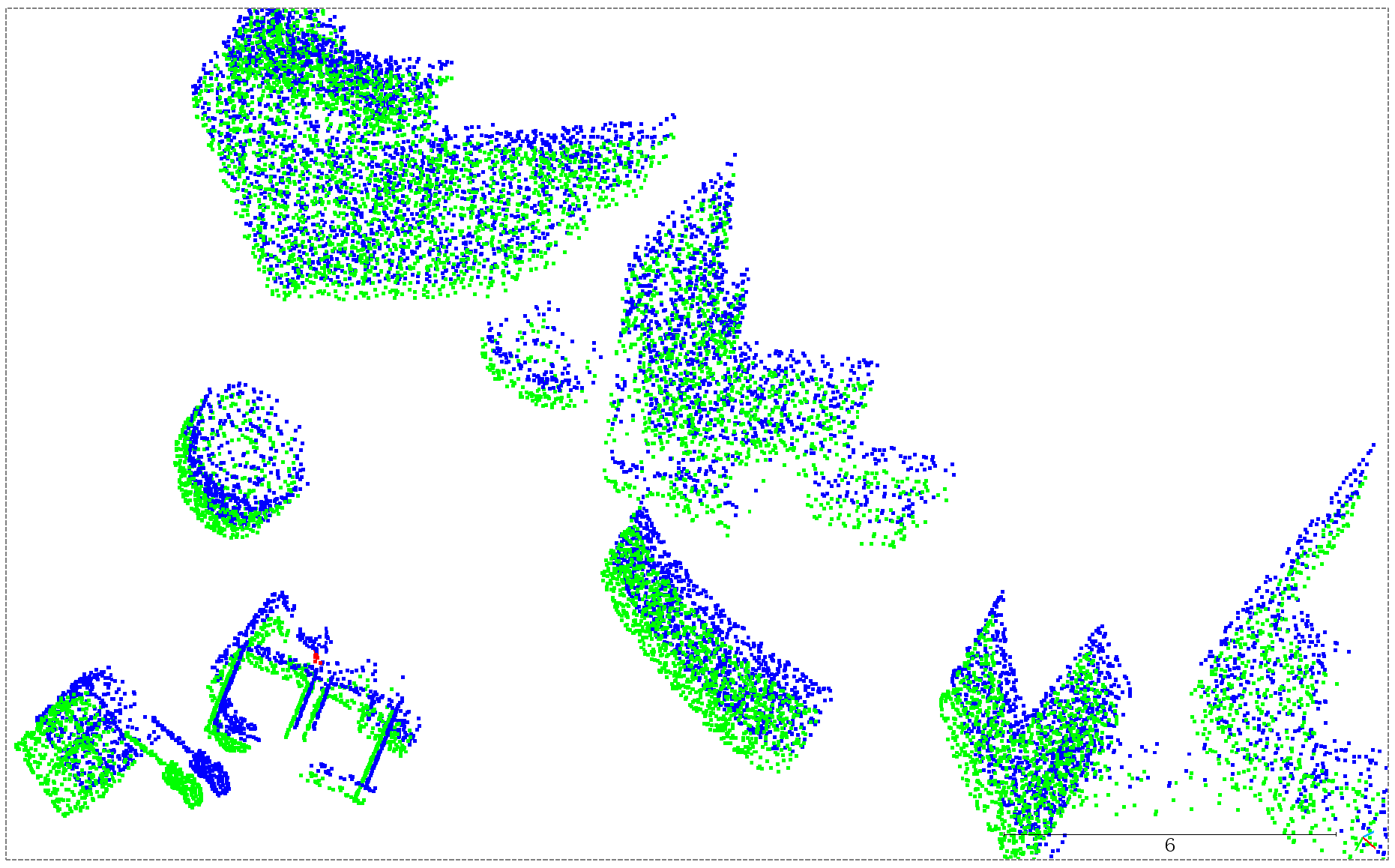}}
\hfil
\subfloat[KITTIo~\cite{cheng2022bi-flow}]{
% \centering\scriptsize{\textbf{Bi-PointFlow~\cite{cheng2022bi-flow}}}
\includegraphics[width=\w\textwidth]{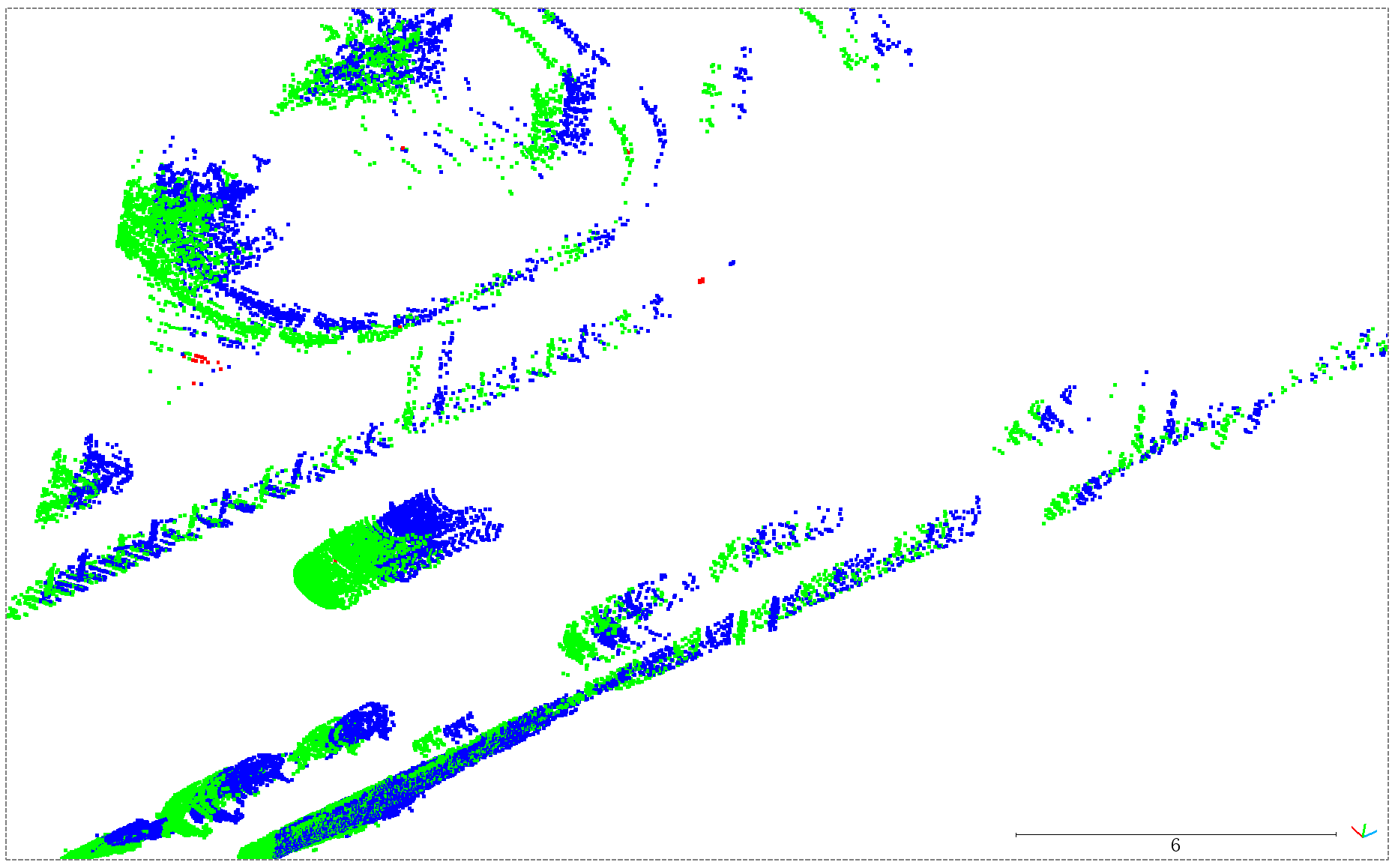}}
\hfil
\subfloat[SF-KITTI~\cite{ding2022fhnet-flow-SFKT}]{
% \centering\scriptsize{\textbf{Bi-PointFlow~\cite{cheng2022bi-flow}}}
\includegraphics[width=\w\textwidth]{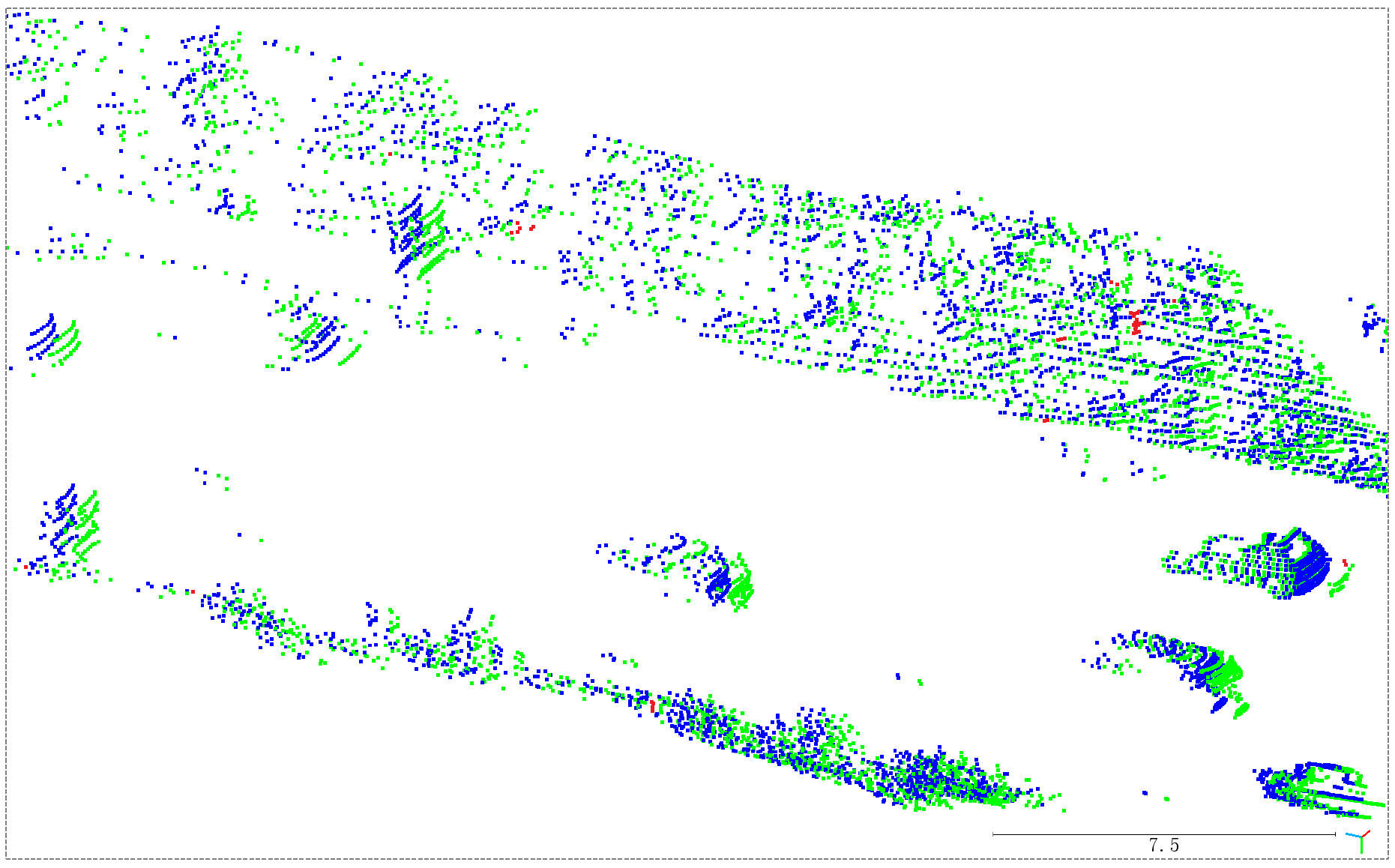}}
\hfil
\subfloat[LiDAR-KITTI~\cite{geiger2012we-lidar-kt}]{
% \centering\scriptsize{\textbf{SSRFlow (Ours)}}
\includegraphics[width=\w\textwidth]{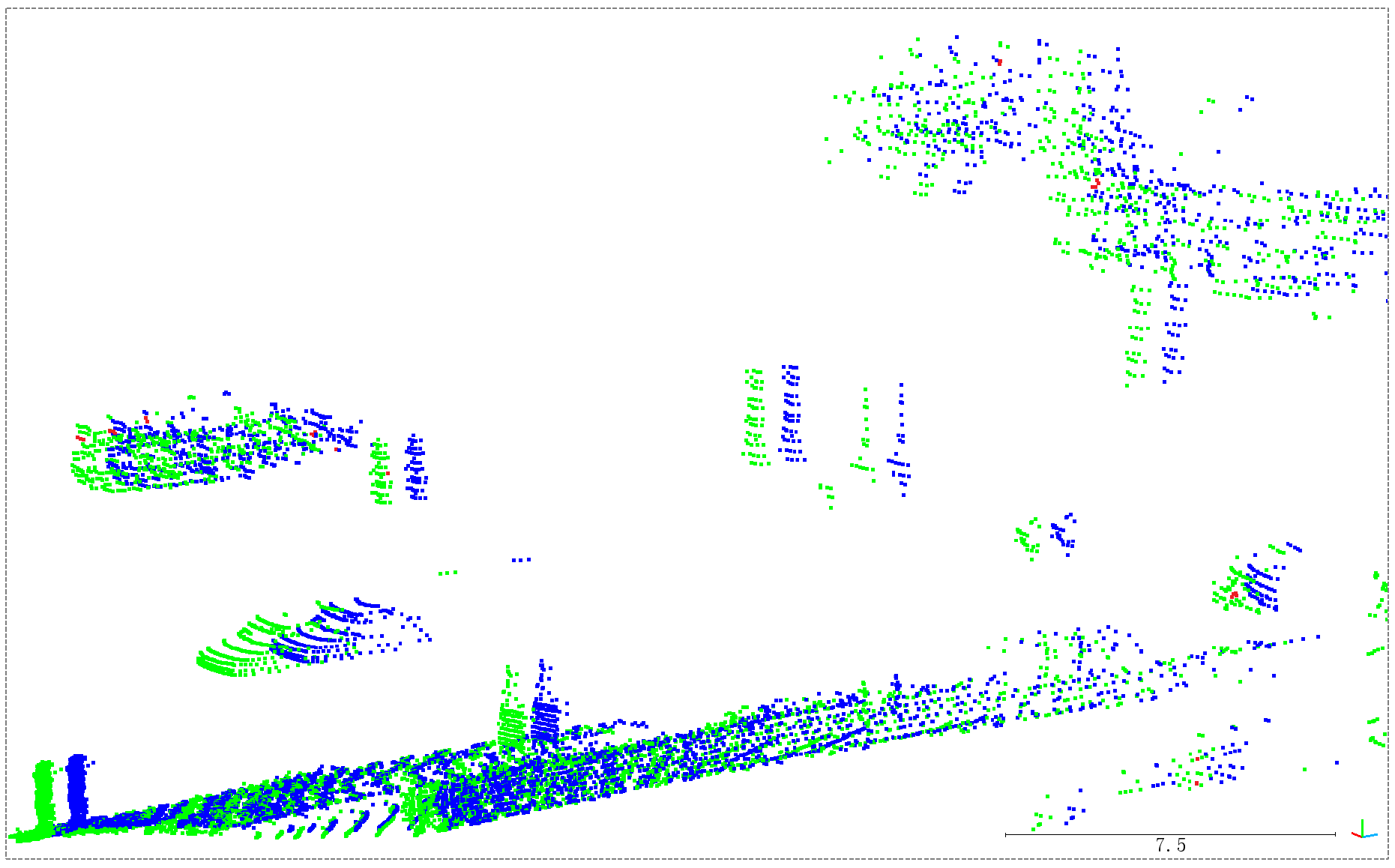}}
\caption{Illustration of results on other datasets of our proposed SSRFlow method. Colors mean the same as Figure \ref{fig:FtsKts}. More visualization results are exhibited in Appendix, Sec \ref{sup-sec::moreResults}}
\label{fig:4datasets}
\vspace{-5pt}
\end{figure}

\begin{table}[!htbp] \footnotesize
\centering
\setlength{\tabcolsep}{4.1pt}
\caption{Evaluation results on LiDAR-KITTI.} \label{tab:lidarkt}
\begin{tabular}{c|ccccc|ccccc}
\toprule
\multirow{2}{*}{Methods} & \multicolumn{4}{c}{FT3Ds$\to$LiDAR-KITTI} & & & \multicolumn{4}{c}{SF-KITTI$\to$LiDAR-KITTI} \\
\cmidrule{2-5}
\cmidrule{8-11}
& EPE3D${\downarrow}$ & AS3D${\uparrow}$ & AR3D${\uparrow}$  & Out3D${\downarrow}$ & & & EPE3D${\downarrow}$ & AS3D${\uparrow}$ & AR3D${\uparrow}$  & Out3D${\downarrow}$ \\
\midrule
FlowNet3D\cite{liu2019flownet3d} & 0.722 & 0.030 & 0.122 & 0.965 & & & 0.289 & 0.107 & 0.334 & 0.749 \\
PointPWC\cite{wu2020pointpwc} & 0.390 & 0.387 & 0.550 & 0.653 & & & 0.275 & 0.151 & 0.405 & 0.737 \\
FLOT\cite{puy2020flot} & 0.653 & 0.155 & 0.313 & 0.837 & & & 0.271 & 0.133 & 0.424 & 0.725 \\
FH-R\cite{ding2022fhnet-flow-SFKT} & 0.472 & 0.369 & 0.432 & 0.805 & & & 0.156 & 0.341 & 0.636 & 0.612 \\
MSBRN\cite{Cheng_2023_ICCV_MSBRN} & 0.351 & 0.400 & 0.592 & 0.685 & & & 0.138 & 0.433 & 0.790 & 0.412 \\
SSRFlow (Ours) & \textbf{0.205} & \textbf{0.498} & \textbf{0.712} & \textbf{0.552} & & & \textbf{0.108} & \textbf{0.570} & \textbf{0.892}  & \textbf{0.401} \\
\bottomrule
\end{tabular}
% \vspace{-5pt}
\end{table}

\begin{table}[t] \footnotesize
\hspace{4pt}
\begin{minipage}[c]{0.4\textwidth}
\centering
\setlength{\tabcolsep}{4pt}
% \vspace{-1.5cm}
\caption{Ablation studies of distinct modules. All module combinations are trained on FT3Ds.}
\label{tab:abAll}
\begin{tabular}{ccc|cc}
\toprule
 STR & GF & DA Loss & \makecell{FT3Ds\\EPE3D${\downarrow}$} & \makecell{KITTIs\\EPE3D${\downarrow}$} \\
\midrule
\ding{52} &  & \ding{52} & 0.0319 & 0.0208 \\
 & \ding{52} & \ding{52} & 0.0301 & 0.0221 \\
 \ding{52} & \ding{52} &  & 0.0183 & 0.0124 \\
 \ding{52} & \ding{52} & \ding{52} & \textbf{0.0122} & \textbf{0.0059} \\
\bottomrule
\end{tabular}   
\end{minipage}
%----------------------subTab-------------------------
\hspace{0.05\textwidth}
\begin{minipage}[c]{0.45\textwidth}
\centering
\setlength{\tabcolsep}{4pt}
% \vspace{-2cm}
\caption{Detailed ablations of the DA Losses. KNN and Radius signify different neighborhood search ways.}
\label{tab:abLoss}
\begin{tabular}{cccc|cc}
\toprule
 $\mathcal{L}_{cfs}$  & $\mathcal{L}_{lfs}$ & KNN & Radius & \makecell{FT3Ds\\EPE3D${\downarrow}$} & \makecell{KITTIs\\EPE3D${\downarrow}$} \\
\midrule
\ding{52} &           & \ding{52}   & \ding{52} & 0.0171 & 0.0109 \\
          & \ding{52} & \ding{52}   & \ding{52} & 0.0169 & 0.0101 \\
\ding{52} & \ding{52} & \ding{52} &           & 0.0136  & 0.0082 \\
\ding{52} & \ding{52} & \ding{52} & \ding{52} & \textbf{0.0122}  & \textbf{0.0059} \\
\bottomrule
\end{tabular} 
\end{minipage}
\vspace{-5pt}
\end{table}

\begin{table}[!htbp] \footnotesize
%----------------------subTab-------------------------
\begin{minipage}[c]{0.45\textwidth}
\centering
\setlength{\tabcolsep}{1pt}
% \vspace{-1cm}
\caption{Detailed ablations on FT3Ds, where r/w A$\to $B denotes replace A with B.}
\label{table: module detail ablation}
\begin{tabular}{l|c}
\toprule
  Method & \makecell{EPE3D${\downarrow}$} \\
\midrule
 Ours (full equip)  & \textbf{0.0122} \\
\midrule
     \textbf{(a) Global Fusion Flow Embedding} & \\
      w/o DCA Fusion  & 0.0259 \\
      r/w attentive weight $\to $ MaxPooling    & 0.0208 \\
      w/ internal position encoder & 0.0162 \\
     % & with external position encoder & 0.0228 \\
      r/w all-to-all $\to $ KNN  & 0.0203 \\
\midrule
    \textbf{(b) Spatial Temporal Re-embedding} & \\
     w/o Spatial Re-embedding  & 0.0171 \\
      w/o Temporal Re-embedding  & 0.0203 \\
      r/w Fusion net $\to $ element-wise addition & 0.0159 \\
\bottomrule
\end{tabular}    
\end{minipage}
\hspace{0.05\textwidth}
\begin{minipage}[c]{0.4\textwidth}
\centering
\setlength{\tabcolsep}{0.5pt}
% \vspace{-1.5cm}
% \vspace{-2cm}
\caption{Transfer results on FT3Ds.}
\label{tab:transfer}
\begin{tabular}{l|c|cc}
\toprule
Model & EPE3D${\downarrow}$ & \makecell{Param\\size (M)} & \makecell{Run\\time (ms)} \\
\midrule
PointPWC        & 0.0588 & 7.72M & 76ms\\
PointPWC+STR    & 0.0504 & 8.02M & 81ms\\
PointPWC+STR+GF & \textbf{0.0402} & 9.89M & 96ms\\
\midrule
Bi-PointflowNet      & 0.0282 & 7.96M & 80ms \\
Bi-PointFlow+GF & 0.0227 & 9.21M & 87ms \\
Bi-PointFlow+GF+STR & \textbf{0.0187} & 10.85M & 103ms \\
\midrule
FlowNet3D    & 0.1136 & 1.23M & 70ms\\
FlowNet3D+GF & \textbf{0.0837} & 1.36M & 94ms\\
\midrule
WM3DSF     & 0.0281 & 4.77M & 63ms \\
WM3DSF+DCA Fusion & \textbf{0.0209} & 5.37M & 72ms \\
\bottomrule
\end{tabular}
\end{minipage}
\vspace{-5pt}
\end{table}

\textbf{FT3Do and KITTIo}
Similar to the above, we train our model on FT3Do and test on KITTIo without any fine-tuning. The experimental results are listed in Table~\ref{tab:FtoKto}, which reveal the good performance of our model even with occlusion.
% which demonstrates that our model has strong generalization performance and excels in generating accurate motion flow estimation, even in the presence of occlusion.
Specifically, our model achieves 39\% improvement over the previous SOTA method~\cite{Cheng_2023_ICCV_MSBRN} on FT3Do. Furthermore, 
% the exceptional generalization capabilities of SSRFlow enable it to deliver comprehensive results on real-world occluded KITTIo datasets. In fact,
SSRFlow outperforms \cite{Cheng_2023_ICCV_MSBRN} by a significant margin, surpassing it by 33\% and 40\% in terms of EPE3D and Out3D on the real-world occluded KITTIo dataset, respectively. Visualized experimental results are provided in the \figurename~\ref{fig:4datasets}.

\textbf{Generalization on LiDAR-KITTI}
To validate the generalization on real-world LiDAR-scanned datasets, we train our model on FT3Ds and SF-KITTI datasets separately, followed by evaluation on the LiDAR-KITTI dataset. The results are shown in Table~\ref{tab:lidarkt} and \figurename~\ref{fig:4datasets}. Specifically, SSRFlow reduces EPE3D by 41\% and 22\% compared to the second place~\cite{Cheng_2023_ICCV_MSBRN} under training on FT3Ds and SF-KITTI datasets, respectively.
% which proves that our model also has strong motion inference ability even on the sparse and indirect corresponding LiDAR-scanned point clouds.

\subsection{Ablation Study}
To investigate the distinct impacts of GF, STR, and DA Losses, a set of ablation experiments are conducted to perform functional analysis. The comprehensive results of the ablation experiments can be found in Table~\ref{tab:abAll}, while detailed information is presented in Table~\ref{tab:abLoss} and Table~\ref{table: module detail ablation}.

%\textbf{Global Fusion Flow Embedding (GF)}
\textbf{GF}
% Unlike previous methods~\cite{kittenplon2021flowstep3d,wang2022whatmatters,puy2020flot}, the GF module globally fuses the semantic context of two frames by DCA Fusion and leverages cross attentive map to obtain match associations across all point-pairs in both the latent and Euclidean space. 
In~(a) of Table~\ref{table: module detail ablation}, we provide a detailed list of the importance of the DCA Fusion, location of position encoder, aggregation style, and all-to-all point-pair concatenation. 
Firstly, we exclude the DCA Fusion and the subsequent global attentive aggregation in GF and directly utilize the original semantic feature for global flow embedding. Secondly, we test the internal and external position encoder of cross-attention in DCA Fusion. Additionally, we replace the attentive weighted aggregation with MaxPooling. Finally, we substitute the all-to-all match method with KNN. 
After removing the DCA Fusion, the model experienced a substantial decline in accuracy, primarily due to its capability to fuse point features with another frame context before embedding. Moreover, the position encoder outside the DCA Fusion provides additional spatial features that are superior to internal equipment. The KNN method struggles to process long-range distance dependencies.

\textbf{STR}
% The STR module effectively re-embeds the local spatiotemporal features of each point in the source frame after warping, resulting in a significant improvement of flow prediction, as shown in Table \ref{tab:abAll}. 
We remove the Spatial and Temporal Re-embedding sub-modules separately to consider their contribution to the STR module. The detailed results are listed in (b) of Table \ref{table: module detail ablation}. 
It is observed that the Spatial Re-embedding sub-module has brought greater performance improvement, which is in line with common sense, as the relation between the two frames has been taken into account in the subsequent cost volume calculations. 

\textbf{DA Losses}
We conduct a series of ablation experiments exploring the effectiveness of the LFC loss and the CFS loss, as well as neighborhood search strategies. The results are listed in Table~\ref{tab:abLoss}. Detailed analysis of the hyper-parameters is in Sec \ref{sup-sec::daloss} of Appendix.

\textbf{Transfer to Other Models}
To evaluate the effectiveness of the proposed GF and STR modules, we conduct experiments by integrating them directly into PointPWC~\cite{wu2020pointpwc},  FlowNet3D~\cite{liu2019flownet3d}, Bi-PointFlow\cite{cheng2022bi-flow} and WM3DSF\cite{wang2022whatmatters}.  Following the original training strategy as described in the respective papers, the results are listed in Table~\ref{tab:transfer}. Both modules improve network performance.

\section{Conclusion}
We propose SSRFlow network to accurately and robustly estimate scene flow. SSRFlow conducts global semantic feature fusion and attentive flow embedding in both Euclidean and context spaces. Additionally, it effectively re-embeds deformed spatiotemporal features within local refinement. The DA Losses enhance the generalization ability of SSRFlow on various pattern datasets. Experiments show that our method achieves SOTA performance on multiple distinct datasets.

{
\small
\bibliographystyle{apalike}
% \bibliographystyle{IEEEtran}
% \bibliography{Ref}
\bibliography{ref.bib}
}

% \newpage
\appendix
\section{Network}

\subsection{Network Architecture}
\label{sup-sec::network}
Our proposed network architecture mainly consists of three components: 
(1)	Hierarchical point cloud feature extraction.
(2)	Global attentive flow initialization.
(3)	Local flow refinement.
%这一段与framework的caption高度重合，是否可以放在补充材料中去%
To hierarchically extract semantic features from point clouds, we begin by adopting a pyramid feature extraction network. Subsequently, we construct an attentive global flow embedding that accounts for both high-dimensional feature space and Euclidean space. The resulting sparse flow is then upsampled using upsampling and warping layers, enabling denser flows at lower levels. These flows are accumulated onto the source frame, ultimately yielding the warped source frame. Thereafter, spatiotemporal re-embedding features are used for hierarchical refinement of the residual scene flow between the warped source frame and the target frame. Through iterative refinement, full-resolution scene flow is generated as the ultimate result.

% 正文转移
The architecture of our model comprises a total of 5 levels. We take $N_1=M_1=8192$ points as input, randomly sampled from source and target frames if they contain different numbers of points. Within the multi-layer pyramid network, the down-sampled points vary across different levels, with $N_{2}=2048$, $N_{3}=512$, $N_{4}=256$, and $N_{5}=64$, respectively. Notably, the $5th$ layer is referred to as the GF module. 

\subsection{DCA Fusion}
\label{sup-sec::DCA Fusion}

\begin{figure}[h]
\centering
\includegraphics[width=1.0\linewidth]{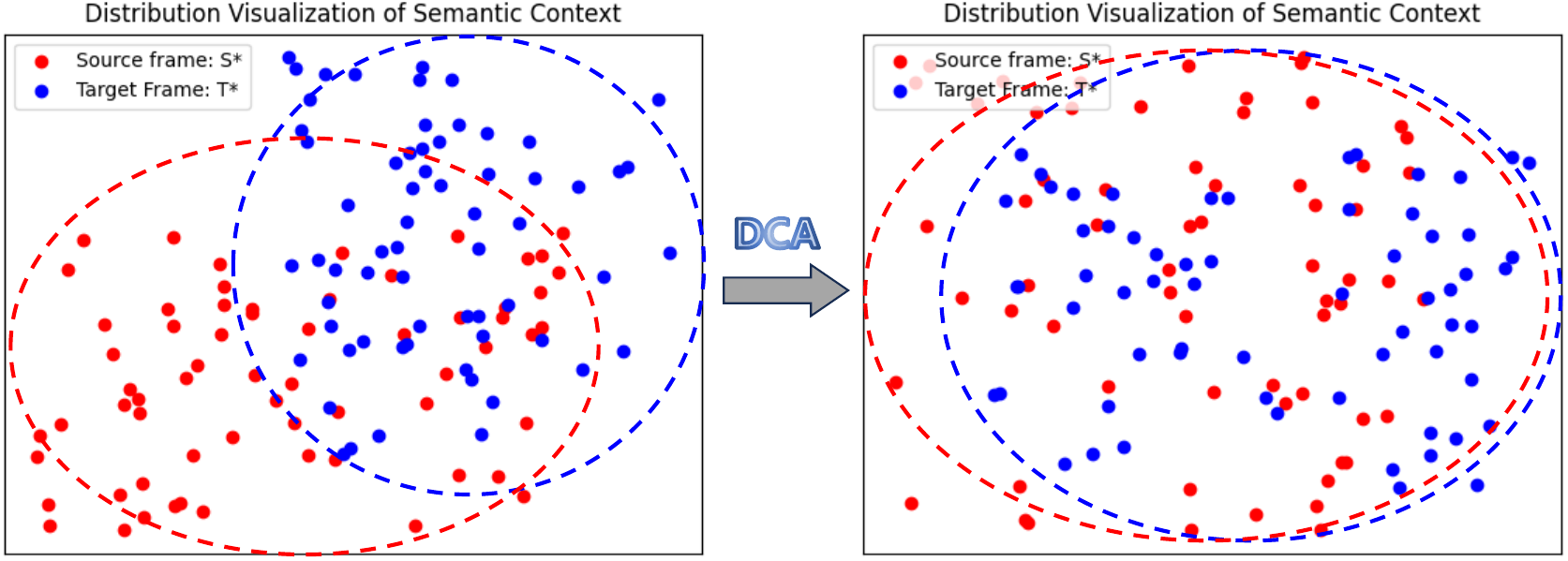}
\caption{Visualization of the distribution of the highest level of point cloud feature pyramid before and after employing DCA. It is evident that DCA Fusion enhances the semantic alignment of the two consecutive point clouds, thereby facilitating subsequent global flow embedding.}
\label{fig::distVis}
\end{figure}

\begin{figure}[t]
\centering
\includegraphics[width=1.0\linewidth]{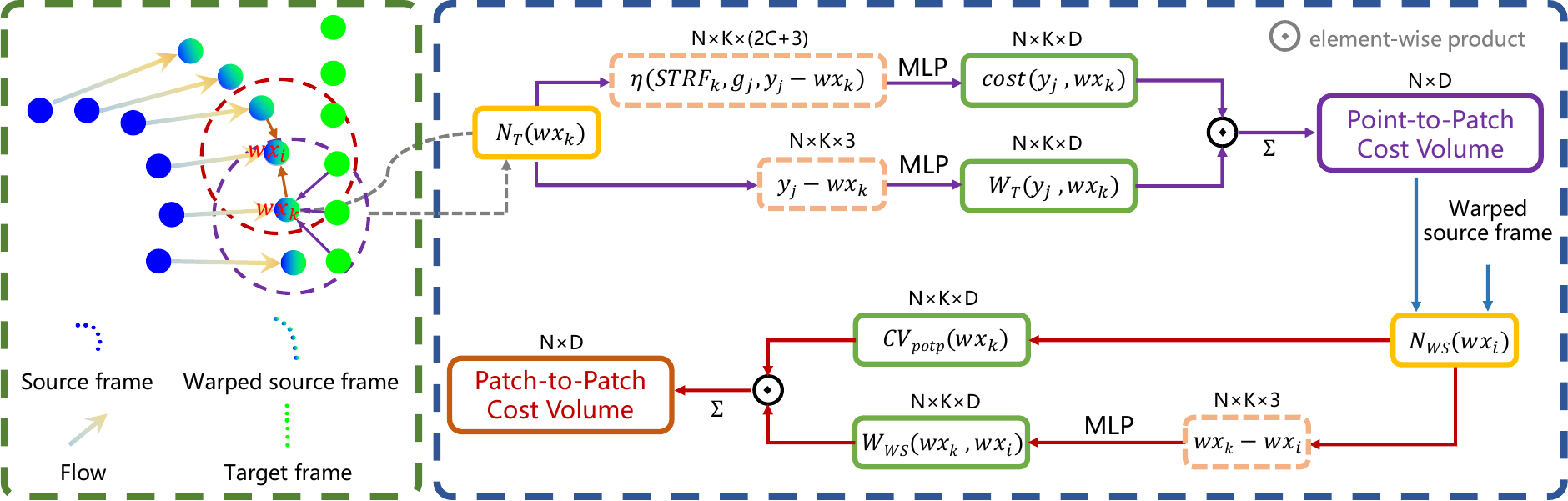}
\caption{
The structure of the Local Flow Embedding (LFE) module. We first generate $\mathcal{N}_{WS}(wx)$ and $\mathcal{N}_{T}(wx)$ using KNN for each warped source point $wx$. Then, we input each point and its neighbors from frames $WS$ and $T$ along with their features into the LFE module.The spatiotemporal re-embedded feature of the warped point $wx_k$ is combined with the features of its neighboring points in $\mathcal{N}_{T}(wx_k)$, along with the direction vector $y_j-wx_k$. Subsequently, they are processed through an MLP to calculate a point-to-patch cost between $wx_k$ and the target frame $T$. Following this, each point-to-patch cost in $\mathcal{N}_{WS}(wx_i)$ is additionally self-aggregated using an MLP to construct a patch-to-patch cost volume.}
\label{fig:lfe}
\end{figure}

\subsection{Local Flow Embedding}
In this section, we input the warped source frame and target frame into the Local Flow Embedding (LFE) module to generate a local flow embedding, which is realized following the patch-to-patch approach~\cite{wu2020pointpwc}. 

The matching cost between two frames of point clouds can be defined as follows:
\begin{equation}
    cost\left(y_{j}, wx_{k}\right)=\text{MLP}(\eta(STRF_k,g_j,y_j-wx_k)),
\end{equation}
where $\eta$ stands for concatenation. $STRF_k$ is the feature of warped source point $wx_k$, and $g_j$ is the feature of target point $y_j$. Afterward, this matching cost can be aggregated into a point-to-patch cost volume between the two frames. 
\begin{equation}
   CV_{point}(wx_{k})=\sum_{ \mathclap{ y_{j} \in \mathcal{N}_{T}\left(wx_{k}\right)} }\text{MLP}(y_j-wx_k) cost\left(y_{j}, wx_{k}\right).
\end{equation}
Finally, the patch-to-patch cost volume for warped source point $wx_i$ can be defined as:
\begin{equation}
CV_{patch}\left(wx_{i}\right)=\sum_{ \mathclap{wx_{k} \in \mathcal{N}_{WS}\left(wx_{i}\right)} } \text{MLP}(wx_k-wx_i)CV_{point}(wx_{k}).
    \label{eq: CV}
\end{equation}
Here $\text{MLP}(wx_k-wx_i)$ and $\text{MLP}(y_j-wx_k)$ are the aggregation weights determined by the direction vectors. Follow the notation in the main text, we use $\mathcal{N}_{WS}(wx_i)$ to define the neighbors of $wx_i$ in frame $WS$ and $\mathcal{N}_T(wx_k)$ to define the neighbors of $wx_k$ in frame $T$. The patch-to-patch manner employed in the cost volume is illustrated in Figure \ref{fig:lfe}.

\section{Experiments Settings}
\subsection{Evaluation Metrics}
\label{sup-sec::metrics}
For fair comparisons, we leverage the same set of evaluation metrics as in the previous methods~\cite{liu2019flownet3d,wu2020pointpwc,cheng2022bi-flow,wang2022whatmatters}.

\begin{itemize}
    \item EPE3D: the main evaluation measuring average 3D end-point-error which is formulated as   $\text{EPE3D} = \frac{1}{N}  { \sum_{i}^{}}  \| \tilde{sf_{i}}-sf_{i}  \| _{2}$
    \item AS3D: the strict version for scene flow accuracy which denotes the percentage of points whose EPE3D $<$ 0.05m or relative error $<$ 5\%.
    \item AR3D: the relax version for scene flow accuracy which denotes the percentage of points whose EPE3D $<$ 0.1m or relative error $<$ 10\%.
    \item Out3D: the proportion of points that exhibit significant miscalculations, characterized by an EPE3D $>$ 0.3m or relative error $>$ 30\%.
    \item EPE2D: the main measurement for 2D optical flow which represents the end-point-error by projecting points black to the 2D image plane.
    \item Acc2D: the percentage of points whose EPE2D $<$ 3px or relative error $<$ 5\%.
\end{itemize}

\begin{figure}[t]
\centering
\subfloat[KNN]{\includegraphics[width=0.46\linewidth]{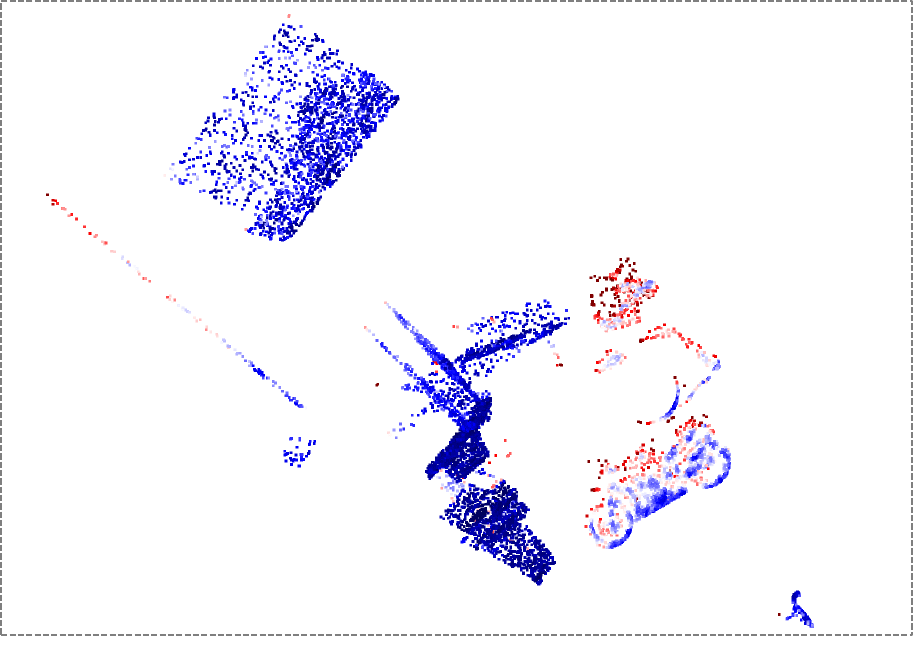}}
\hfil
\subfloat[KNN+Radius]{\includegraphics[width=0.46\linewidth]{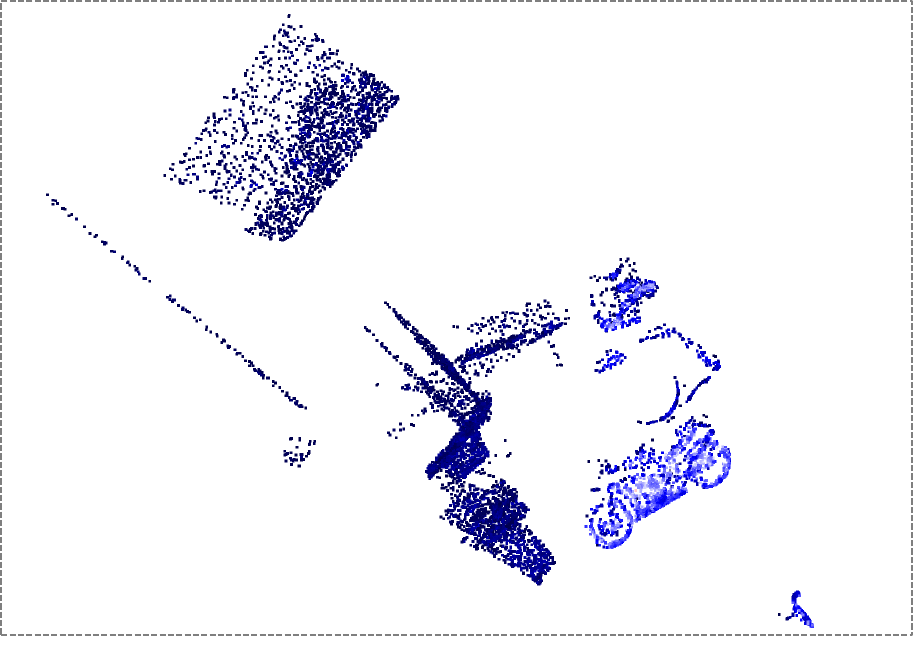}}
\hfil
\subfloat{\includegraphics[width=0.05\linewidth]{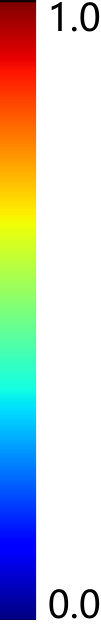}}
\hfil
\caption{The differences of ground truth scene flow local consistency under different nearest point search methods. The normalized value is shown in the right color bar.}
\label{fig:knn_radius}
\end{figure}

\subsection{Search Methods}
\label{sup-sec::searchMethods}
To assess the enhancement of the KNN+Radius approach compared to the pure KNN method, we visualize the flow disparities of different points under the search strategies of KNN and KNN+Radius local point clusters based on the ground truth of the scene flow.

From \figurename~\ref{fig:knn_radius} it can be observed that using only KNN introduces noise points that do not belong to the block, thereby causing greater local flow disparities and leading to deviating in the network model learning.

\def\w{0.45}
\begin{figure}[t]
\centering
\subfloat[The visualization of a KITTIs dataset scene in which there are rare background points observed.]{\includegraphics[width=0.45\linewidth]{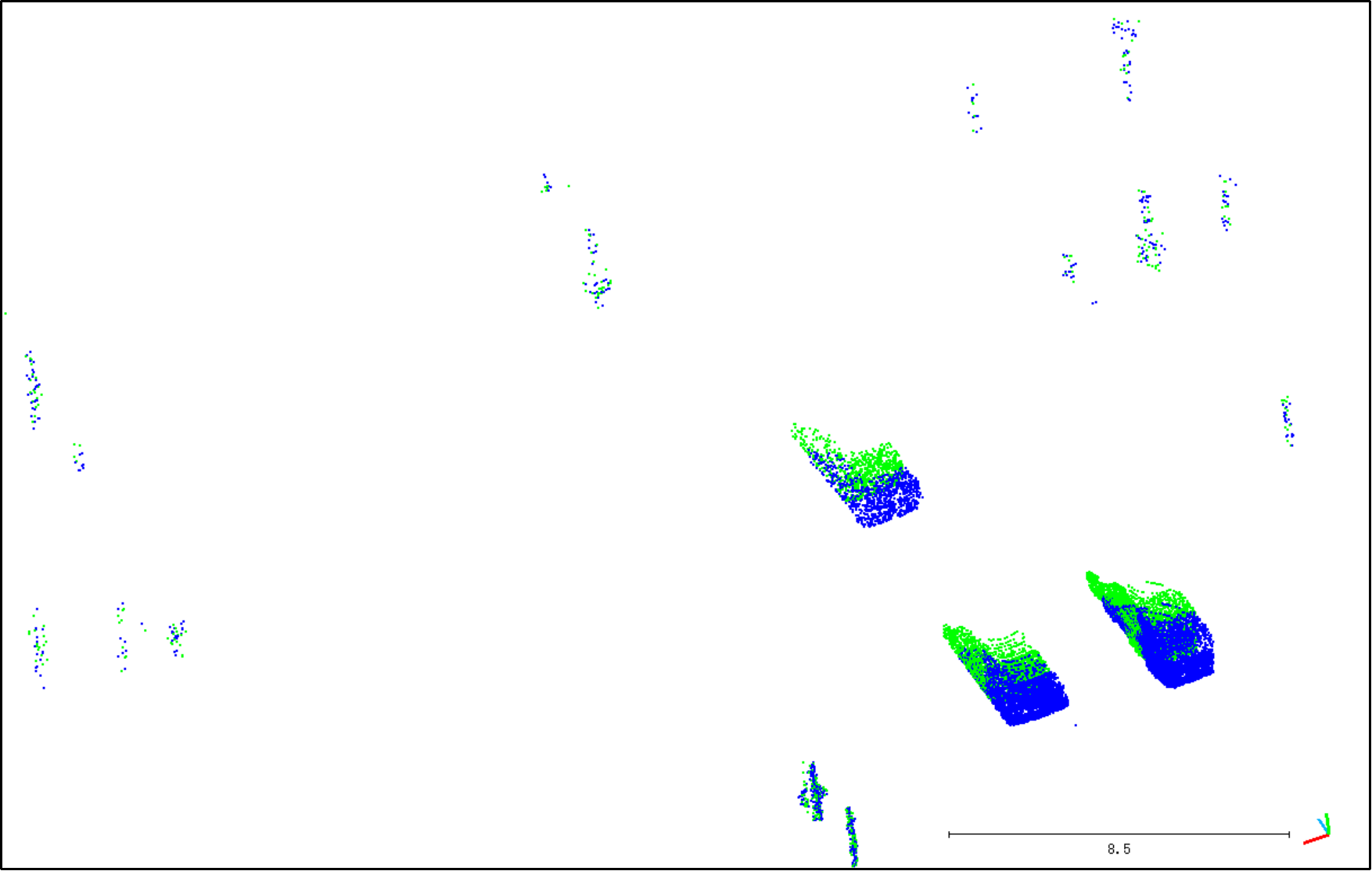}}\label{supFig:scene}
\hfil
\subfloat[A color level plot to visualize the average local differences in GT flow for different combinations of $ R \text{ and } K $.]{
\includegraphics[width=\w\textwidth, frame]{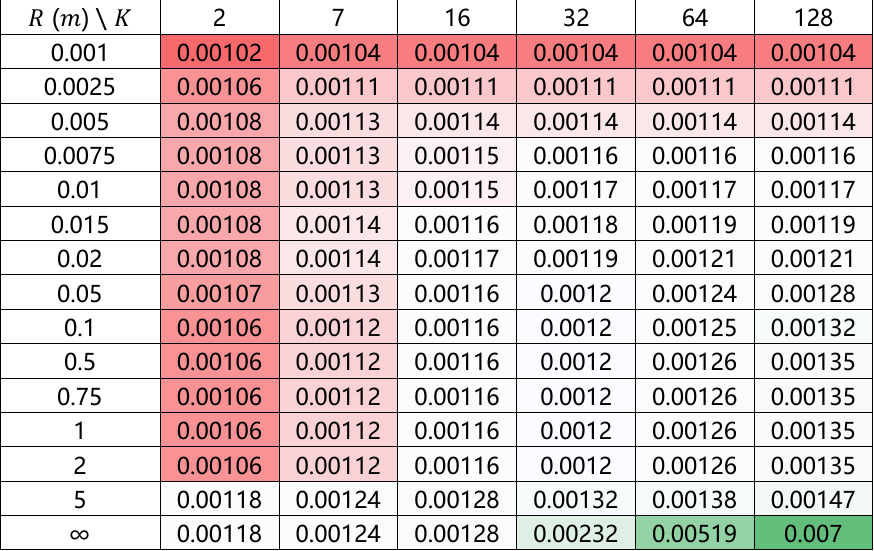}}\label{supFig:flowDif}
\quad
\subfloat[As the value of $R$ progressively increases, a color level plot illustrates the proportion of points accessible through KNN + Radius compared to the original KNN. As the radius approaches infinity, the original KNN behavior is reinstated.]{
\includegraphics[width=\w\textwidth, frame]{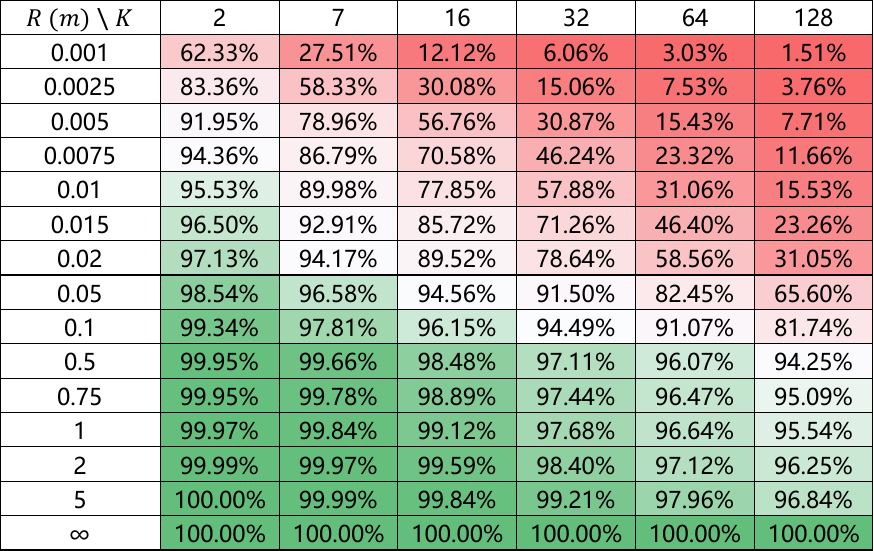}}\label{supFig:validNum}
\hfil
\subfloat[The EPE3D of KITTIs with increasing the number of KNN searches on a reasonable local rigid radius ($R = 0.05$ m). Notably, our model is tested on the KITTIs dataset without any fine-tuning.]{
\includegraphics[width=\w\textwidth, frame]{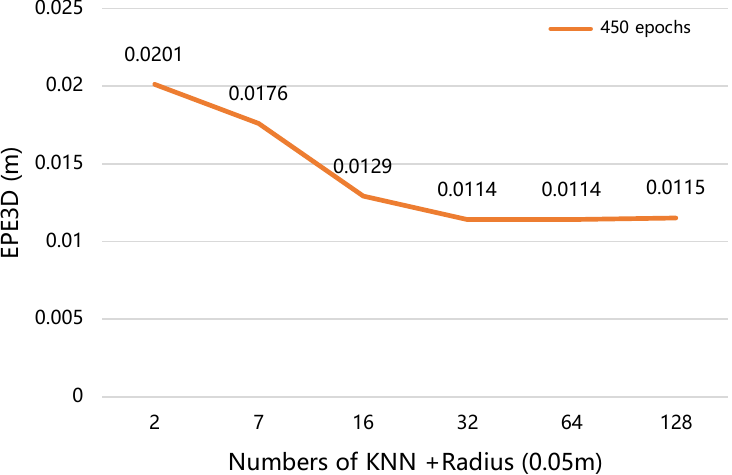}}\label{supFig:epeRK}
\caption{Ablation studies and analysis of adaption losses.}
\label{fig:abLossesAll}
\end{figure}

\subsection{Hyper-parameters in DA Losses}
\label{sup-sec::daloss}
We conduct a comprehensive experimental analysis on the hyper-parameters $ K, R, \text{ and } TH $ in the context of DA Losses, where $ R $ represents the radius truncation of neighbor search, $ K $ represents the number of neighbors in KNN, and $ TH $ represents the cross-frame feature similarity threshold. It is worth mentioning that $ R \text{ and } K $ are uniform across both Local Flow Consistency (LCF) and Cross-frame Feature Similarity (CFS) loss functions since they are utilized to identify local rigid blocks with similar semantic features.

To analyze the local consistency of ground truth (GT) flow, we initiate the process by selecting a suitable scenario in the KITTIs dataset that exhibits minimal visible background points, as depicted in Figure~\ref{fig:abLossesAll}. Subsequently, we systematically augment the ranges of $R$ and $K$
%both the number of KNN and the truncation radius range
to acquire diverse consistency outcomes. As shown in ~\figurename~\ref{fig:abLossesAll} and~\figurename\ref{fig:abLossesAll}, the quantity of KNN group points exhibits a gradual decrease as the truncation radius decreases. Simultaneously, the average flow differences within each group also decrease, indicating that the KNN+Radius search method effectively eliminates the noise points.

We further explore the disparities between KNN and KNN+Radius search methods. The KNN+Radius search method results in certain points without any neighboring points. These isolated points, known as exceptionally sparse or invalid points, function as noise points that impede flow smoothing. However, utilizing the pure KNN search method would optimize flow consistency between noise points and mandatory surrounding points, which is an incorrect outcome.
By observing the size of local rigid blocks in real-world scenarios, we finally select $ R=0.05 $m as the truncation radius to perform ablation experiments on different KNN search numbers. The results are shown in Figure ~\ref{fig:abLossesAll}.

\begin{table}[ht]  \footnotesize
    \centering
    \caption{The EPE3D of KITTIs with different cross-frame feature similarity threshold.}
    \label{tab:TH}
    \begin{tabular}{c|c|c|c|c|c|c}
    \toprule
      {Threshold $TH$} & 0.99 & 0.95 & 0.9 & 0.8 & 0.7 & 0.6 \\
       \midrule
      {EPE3D} &0.0149 & 0.0114 & 0.0122 & 0.0138 & 0.0151  & 0.0180 \\
    \bottomrule
    \end{tabular}
\end{table}

Regarding the selection of hyper-parameter $TH$ in the CFS loss function, we fix $ K=32 \text{ and } R=0.05$m that represent a local rigid block for capturing cross-frame similar features and test different $TH$ values in the ablation experiments. The corresponding results are listed in Table~\ref{tab:TH}. Our model was trained for a limited number of \textbf{450 epochs} on the FT3Ds dataset and subsequently evaluated on the KITTIs dataset without undergoing any fine-tuning, which is enough for observing the trend.

\def\w{0.32}
\begin{figure}[ht] \footnotesize
\centering
\subfloat[]{
\includegraphics[width=\w\textwidth]{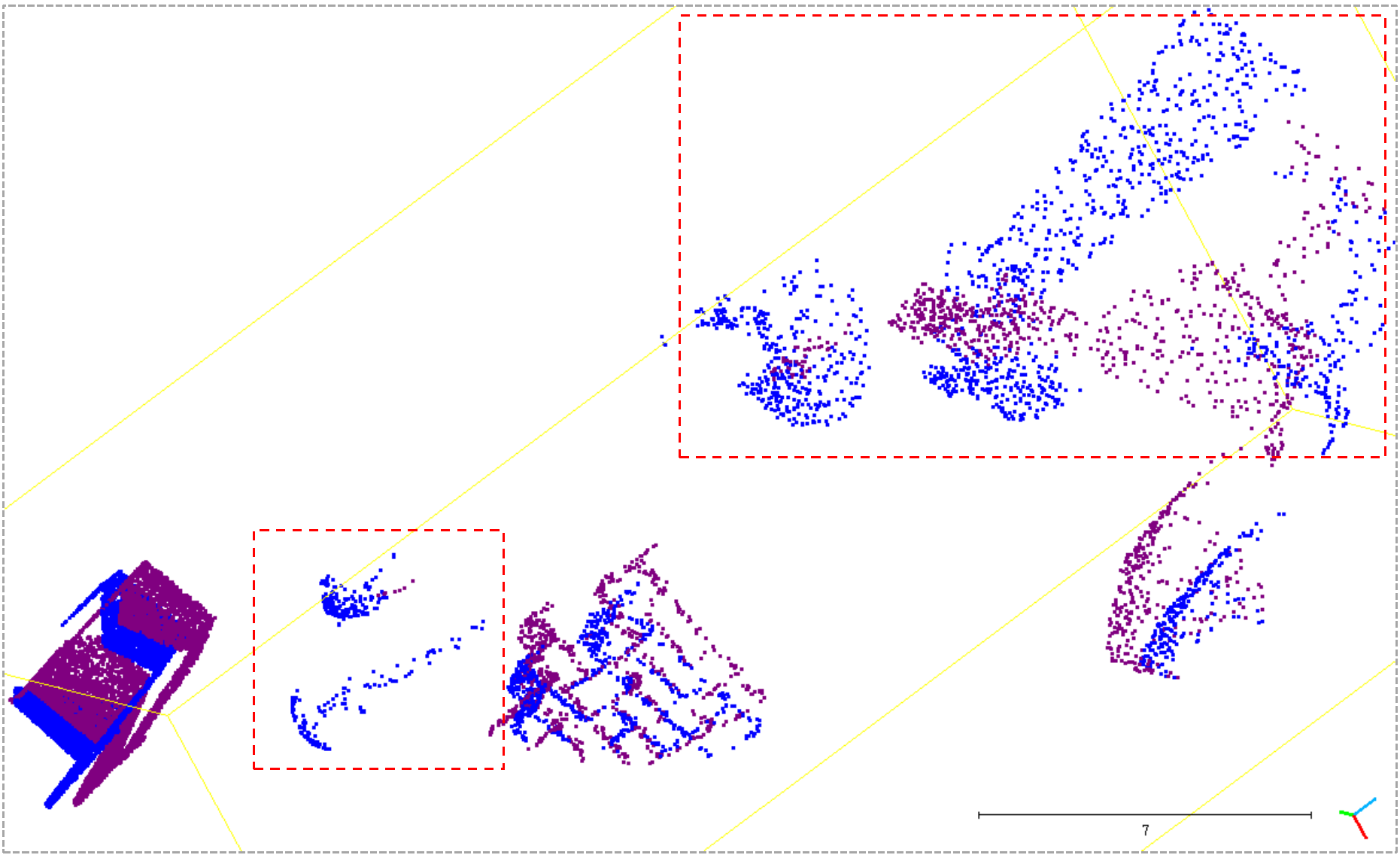}}
\hfil
\subfloat[]{
\includegraphics[width=\w\textwidth]{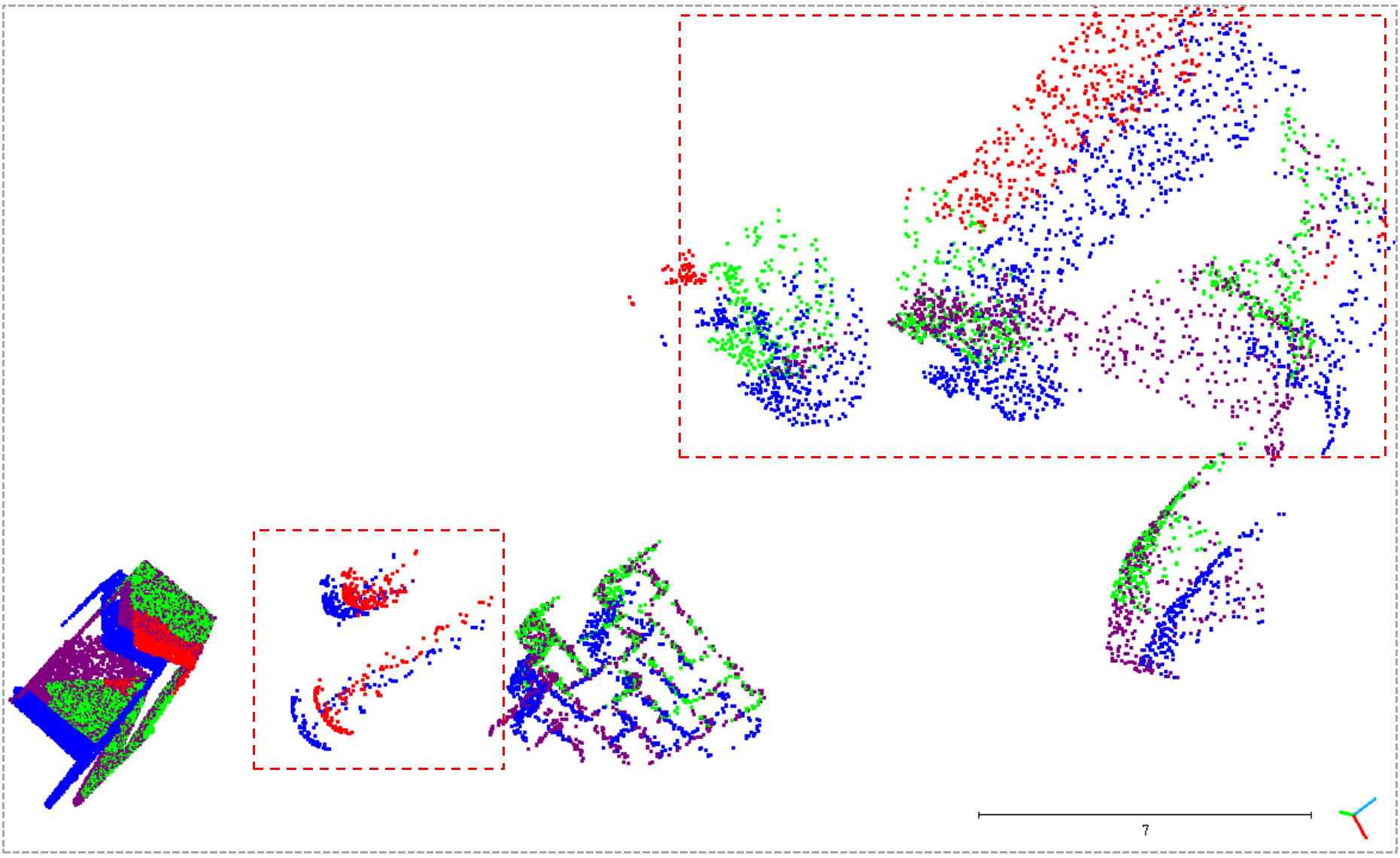}}
\hfil
\subfloat[]{
\includegraphics[width=\w\textwidth]{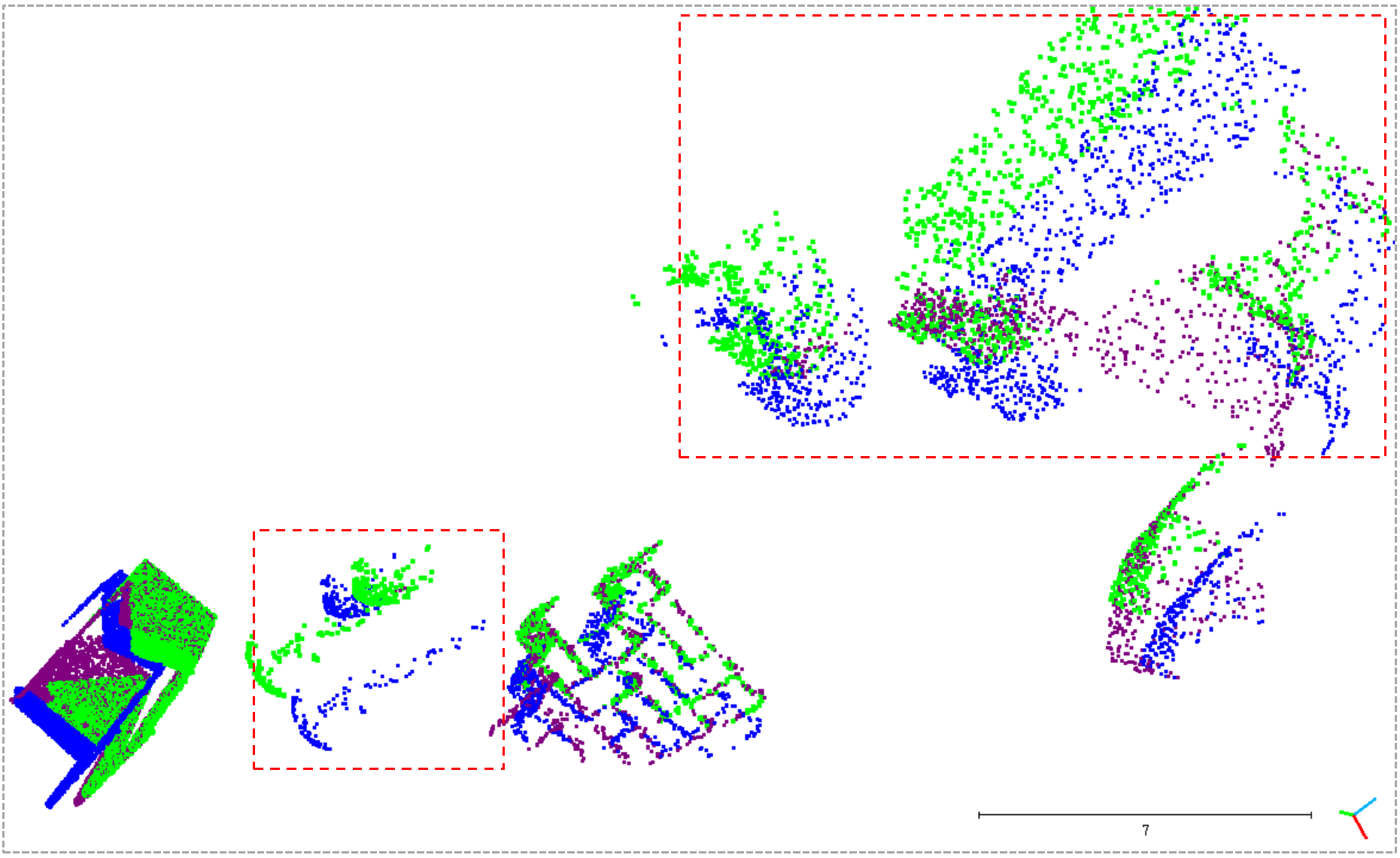}}
\caption{(a) The occlusion occurs between the \textcolor{blue}{source frame} and the \textcolor[RGB]{128,0,128}{target frame}. In this scenario, red bounding boxes delineate points in the source frame that vanish when transitioning to the target frame. (b) The \textcolor{green}{green points} represent the result of warping the source frame using our predictions. Here, the \textcolor{red}{red points} indicate incorrectly predicted warped points that have an EPE3D greater than 0.1m.
(c) The \textcolor{green}{green points} depict the result of warping the source frame using ground truth.}
\label{Fig: limits}
\end{figure}

\section{Limitations}
\label{sec:limitations}
Inferring motion flow for occluded objects has always been a challenging task. Particularly, in situations where objects are completely occluded, as defined in~\cite{ouyang2021-occSF-selfSF}, motion inference relying on the consistency of local motion becomes implausible. In real-world scenarios, taking into account roadway regulations, it might be feasible to estimate the movement of occluded vehicles on the lane by considering the overall directionality of traffic flow. However, for synthetic datasets like FT3Do, where random motion is assigned, it is challenging to infer the motion of completely occluded objects even for humans due to the unpredictable and random nature of the motion generation. Our model exhibits poor performance in some completely occluded synthetic scenes, as illustrated in Figure~\ref{Fig: limits}. An effective approach involves integrating a specific pattern of motion (object-wise relation) during the generation of a synthesized scene flow dataset.

\section{Related Works}
% \subsection{Coarse-to-Fine flow embedding}
% \subsection{All-to-All flow embedding}
% \subsection{Domain adaptive scene flow estimation}
% Scene flow, originally introduced in \cite{vedula1999three}, embodies the vectorized motion field characterizing the movements of objects within a 3D scene. 
% In previous research, RGB images were conventionally employed as the primary input \cite{vogel20153d,vogel2013piecewise,menze2015joint-kitti2015,ma2019deep}. However, given that scene flow prediction pertains to describing the motion of 3D real objects, recent advancements in 3D sensor technology and the emergence of deep learning networks designed specifically for point cloud processing have motivated a shift towards the direct utilization of raw point clouds as scene representations. This transition aims to enhance the accuracy of scene flow prediction, thereby ensuring more precise results. 
FlowNet3D~\cite{liu2019flownet3d} pioneers in leveraging deep learning network PointNet++\cite{qi2017pointnet++} for scene flow embedding based on raw point clouds, which surpasses traditional methods by a large margin. 
Afterward, HPLFlowNet\cite{gu2019hplflownet} proposed novel DownBCL, UpBCL, and CorrBCL operations inspired by  Bilateral Convolutional layers to abstract and fuse structural information from consecutive point clouds.
FlowNet3D++~\cite{wang2020flownet3d++-fullSF} enhances FlowNet3D by incorporating geometric constraints based on point-to-plane distance and angular alignment. FESTA~\cite{wang2021festa} expands on FlowNet3D by utilizing a trainable aggregate pooling to stably down-sample points instead of Farthest Point Sampling (FPS). 
These above methods employ the SetConv\cite{liu2019flownet3d}, composed of PointNet++, to conduct local flow embedding from two frames. However, this local embedding approach lacks global representation and fails to multi-scale processing.

Inspired by \cite{Sun_2018_CVPR_PWC_opticalflow} in optical flow,
PointPWC~\cite{wu2020pointpwc} incorporates the Pyramid, Warp, and Cost volume (PWC) to scene flow estimation. PointPWC utilizes semantic features from point cloud pyramids at different levels to generate the cost volume, which is then used to compute patch-to-patch local flow embedding.
HALF~\cite{wang2021hierarchical-fullSF} introduces a novel double attentive flow embedding in cost volume. RMS-FLowNet~\cite{battrawy2022rms-fullSF-largeScale} integrates random sampling to efficiently process large-scale scenes instead of FPS. Inspired by BERT~\cite{devlin2019bert}, Bi-PointFlow~\cite{cheng2022bi-flow} applies bidirectional flow embedding to produce cost volume using the sequence information. Res3DSF~\cite{wang2022residual-Res3DSF-fullSF} presents a novel context-aware set convolution layer to enhance the detection of recurrent patterns in 3D space. Nonetheless, these coarse-to-fine methods focus on local flow regression layer-by-layer which lacks global information. To address this issue, 
Some methods adopt an all-to-all approach. FLOT\cite{puy2020flot} redefines scene flow prediction as an optimal transmission problem, gauging the transmission cost by evaluating the global cosine similarity of semantic features. FlowStep3D~\cite{kittenplon2021flowstep3d} aims to directly compute the initial scene flow by leveraging an unlearnable feature similarity matrix in the global unit. WM3DSF~\cite{wang2022whatmatters} proposes an all-to-all point mixture module with backward reliability validation. Additionally, PT-Flow~\cite{fu2023pt-flownet-transformerSF-fullSF} and PV-RAFT equip the point-voxel branches to the flow embedding to enlarge the receptive field.

However, these methods solely consider the individual point cloud semantic to make a hard flow embedding, which lacks fusion of the global semantic context of another embedded frame. We propose the GF module to fuse the semantic features of two frames and perceive the global correlation in each other's context, which is essentially in the global flow estimation of long-range and complex geometric situations such as occluded\cite{jiang2021learning-gma} and repetitive\cite{wang2022residual-Res3DSF-fullSF} pattern.

\section{Datasets}
\label{sup-sec::datasets}
Experiments are conducted on four datasets, namely the synthetic dataset FlyThings3D (FT3D)~\cite{mayer2016large-fly3d} and three real-world datasets, including Stereo-KITTI~\cite{menze2015joint-kitti2015,menze2018object-kitti2018}, LiDAR-KITTI~\cite{geiger2012we-lidar-kt}, and SF-KITTI~\cite{ding2022fhnet-flow-SFKT}.

The synthetic dataset FT3D is derived from a large-scale collection of stereo videos. Each pair of point cloud scenes is generated from RGB stereo images in the ShapeNet~\cite{chang2015shapenet}, with random motion assigned to multiple objects within the scenes. Another stereo dataset Stereo-KITTI is a real-world dataset comprising 200 training sets and 200 testing sets. Building on the previous literature, two distinct preprocessing techniques are employed for these two datasets. A technique from~\cite{gu2019hplflownet} is utilized to remove points that do not correspond between consecutive frames, resulting in processed datasets referred to as FT3Ds and KITTIs. The second technique, proposed by~\cite{liu2019flownet3d}, takes a different approach by preserving the occluded points. Instead of removing them, mask labels are used to indicate occluded points for evaluation purposes. The datasets generated using this method are referred to as FT3Do and KITTIo. The FT3Ds dataset comprises 19,640 pairs of training data and 3,824 pairs for evaluation. On the other hand, the FT3Do dataset consists of 20,008 point cloud pairs for training and 2,008 pairs for testing. Regarding the KITTIs and KITTIo datasets, they contain 142 and 150 pairs of test-only data, respectively.

However, it is noteworthy that the FT3D and Stereo-KITTI datasets, which are derived from dense and regular disparity images, differ from real-world LiDAR-scanned datasets. To demonstrate the robust generalization of our SSRFlow on LiDAR-scanned datasets, we conducted additional experiments using datasets~\cite{ding2022fhnet-flow-SFKT,gojcic2021weakly-rigid3dSF,geiger2012we-lidar-kt}. Specifically, SF-KITTI~\cite{ding2022fhnet-flow-SFKT} is a large-scale real-world scene flow dataset that is based on LiDAR-scanned data. It comprises 7,185 pairs of data. Following~\cite{ding2022fhnet-flow-SFKT},  we further divide the dataset into 6,400 pairs for training and 600 pairs for testing. In addition, the LiDAR-KITTI~\cite{geiger2012we-lidar-kt,gojcic2021weakly-rigid3dSF}, dataset includes 142 pairs of real-world scenarios captured by the Velodyne 64-beam LiDAR, specifically designed for testing purposes. This dataset is particularly valuable as it highlights the sparse and non-corresponding point characteristics that are often observed between consecutive frames of the real-world LiDAR-scanned point clouds.

\section{More Results}
\label{sup-sec::moreResults}

\def\w{0.32}
\begin{figure}[h] \footnotesize
\centering
\captionsetup[subfloat]{labelsep=none,format=plain,labelformat=empty}
\subfloat{
\includegraphics[width=\w\textwidth, frame]{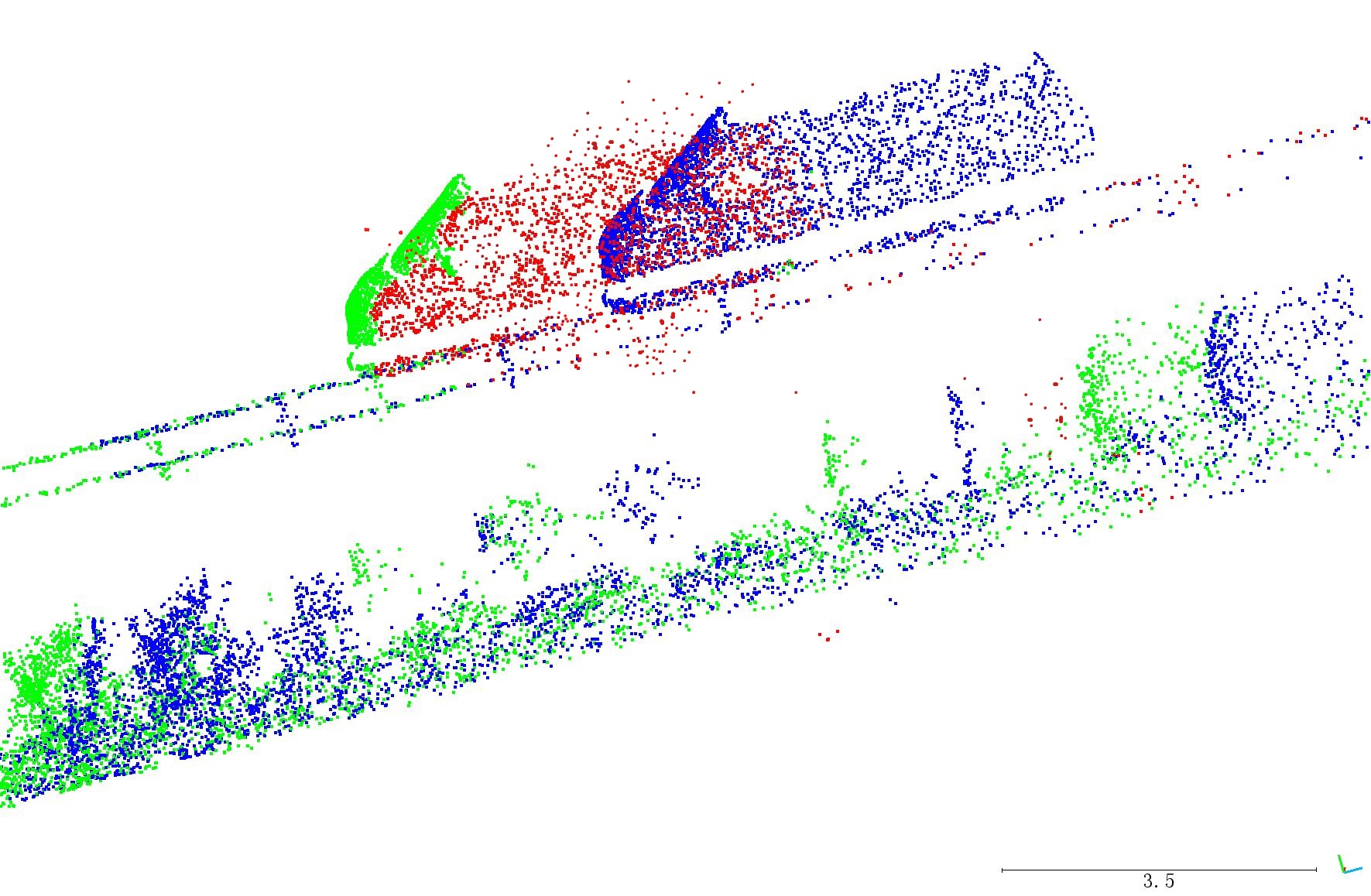}}
\hfil
\subfloat{
\includegraphics[width=\w\textwidth, frame]{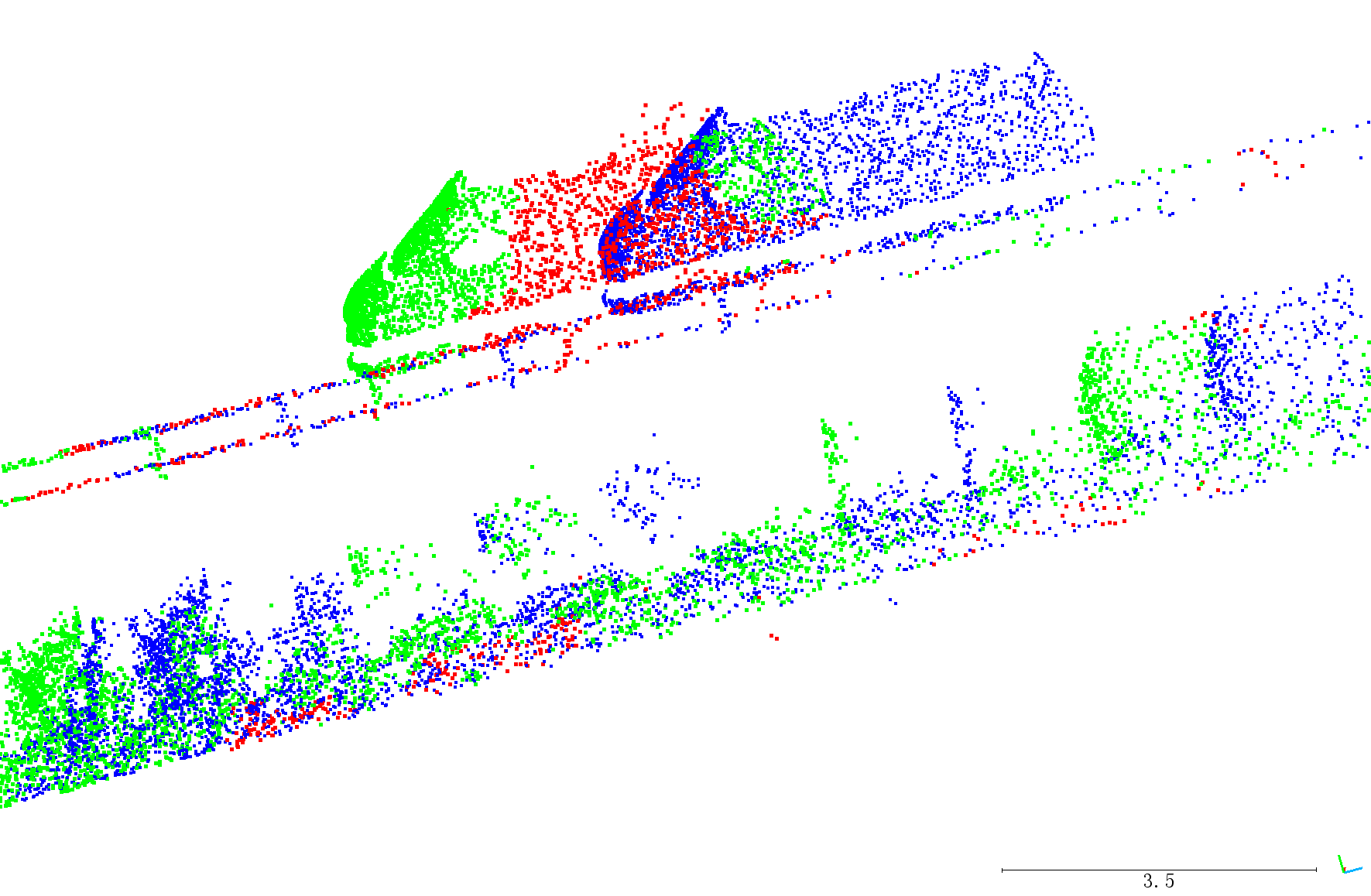}}
\hfil
\subfloat{
\includegraphics[width=\w\textwidth, frame]{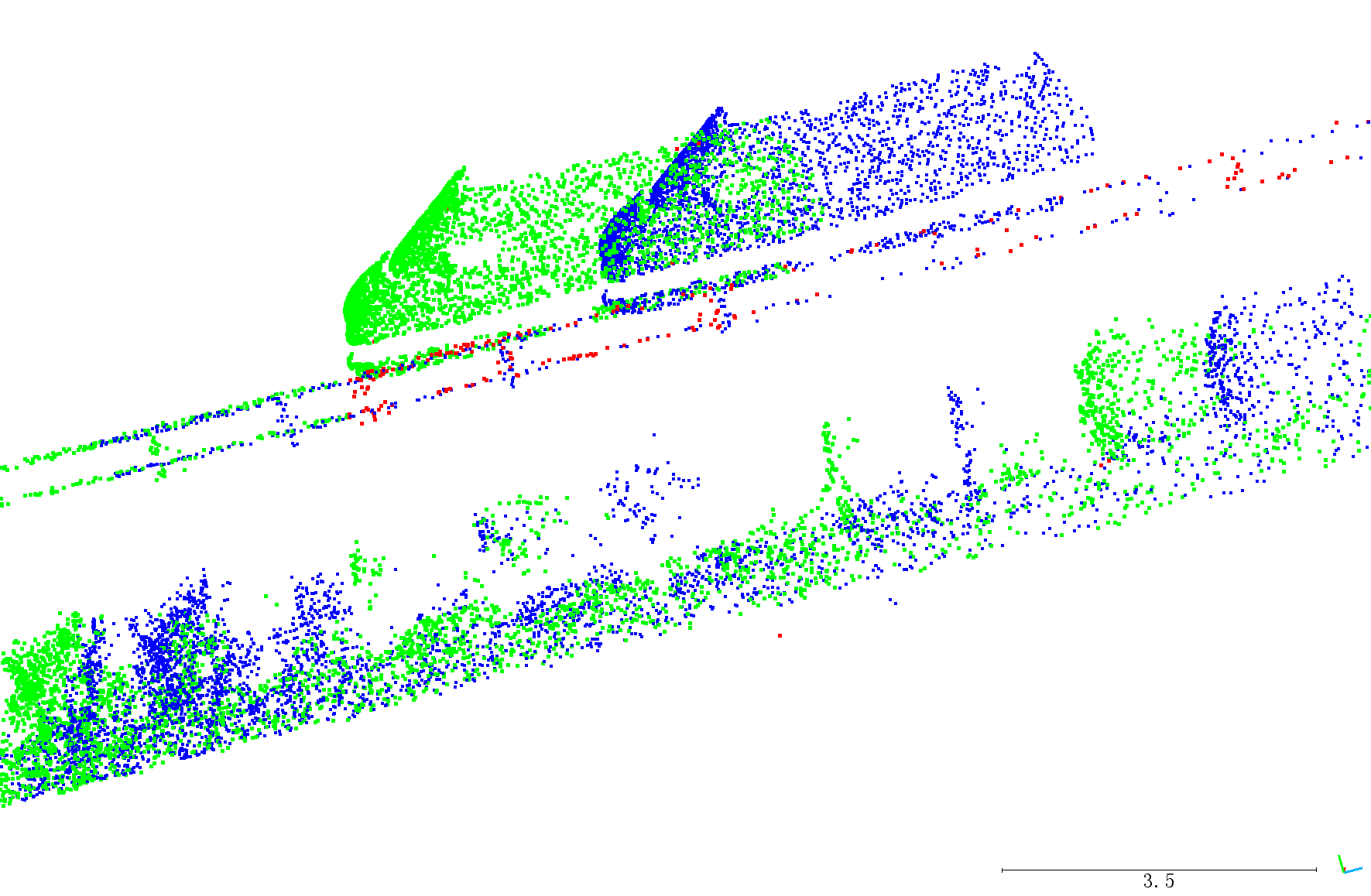}}

\subfloat{
\includegraphics[width=\w\textwidth, frame]{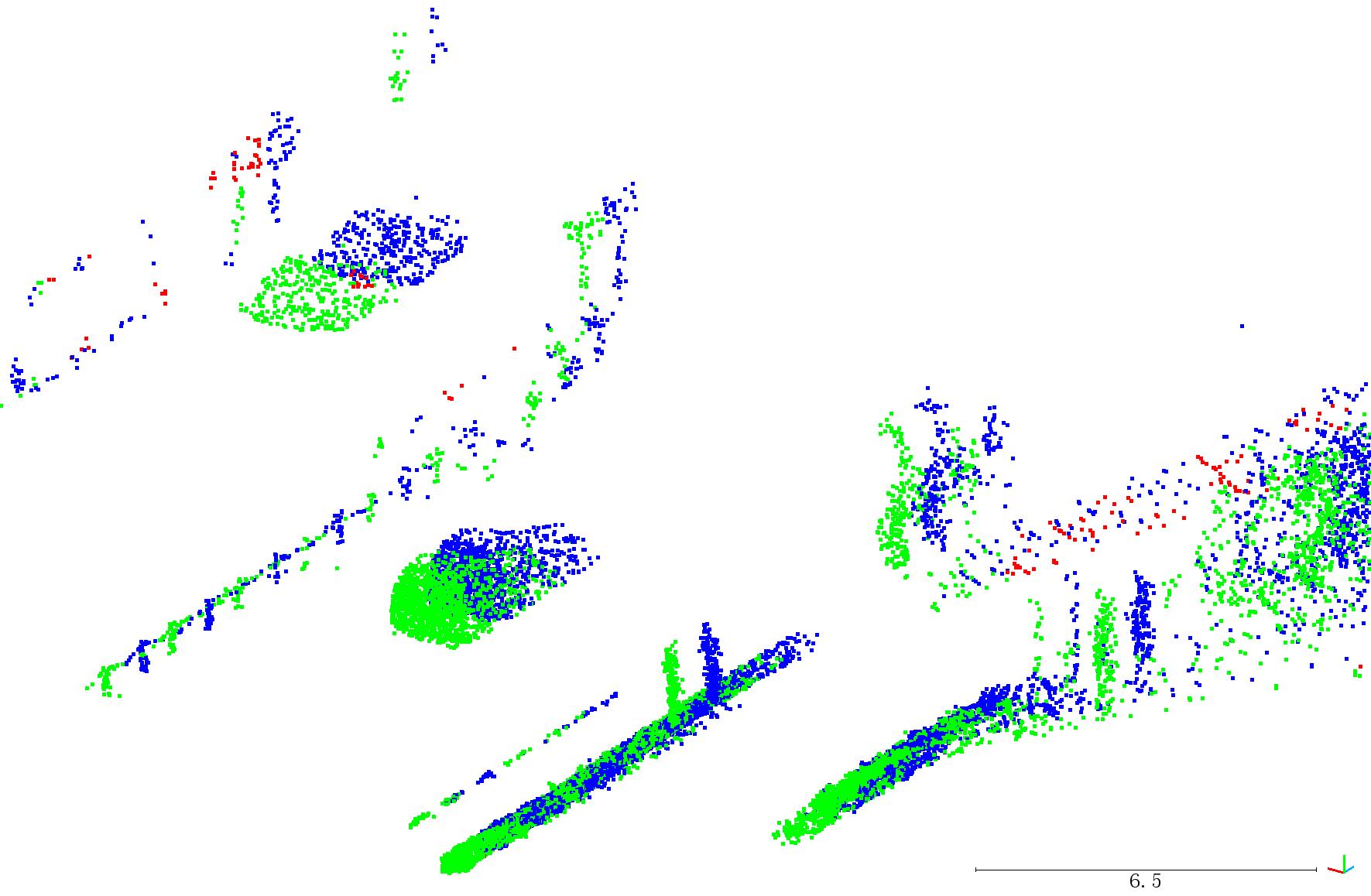}}
\hfil
\subfloat{
\includegraphics[width=\w\textwidth, frame]{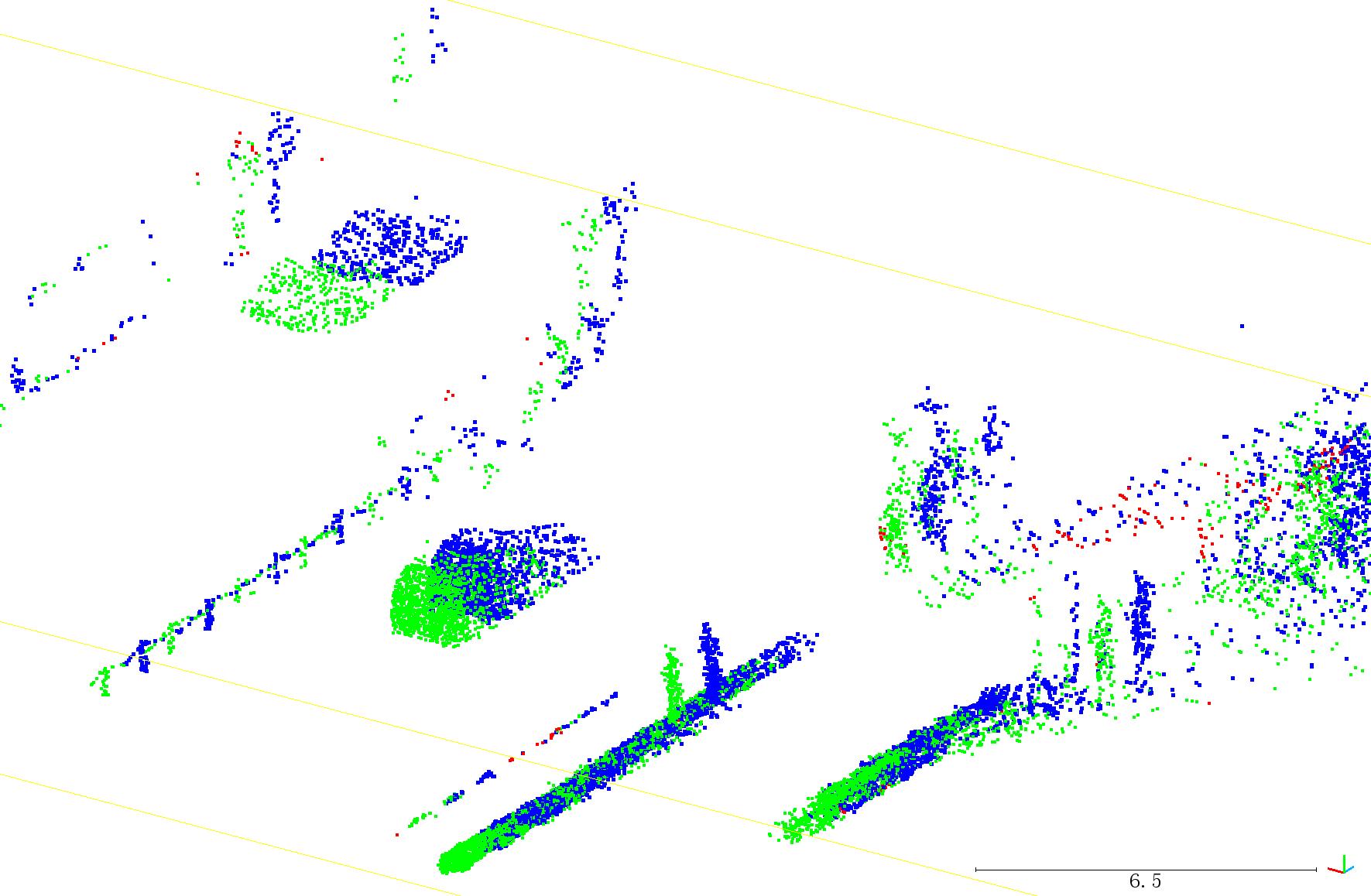}}
\hfil
\subfloat{
\includegraphics[width=\w\textwidth, frame]{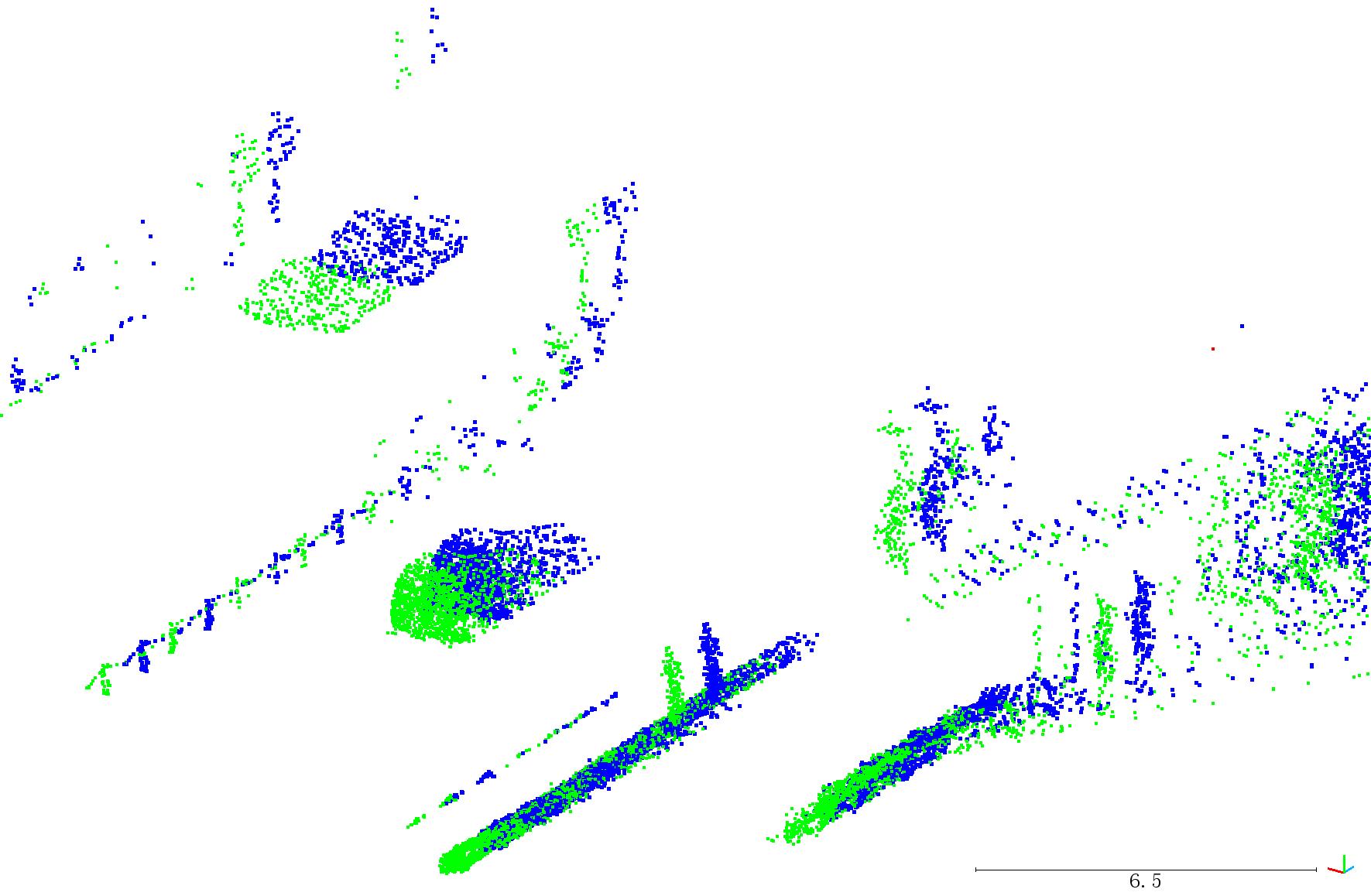}}

\subfloat{
\includegraphics[width=\w\textwidth, frame]{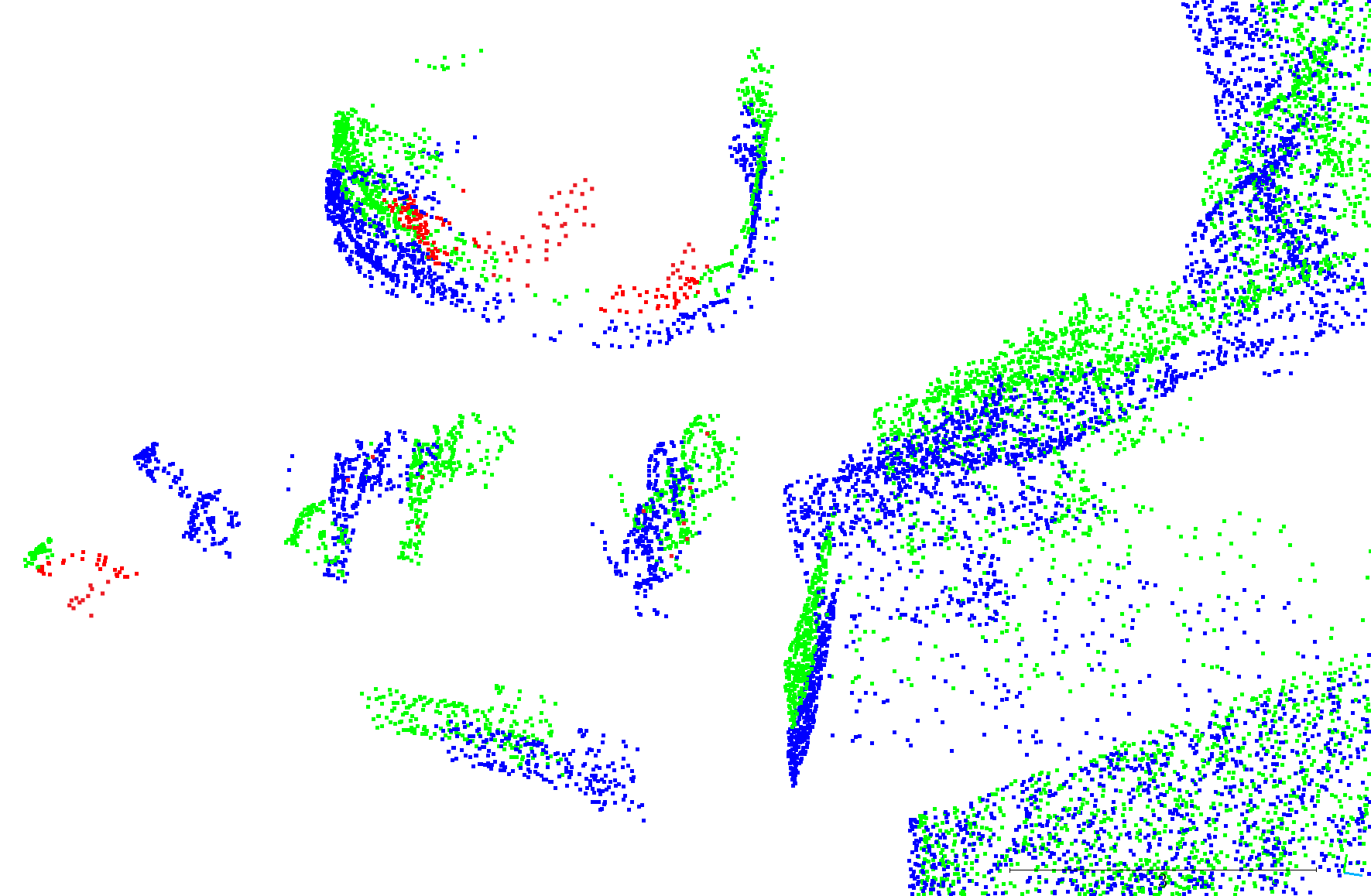}}
\hfil
\subfloat{
\includegraphics[width=\w\textwidth, frame]{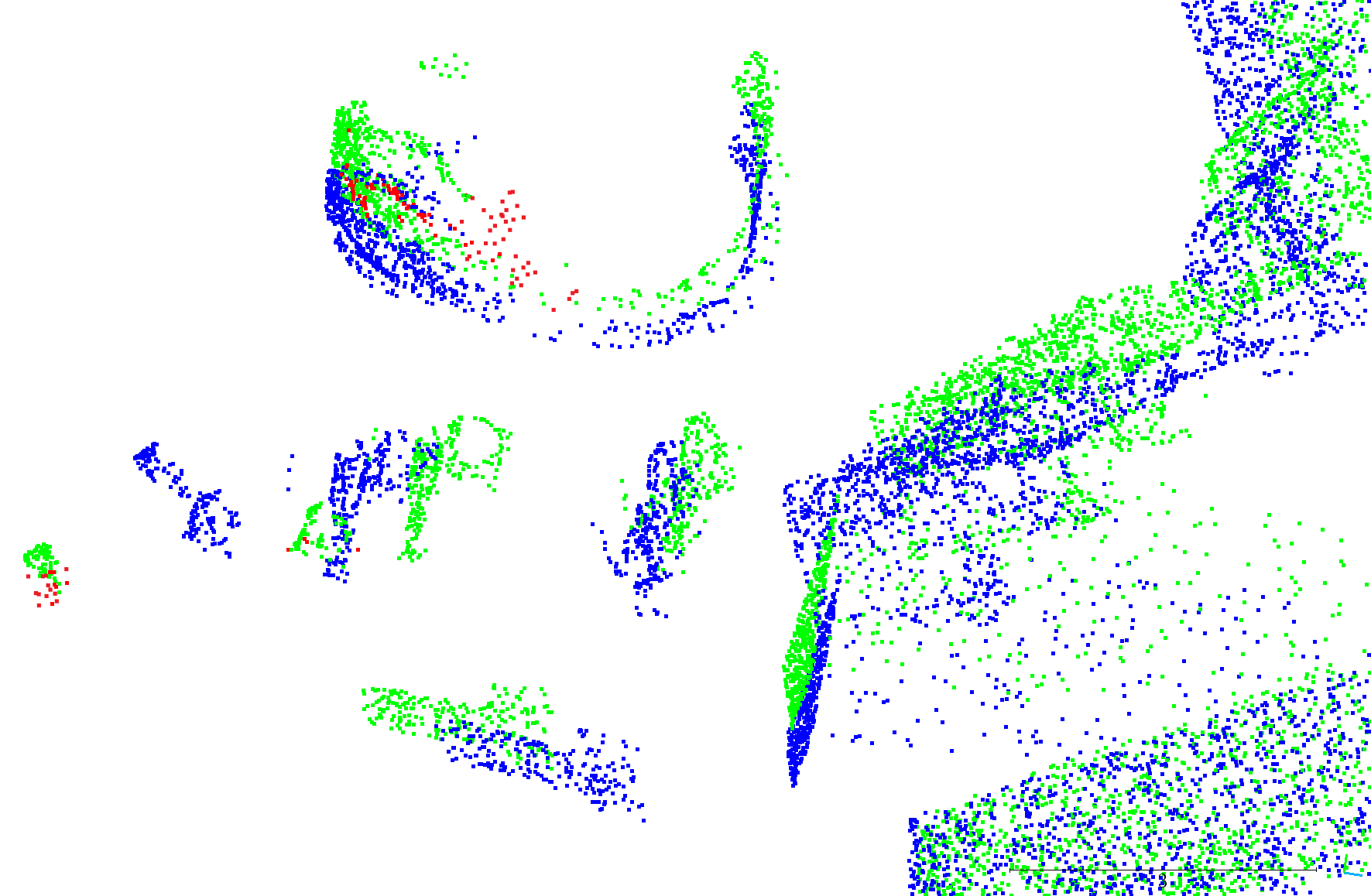}}
\hfil
\subfloat{
\includegraphics[width=\w\textwidth, frame]{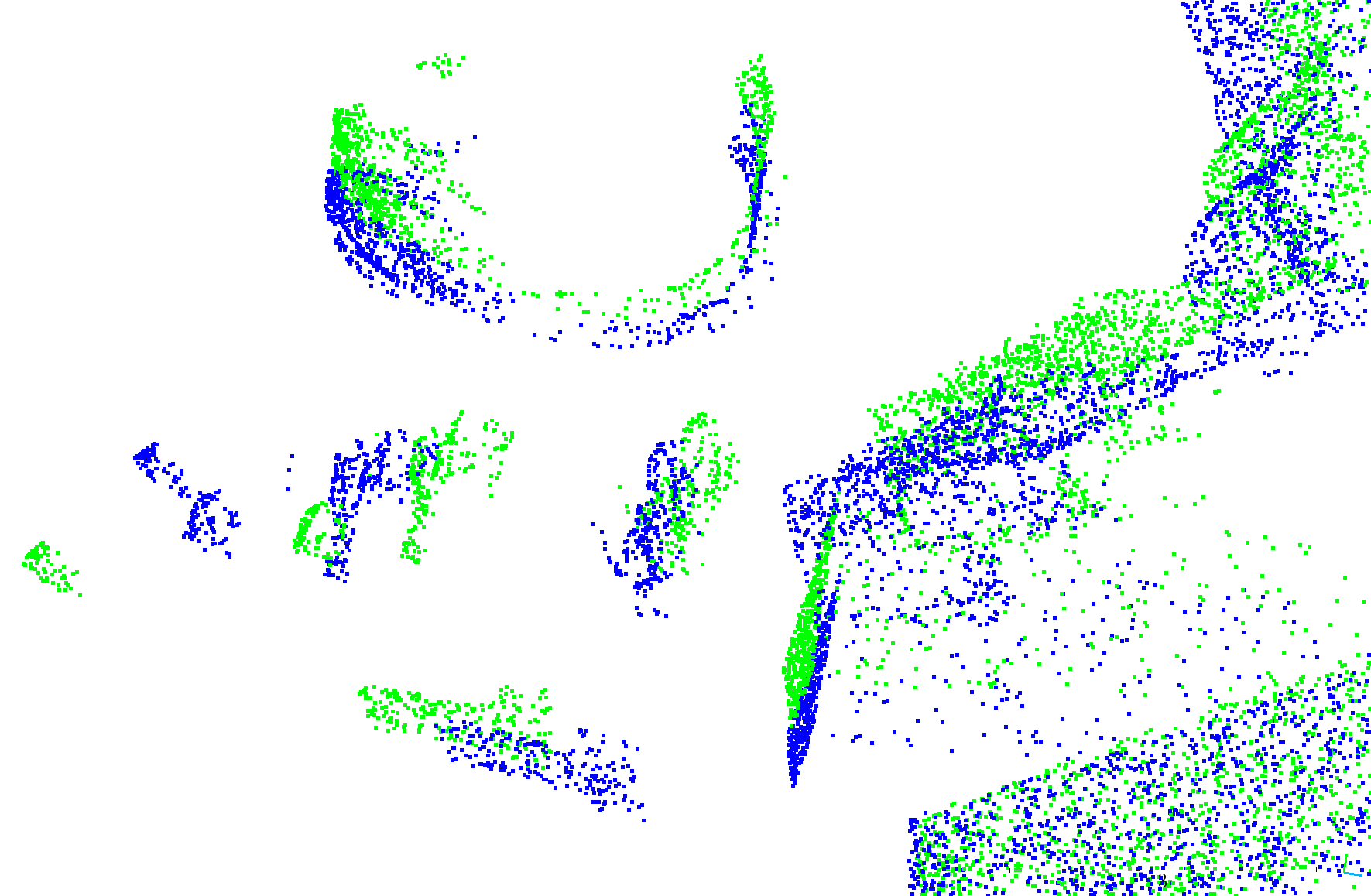}}

\subfloat{
\includegraphics[width=\w\textwidth, frame]{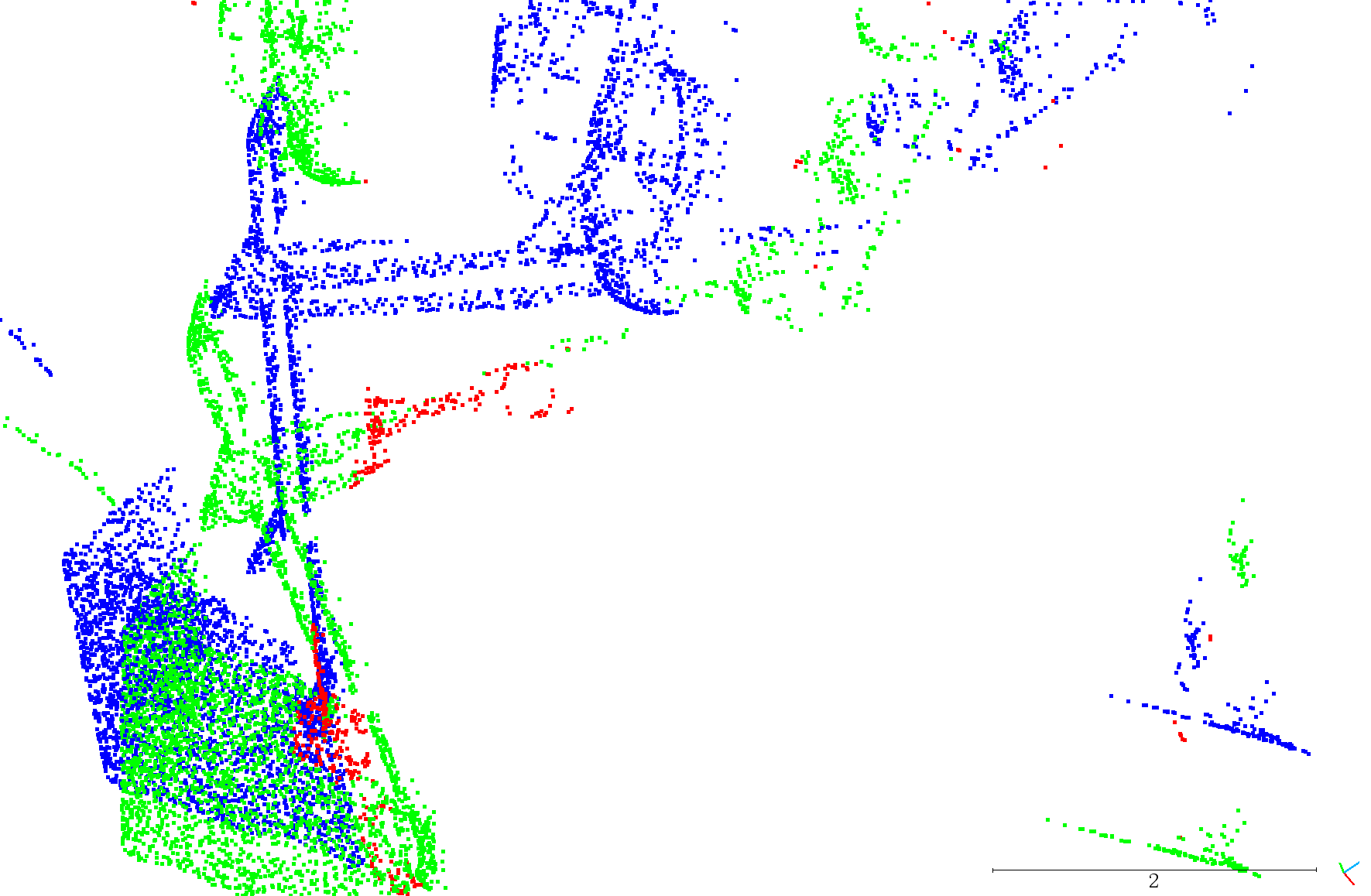}}
\hfil
\subfloat{
\includegraphics[width=\w\textwidth, frame]{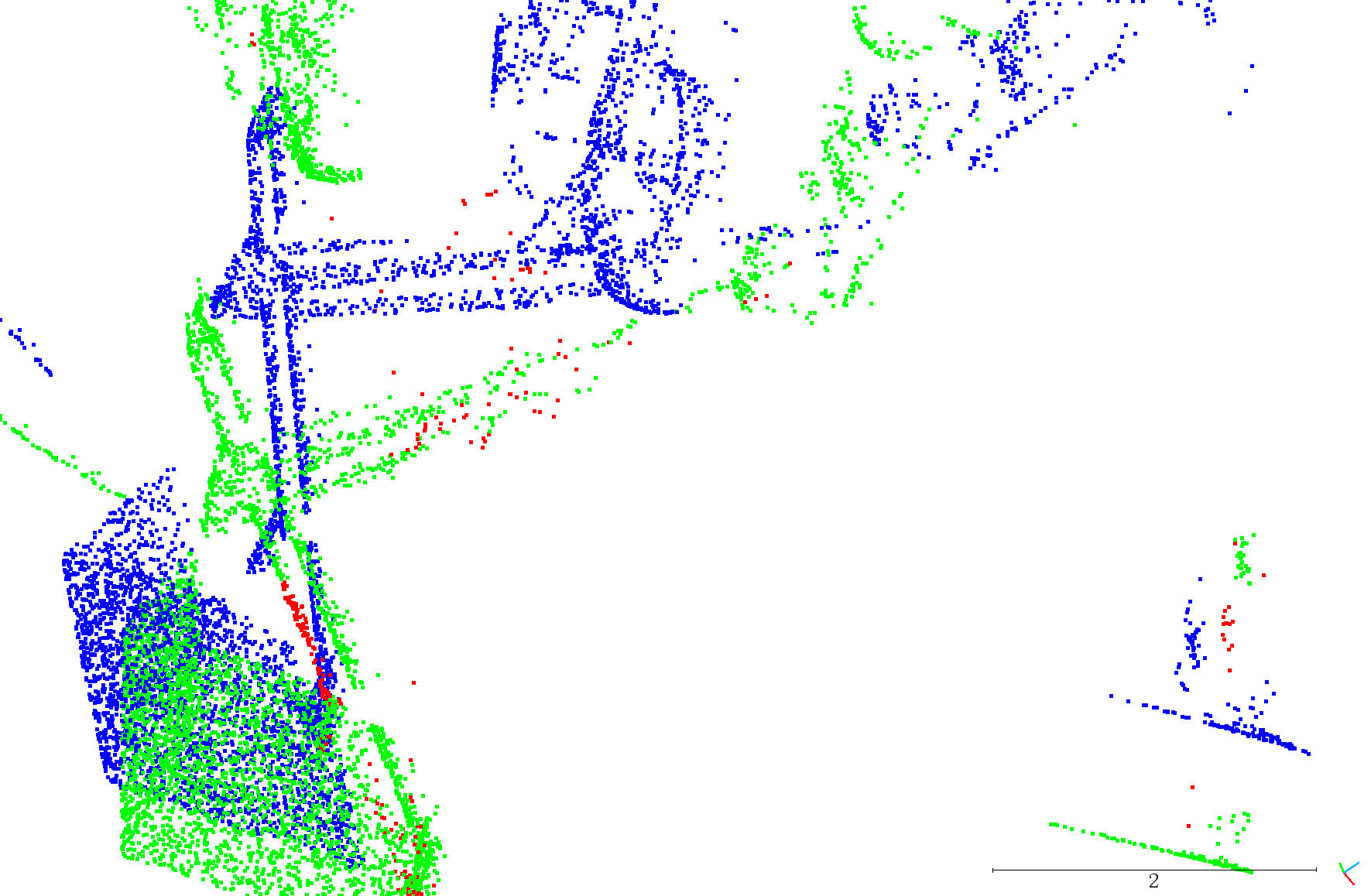}}
\hfil
\subfloat{
\includegraphics[width=\w\textwidth, frame]{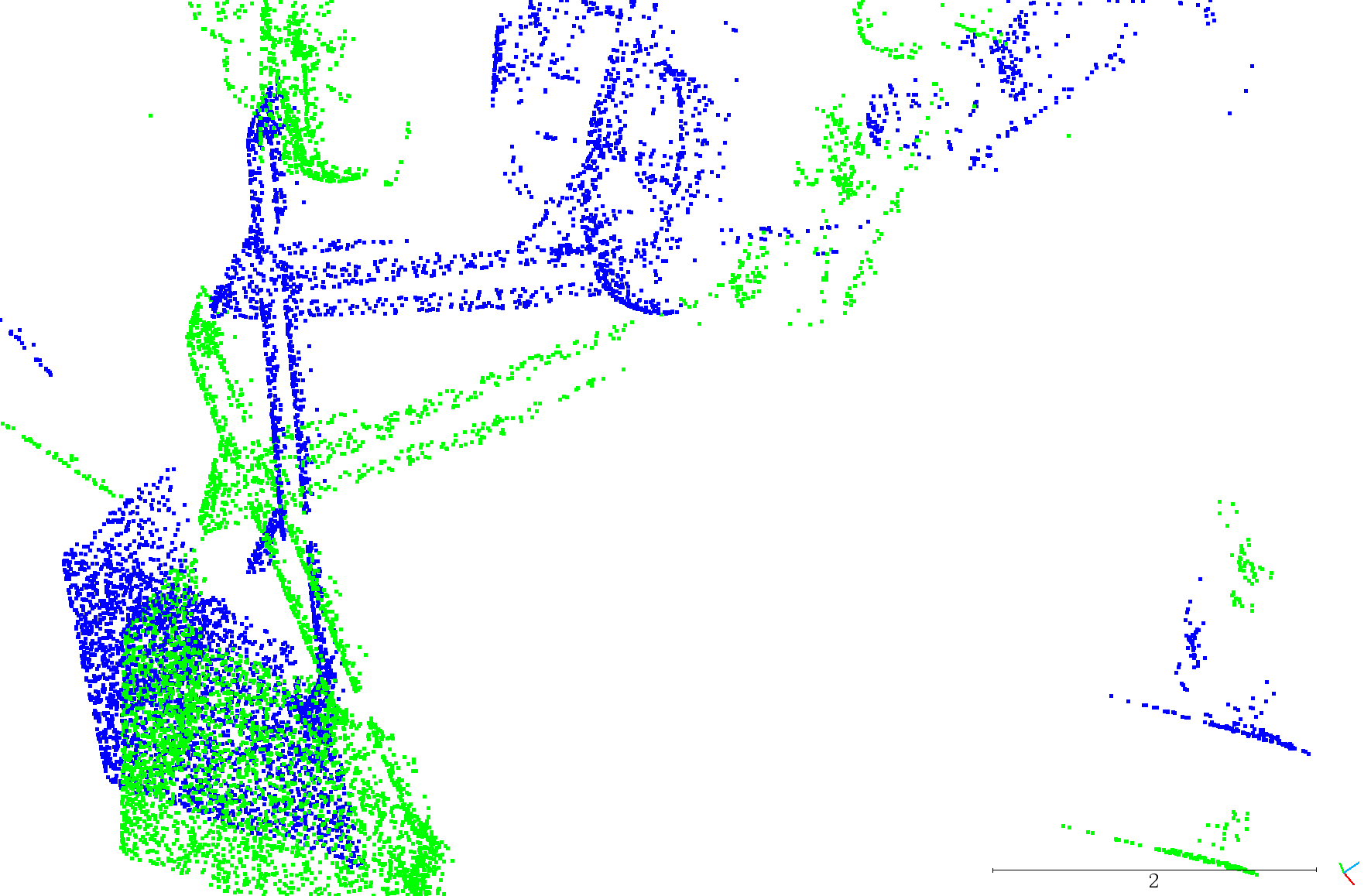}}

\end{figure}

\begin{figure}[h] \footnotesize
\centering
\captionsetup[subfloat]{labelsep=none,format=plain,labelformat=empty}

\subfloat{
\includegraphics[width=\w\textwidth, frame]{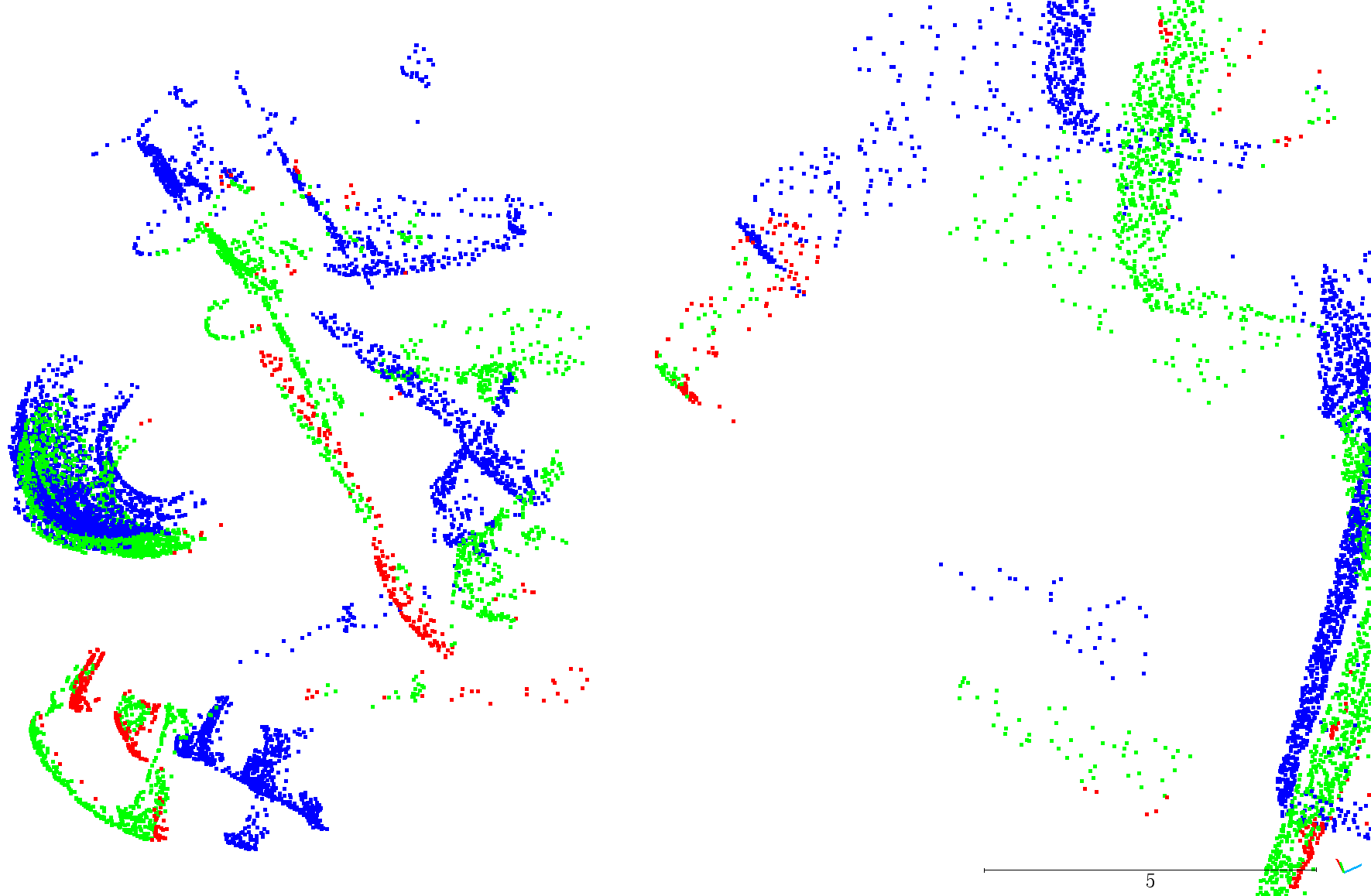}}
\hfil
\subfloat{
\includegraphics[width=\w\textwidth, frame]{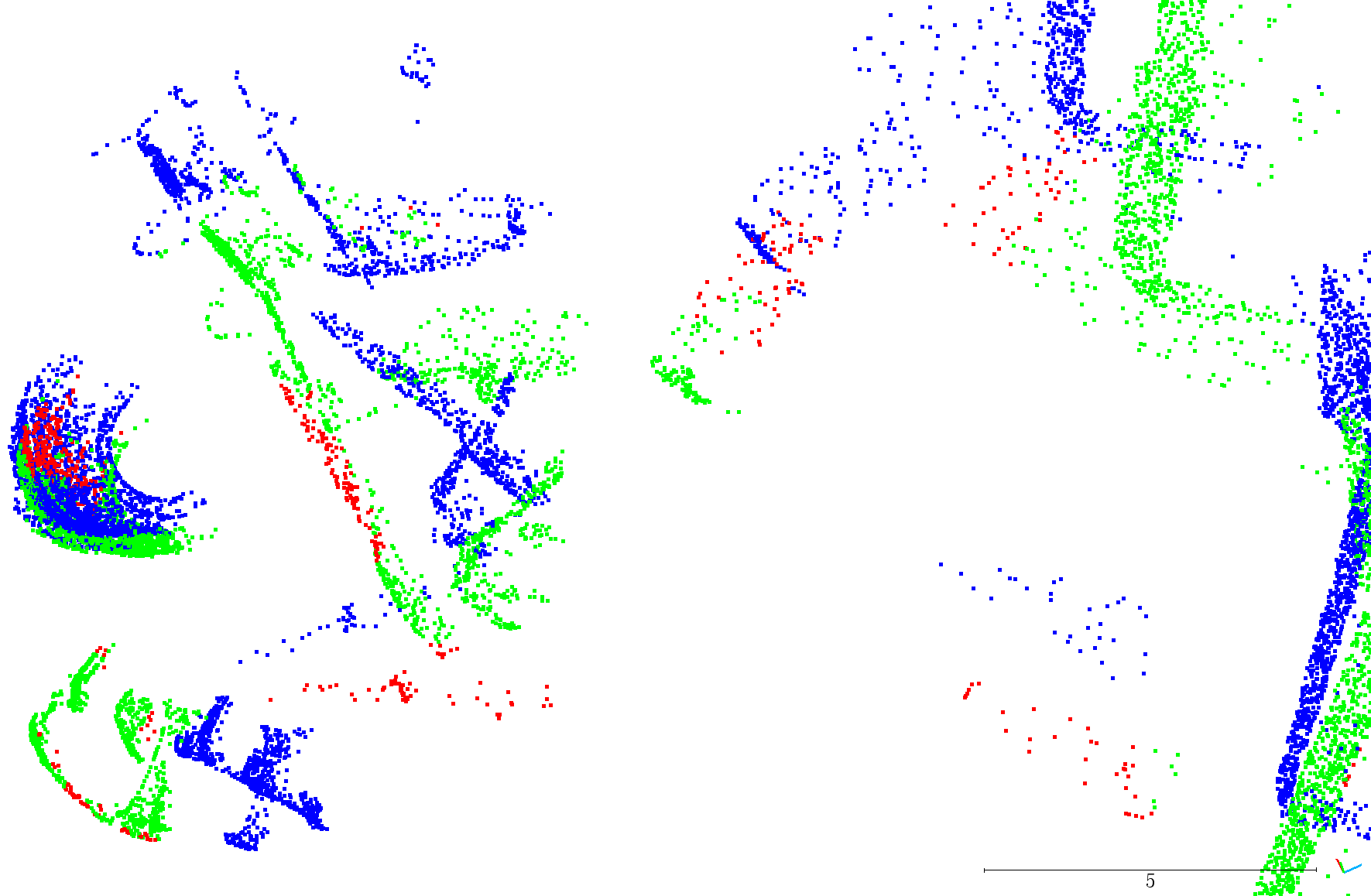}}
\hfil
\subfloat{
\includegraphics[width=\w\textwidth, frame]{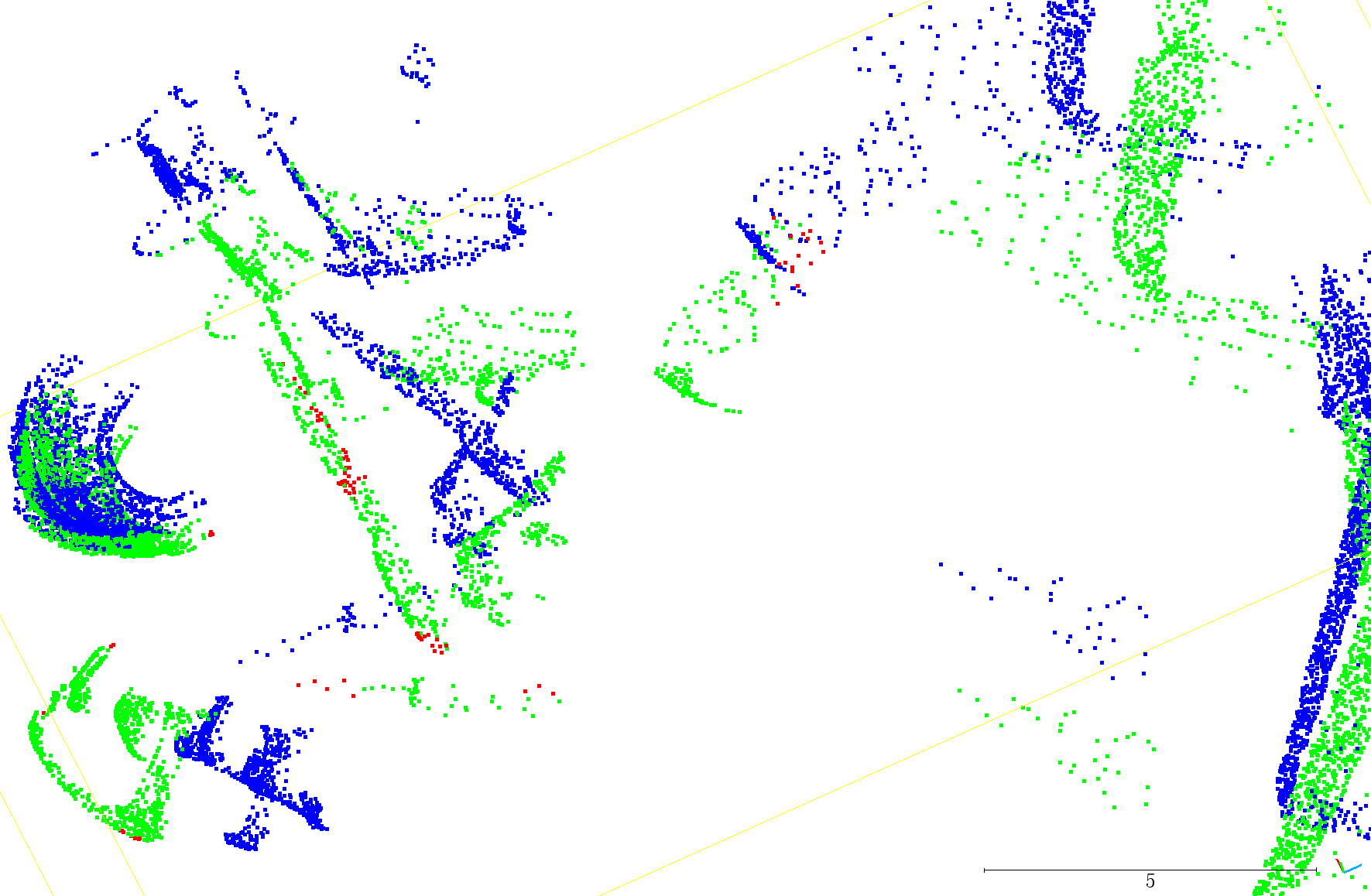}}

\subfloat{
\includegraphics[width=\w\textwidth, frame]{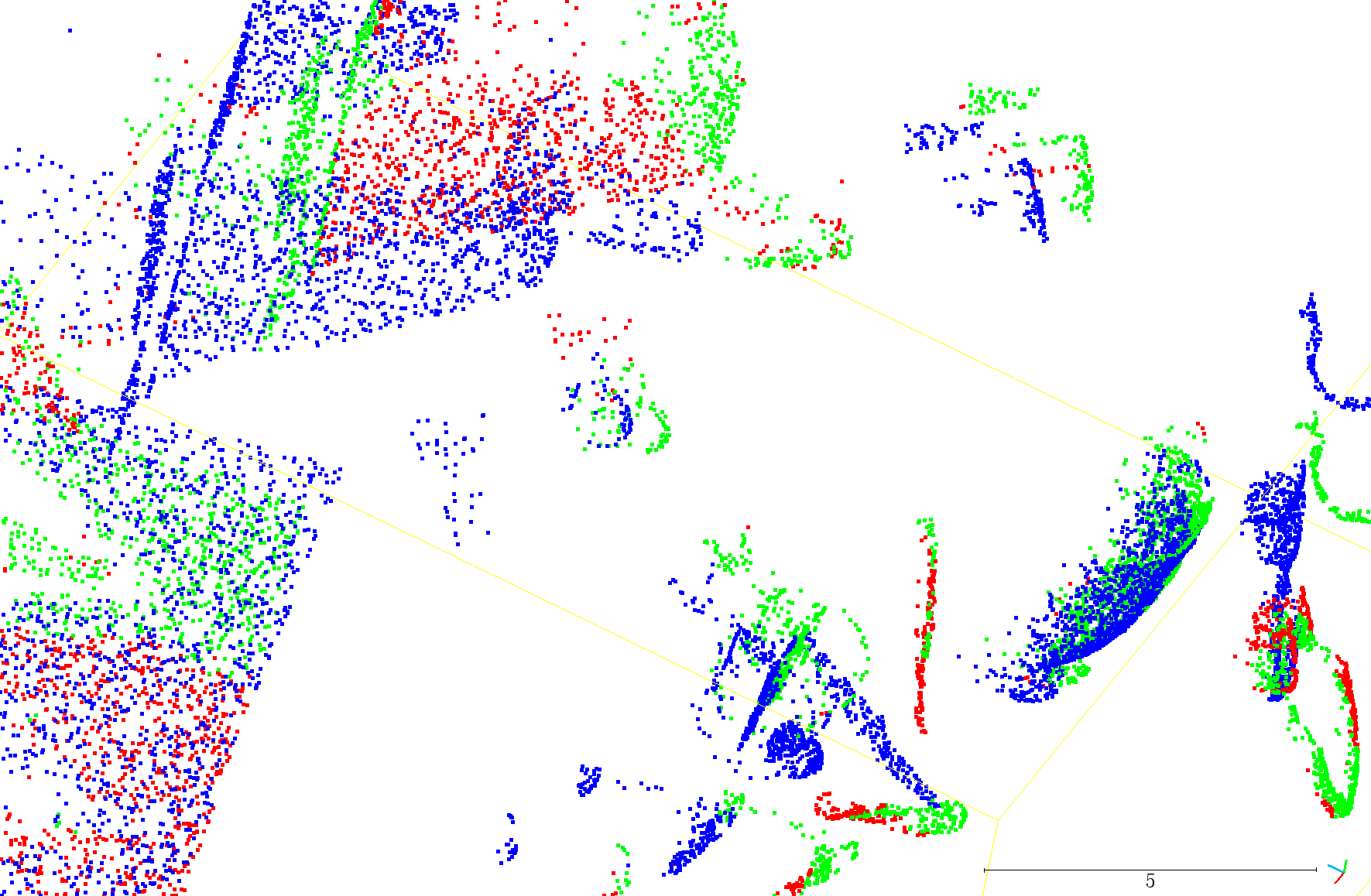}}
\hfil
\subfloat{
\includegraphics[width=\w\textwidth, frame]{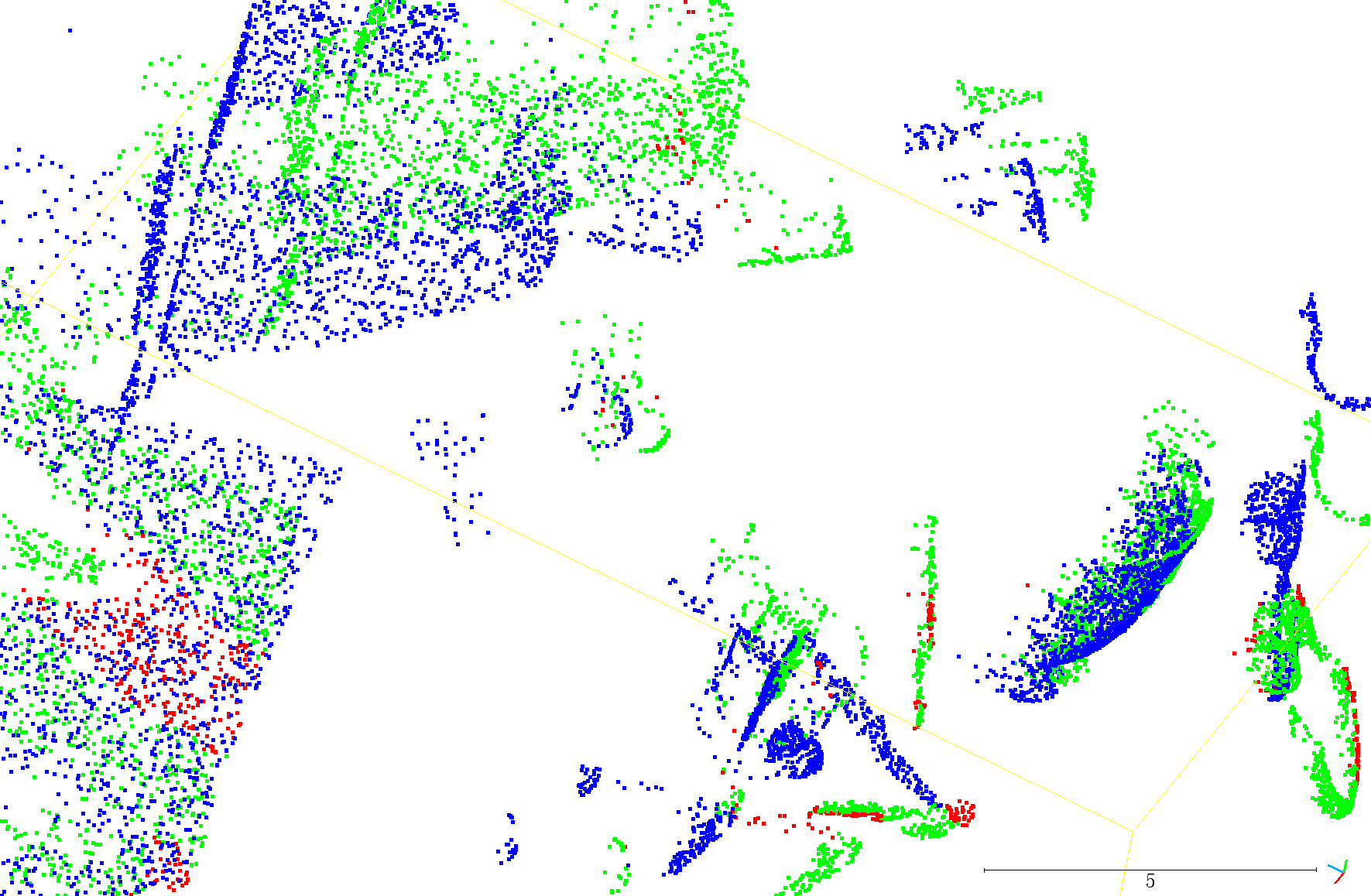}}
\hfil
\subfloat{
\includegraphics[width=\w\textwidth, frame]{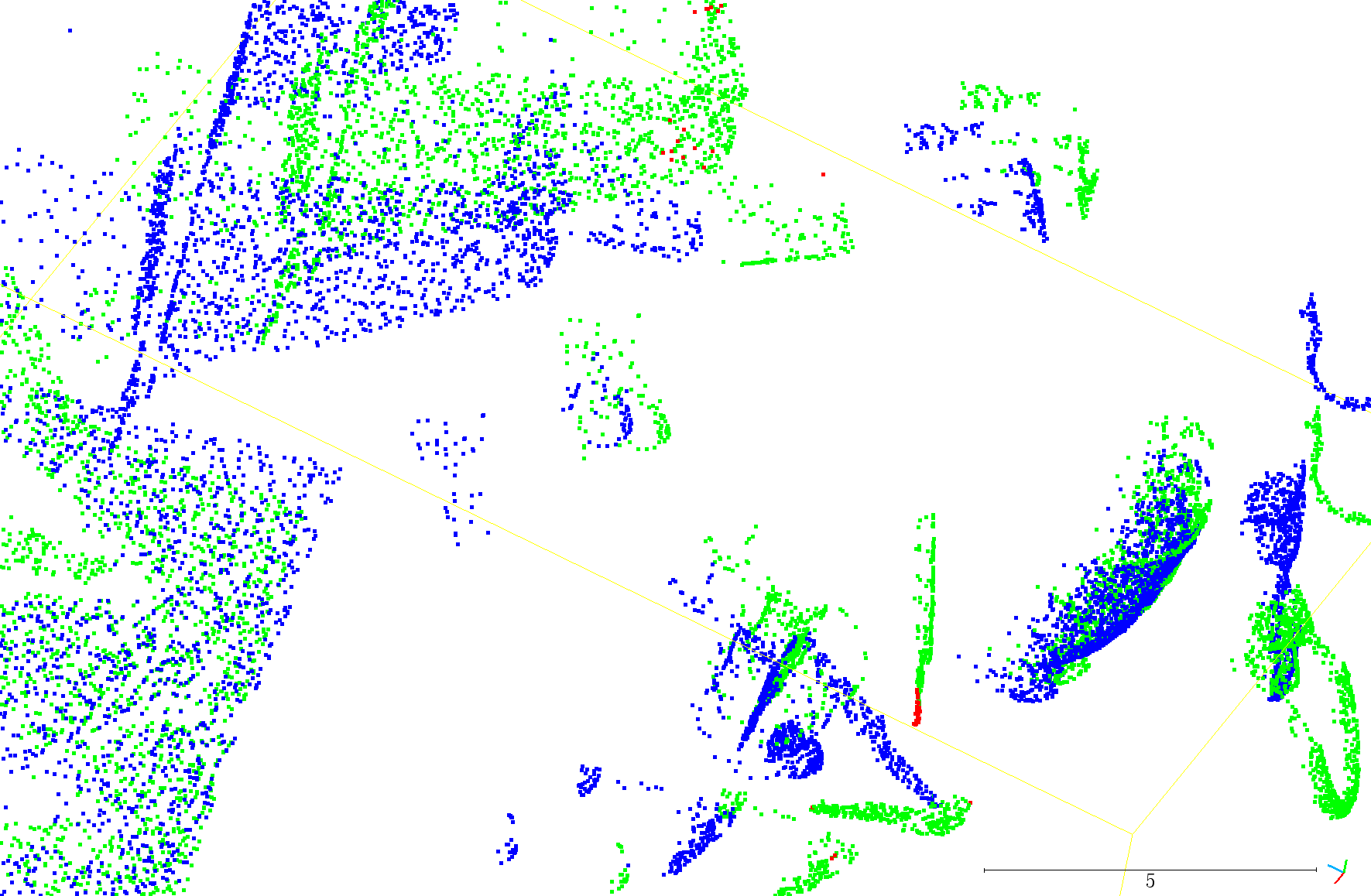}}

\subfloat[PointPWC\cite{wu2020pointpwc}]{
\includegraphics[width=\w\textwidth, frame]{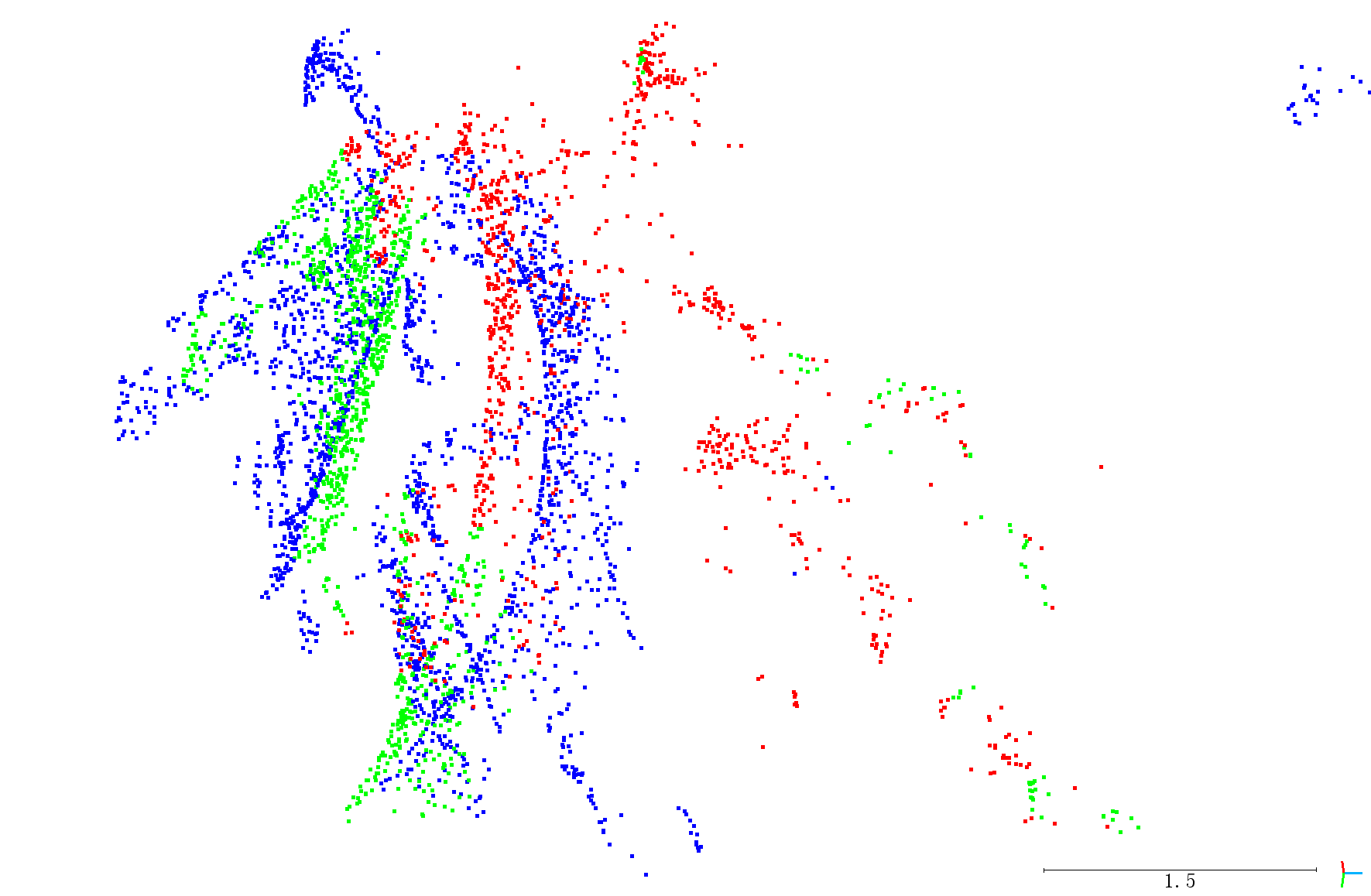}}
\hfil
\subfloat[Bi-PointFlow\cite{cheng2022bi-flow}]{
\includegraphics[width=\w\textwidth, frame]{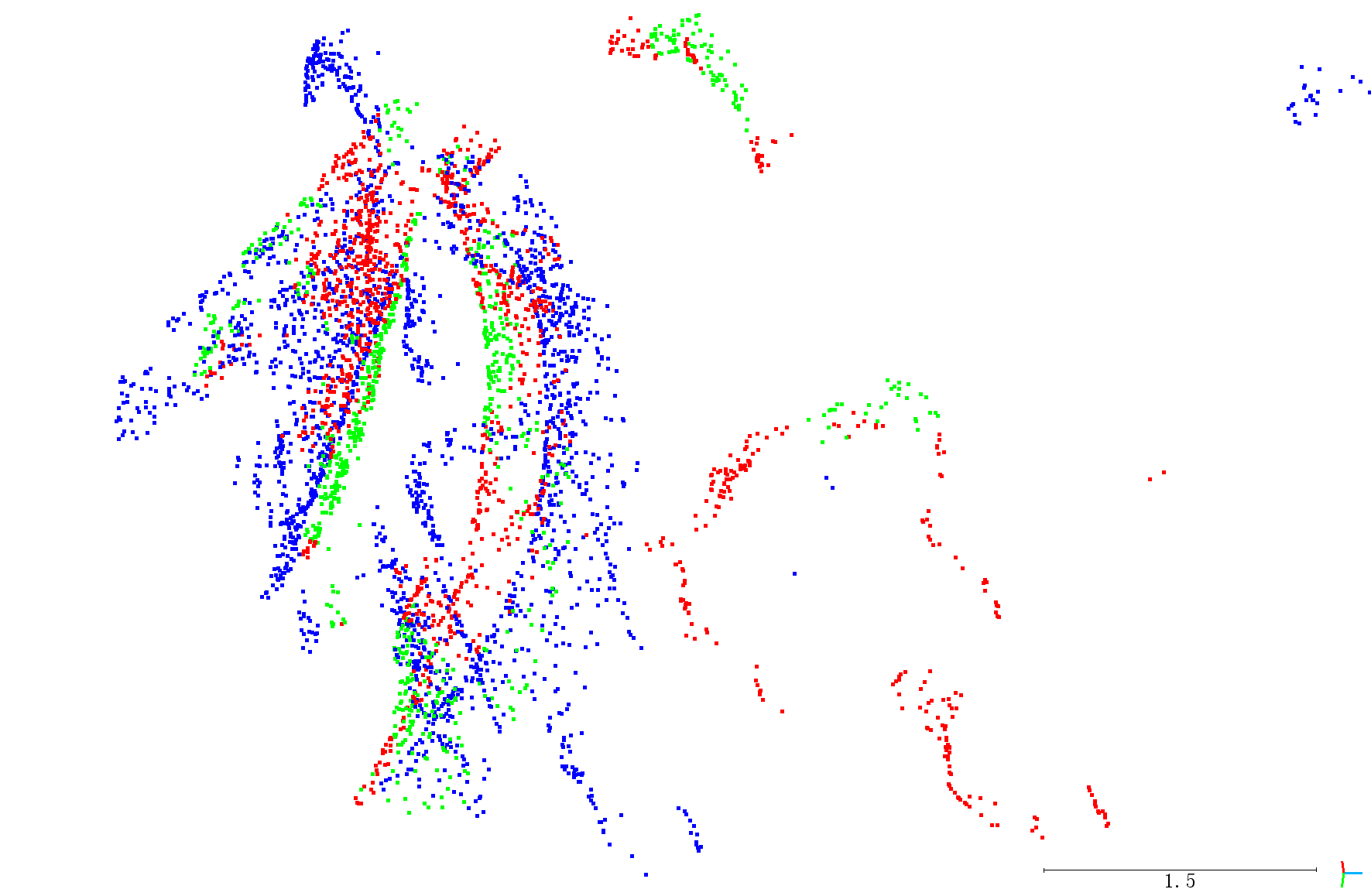}}
\hfil
\subfloat[SSRFlow]{
\includegraphics[width=\w\textwidth, frame]{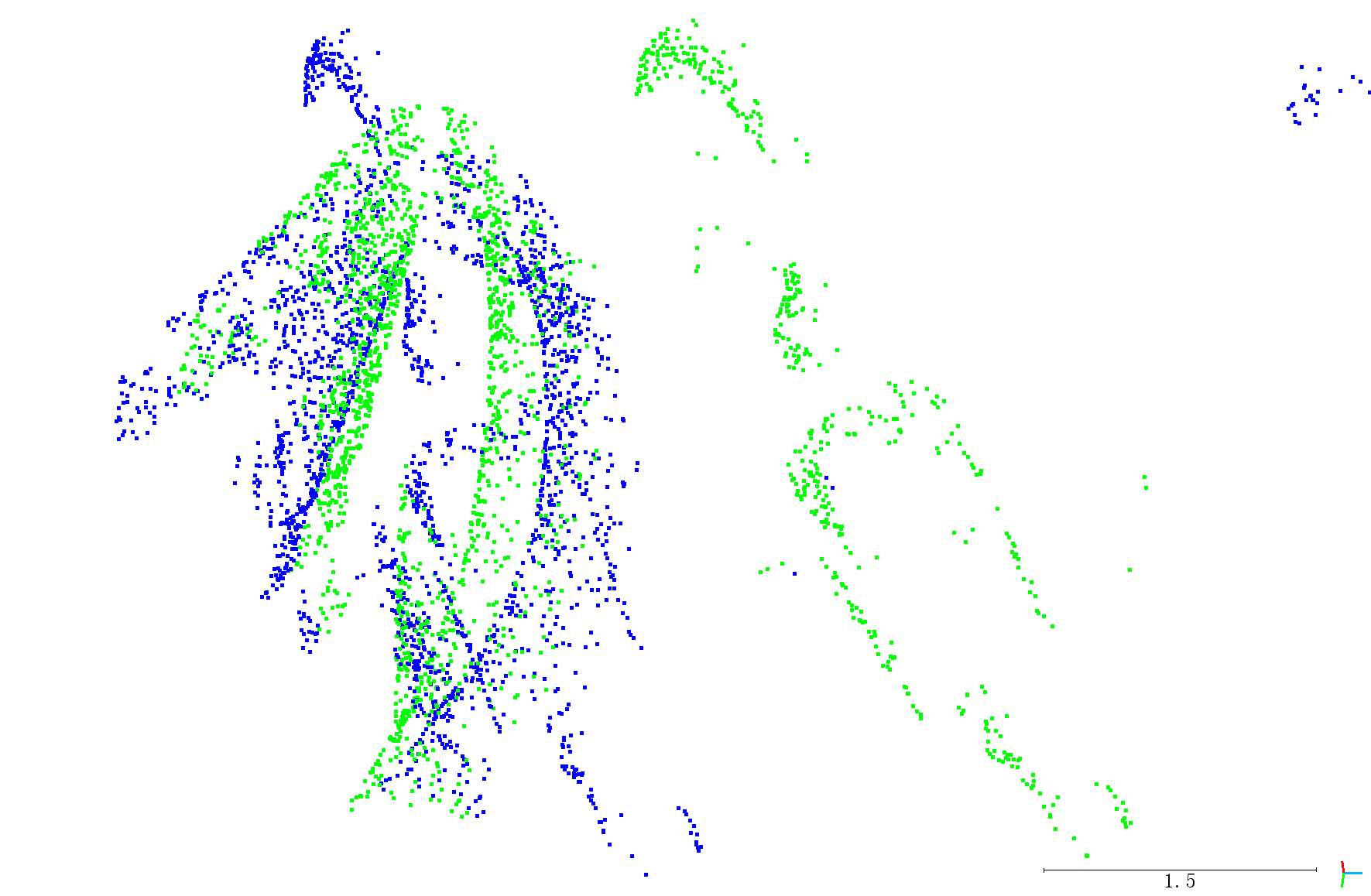}}

\caption{The \textcolor{blue}{blue points} denote the source frame. The \textcolor{green}{green points} represent the result of warping the source frame using predictions. Here, the \textcolor{red}{red points} indicate wrong predicted warped points that have an EPE3D greater than 0.1m.}
\label{SupFig::moreRes}
\end{figure}

\end{document}